\pdfoutput=1

\documentclass[11pt]{article}

\usepackage[final]{acl}

\usepackage{times}
\usepackage{latexsym}

\usepackage[T1]{fontenc}

\usepackage[utf8]{inputenc}

\usepackage{microtype}

\usepackage{inconsolata}

\usepackage{enumitem}
\setlist[itemize]{noitemsep, topsep=0pt}
\usepackage{multirow}
\usepackage{graphicx}
\usepackage{pifont}
\usepackage{booktabs}
\usepackage{subfigure}
\usepackage{algorithm}
\usepackage{algorithmic}
\usepackage{makecell}

\usepackage{amsfonts}
\usepackage{amsmath}
\usepackage{amssymb}
\usepackage{mathrsfs}
\usepackage{xcolor,colortbl}
\usepackage{caption}
\usepackage{subcaption}

\definecolor{Gray}{gray}{0.88}
\definecolor{LightCyan}{rgb}{0.88,1,1}
\newcommand{\colorrow}{\rowcolor{gray!20}}
\newcommand{\colorcell}{\cellcolor{gray!20}}

\newcommand{\name}{\texttt{GPTfluence}}

\title{On Training Data Influence of GPT Models}

 \author{Yekun Chai$^{\spadesuit}$\, Qingyi Liu\thanks{Work done during QL's internship at Baidu. 
 }$^\heartsuit$\, Shuohuan Wang$^\spadesuit$ \, \\ 
\textbf{Yu Sun$^\spadesuit$\, Qiwei Peng$^\diamondsuit$\, Hua Wu$^\spadesuit$} \\
  $^\spadesuit$Baidu Inc. \, 
  $^\heartsuit$Sun Yat-sen University \, 
  $^\diamondsuit$University of Copenhagen 
  \\
  \texttt{\{chaiyekun,wangshuohuan\}@baidu.com} \,
  \\   
  \texttt{\{liuqy95\}@mail2.sysu.edu.cn}  \\
} 

\begin{document}
\maketitle
\begin{abstract}
Amidst the rapid advancements in generative language models, the investigation of how training data shapes the performance of GPT models is still emerging. This paper presents {\name}, a novel approach that leverages a featurized simulation to assess the impact of training examples on the training dynamics of GPT models. Our approach not only traces the influence of individual training instances on performance trajectories, such as loss and other key metrics, on targeted test points but also enables a comprehensive comparison with existing methods across various training scenarios in GPT models, ranging from 14 million to 2.8 billion parameters, across a range of downstream tasks.
Contrary to earlier methods that struggle with generalization to new data, {\name} introduces a parameterized simulation of training dynamics, demonstrating robust generalization capabilities to unseen training data. This adaptability is evident across both fine-tuning and instruction-tuning scenarios, spanning tasks in natural language understanding and generation. We make our code and data publicly available at \url{https://github.com/ernie-research/gptfluence}.

\end{abstract}

\section{Introduction}

The advent of generative language models, particularly the GPT series~\cite{radford2019gpt2,brown2020gpt3,zhang2022opt}, has marked a paradigm shift in natural language processing (NLP)~\cite{touvron2023llama2,jiang2023mistral}, code generation~\cite{lozhkov2024starcoder,chai2023erniecode}, visual and language understanding~\cite{achiam2023gpt4,team2023gemini}. These models have redefined performance standards across an extensive range of tasks, igniting detailed investigations into the process of training dynamics and the intricate nature of learned representations. Despite these strides, the specific influence of individual training examples on the performance of GPT models remains a significantly underexplored area. This oversight presents a critical challenge in optimizing training processes, a challenge that grows in tandem with the increasing complexity and scale of these models.

Current research has \emph{yet} to focus comprehensively on the influence of training data on autoregressive language models. Prior studies, such as those utilizing the BERT~\cite{park2023trak} or T5 architecture~\cite{guu2023simfluence}, have predominantly concentrated on natural language understanding tasks, leaving a considerable void in the exploration of generative language models.

Furthermore, the majority of this research~\cite{pruthi2020tracin,guu2023simfluence,k2021revisiting,koh2017influence-function, yeh2018representer} has focused on test loss as the primary metric of interest, neglecting other vital performance indicators. Metrics such as BLEU~\cite{papineni2002bleu} and ROUGE~\cite{lin2004rouge} scores are crucial for a thorough evaluation of a model's capabilities, particularly in the context of generative language models where downstream task performance is paramount.
Additionally, the challenge of generalizability—extending methodologies to accommodate unseen data—persists as a significant barrier~\cite{guu2023simfluence}. This is particularly critical for models expected to adapt to the dynamic and evolving trajectory of NLP tasks.

In response to these gaps, we introduce {\name}, a novel framework designed to extend the analysis of training data influence beyond the limitations of existing methodologies and across a broader spectrum of tasks. Employing a featurized simulation approach, {\name} estimates the impact of individual training examples on the performance of GPT models, covering both natural language understanding and generation tasks. This expanded focus facilitates a comprehensive understanding of model training dynamics, providing insights into a wide array of evaluation metrics beyond mere test loss.

Extensive experiments on selected subsets from FLAN datasets~\cite{wei2022flan}, across a variety of tasks and GPT model variants~\cite{biderman2023pythia}, ranging in size from 14 million to 2.8 billion parameters, validate the effectiveness and superiority of our approach. Notably, our method not only sheds light on the training dynamics of GPT models but also demonstrates remarkable generalization capabilities to unseen data. 

\paragraph{Contribution}
To summarize, our contributions are as follows:
\begin{itemize}[noitemsep, left=0pt, labelsep=4pt,]
  \item We introduce {\name}, a featurized simulation approach that significantly advances the analysis of training data influence on GPT models. This approach not only enables a comprehensive comparison with existing methodologies but also marks the first extensive foray into the extensive investigation of training data's impact on the performance of GPT models across various scales.
  \item Our approach demonstrates effectiveness on GPT models across different scales, showing its generalization capability on unseen data. 
  \item We release the \textit{GPTDynamics} dataset, a collection encompassing over 350 runs of training dynamics data spanning six distinct model sizes and five NLP tasks, to facilitate further research advancement.
\end{itemize}


\section{Preliminaries}
\label{sec:pre}

In this section, we revisit the conceptual framework of training data attribution (TDA) methods, aiming to quantify the impact of individual training instances on the performance of models with respect to test data points. 

\subsection{Task Definition} 
\label{sec:definition}
Considering the data space $Z$, such as datasets utilized for instruction-tuning, we denote a training example by $z$ and a test example by $z'$ in $Z$. We employ a model, specifically a GPT variant in our experiments, parameterized by weights $\theta \in \mathbb{R}^p$. Our objective is to forecast the model's performance on a target metric $\phi(\theta, z): \mathbb{R}^p \times Z \rightarrow \mathbb{R}$, with a main focus in existing literature on predicting test set loss~\cite{pruthi2020tracin,guu2023simfluence}.

Practically, this involves working with a sequence of training batches $c = (c_1, c_2, \dots, c_T)$, delineating a training curriculum. Here, $c_t$ symbolizes the batch of training examples utilized at step $t$. The crux of our task is to ascertain the influence of training examples $z$ on a test example of interest $z'$, specifically in terms of a test metric score $\phi(\theta, z')$, given the training curriculum $c$. This involves tracking changes in performance trajectory as a function of the curriculum $c$, with prior research predominantly focused on test loss prediction, rather than a broader spectrum of performance metrics.

\subsection{Training Data Attribution}
\paragraph{TracIn} 
Inspired by the \textit{fundamental theorem of calculus}—which posits that the integral of a function's gradient over an interval equals the function's value difference across that interval—TracIn~\cite{pruthi2020tracin} employs the first-order Taylor expansion to quantify the data influence on test example loss at each step as follows:

\begin{equation}
    \mathcal{L}_{t+1} (z) \approx \mathcal{L}_t(z) - \eta_t \langle \nabla \mathcal{L}_t (z_i),  \nabla_\theta \mathcal{L}_t (z') \rangle
\end{equation}{
where $\eta_t$ represents the learning rate at step $t$, and $\nabla_\theta \mathcal{L}_t(\cdot)$ signifies the gradient of the loss function with respect to the model weights $\theta$. 
}

It adopts an influence measurement that utilizes checkpoint ensembling, dubbed \textit{TracInCP}. This approach aggregates the influences calculated at predefined intervals throughout the training, providing a comprehensive view of the training data's impact over time.

\begin{equation}
    \mathcal{I}_\text{TracIn} (z_i, z') = \sum_{i=1}^N \eta_i \nabla_\theta \mathcal{L}_t (z_i; \theta_i)^\top \nabla_\theta \mathcal{L}_t (z'; \theta_i)
\end{equation}{
  where $\mathcal{I}$ denotes the loss change \emph{w.r.t.} the training example $z$, and $N$ indicates the total number of model checkpoints saved during training.
}

\noindent\textbf{Simfluence}~\cite{guu2023simfluence} approaches the challenge by learning a linear function $f$ that correlates training samples $z$ with the test loss $\mathcal{L}(z';\theta)$, expressed as:

\begin{equation}
    \mathcal{L}_t (z) = \alpha (c_t) \mathcal{L}_{t-1}(z) + \beta (c_t)
\end{equation}{
Here, $\alpha(c_t)$ and $\beta(c_t)$, the multiplicative and additive factors respectively, are determined using a linear model, with $c_t$ indicating the batch of examples consumed at training step $t$. 
}
Although it offers a data-driven simulator derived from training dynamics trajectories, its mapping from training data indices to test data points constrains generalizability to new, unseen data.

While TracIn leverages the neural model's first-order gradients and Simfluence employs a data-driven simulation approach, both primarily focus on predicting test loss. Our proposed method aligns with Simfluence's direction but seeks to overcome its limitations, extending our focus to encompass a wider array of performance metrics beyond mere test loss prediction.


\section{{\name}: Featurized Simulation-based Approach}
\label{sec:method}

\begin{figure*}
    \centering
    \includegraphics[width=0.95\linewidth]{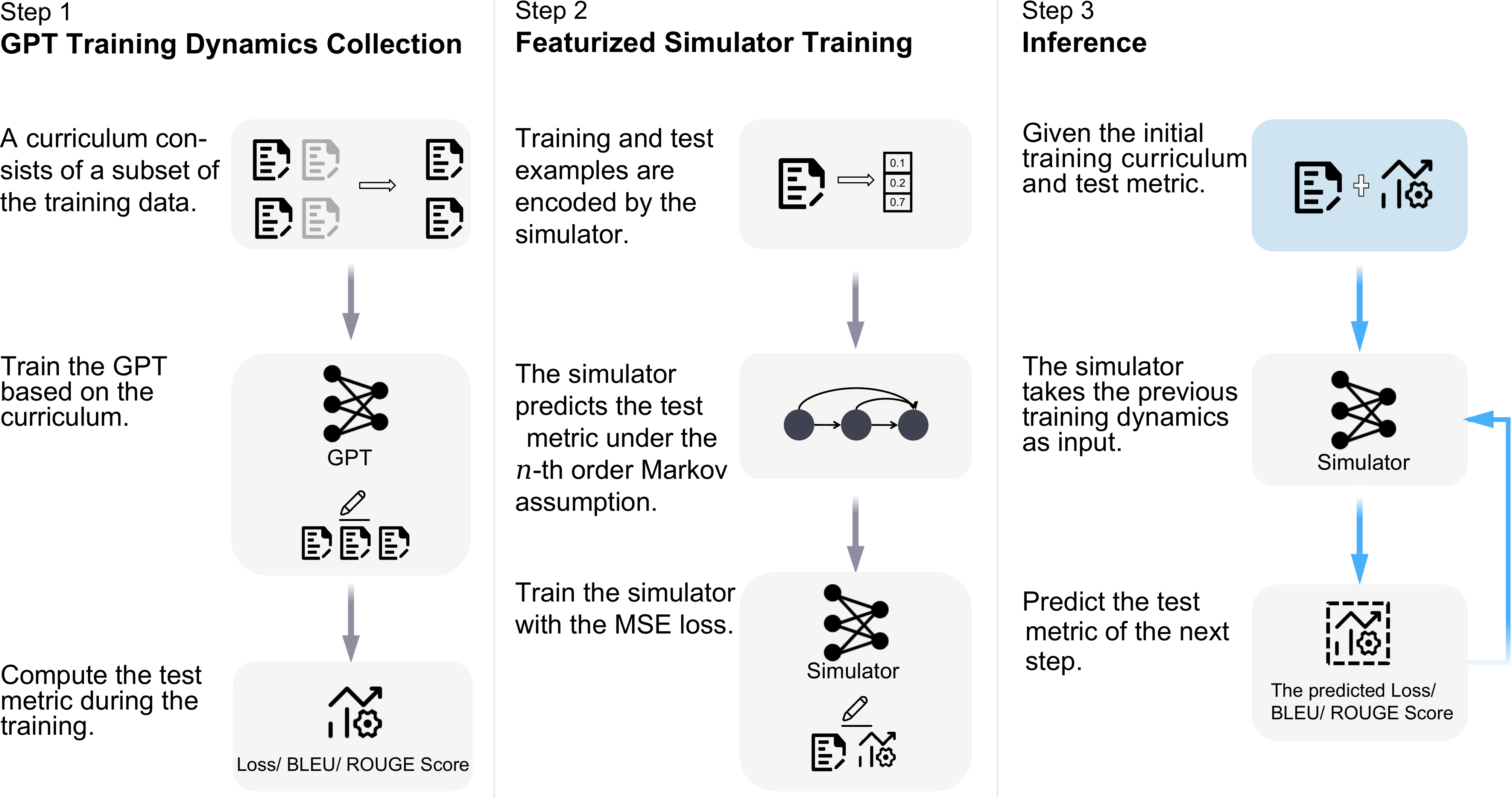}
    \caption{Overview of \name. \textbf{Step 1:} We sample training data to create curricula for training GPT models and compute the test metrics of test examples at each training step. All the training curricula and the ground-truth metrics are referred to as \textit{GPTDynamics}. \textbf{Step 2:} We train our \textit{featurized} simulator on \textit{GPTDynamics}, taking into account training examples at current and previous steps with the test example as input and predicts the ground-truth metric. \textbf{Step 3:} Given a new curriculum with the test example of interest, start from the test metric at the first step, the simulator simulates the test metric in the future training steps in an autoregressive manner.}
    \label{fig:overview}
\end{figure*}

\subsection{Overview}
We present {\name}, a novel approach for tracking the impact of training examples on the training dynamics of GPT models using a featurized simulator. Figure~\ref{fig:overview} depicts the process of {\name}, encompassing the collection of training dynamics, the training of the simulator, and the execution of the final simulation. Similar to \citet{guu2023simfluence}, our initial step involves gathering a comprehensive dataset of training dynamics, which captures both the training curriculum and various target metrics for test examples, extending beyond traditional loss metrics to include performance measures like BLEU and ROUGE scores.

{\name} models these dynamics via an $n$-th order Markov process, incorporating both multiplicative and additive factors to reflect the influence of training examples. At its core, the simulator uses a pre-trained encoder to attain the general representation of training and test examples, ensuring adaptability to new, unseen data. This is achieved by modeling the intricate interplay between examples through the interactions within their condensed hidden vector representations. In its application, it can autoregressively forecast the complete performance trajectory of a test example, starting from its initial performance metrics and following the specified training curriculum.


The collection of training dynamics is pivotal for predicting a test sample's performance trajectory throughout the training process. As outlined in \S\ref{sec:definition}, a $T$ time steps training run is characterized by a sequence of training batches $c$, each contributing to the model's evolving parameters, $\theta_t$, through gradient descent. 

To monitor the performance evolution of a particular test example $z'$, we record its metric scores $y_t = \phi(\theta_{t}, z')$ at every training step $t$, employing a variety of evaluation metrics beyond mere loss, such as BLEU and ROUGE. This comprehensive record, denoted as $y = {\phi_{1:T}}$, tracks the test example's performance across all $T$ steps of training.

From a broader dataset $\mathcal{D}$, we sample $K$ subsets $\mathcal{D}' \subset \mathcal{D}$ for GPT model training, resulting in $K$ distinct training runs. These runs yield a rich dataset of training dynamics $\mathcal{D}_{run}$, encapsulating both the training curricula and the sequential target metric scores $\phi$ for each test point $z'$. This dataset is represented as $\mathcal{D}_{run} = \{c^k, y^k\}_{k=1}^{K}$.


\subsection{Featurized Simulation Approach}
\label{sec:featurized}

In this work, we introduce a featurized simulation methodology designed to capture the effects of training examples on GPT model training dynamics. This method is predicated on conceptualizing the training process as a sequential, time-evolving Markov process, thereby enabling the simulation of metric trajectories across training iterations. Building upon the foundational insights of~\citet{guu2023simfluence}, our model extends the conventional first-order Markov assumption to an \emph{$n$-th order Markov process}. This allows for the consideration of a test sample $z'$, where its performance metric $\phi(\cdot)$ at any given timestep $t$ is influenced by its performance across the preceding $n$ steps, encapsulated as $\{\phi_{t-1}, \phi_{t-2}, \cdots, \phi_{t-n}\}$.

Our approach integrates both multiplicative and additive components within the simulation. The performance trajectory of a test sample $z'$ is thus delineated by a combination of these factors, formulated as follows:

\begin{equation} \label{eq:n-markov}
\phi_{t}(z') = \sum_{j=1}^{n} \alpha_{j}(c_t) \phi_{t-j}(z') + \beta(c_t)
\end{equation}
{where $\alpha_{1:n}(\cdot)$ and $\beta(\cdot)$ represent the learned functions attributed to the current training batch $c_t$. Here, $\alpha_{j}(c_t)$ and $\beta(c_t)$ are determined through the aggregation of influence factors $A_{i,j}$ and $B_i$, respectively, across the training examples in $c_t$:}

\begin{equation}
\alpha_{j}(c_t) = \sum_{i \in c_t} A_{i,j}, \quad \beta(c_t) = \sum_{i \in c_t} B_i
\end{equation}

We introduce a parameterized, featurized simulator that employs a pre-trained encoder $\Psi(\cdot)$ such as BERT~\cite{devlin-etal-2019-bert} and GPT~\cite{radford2019gpt2}. This is adept at processing each training example $z_i$ and test example $z'$, generating predictive influence factors $A_{i,j}$ and $B_i$ through the encoded representations $h_{z_i}$ and $h_{z'}$:

\begin{equation}
h^{z_i} = \Psi(z_i), \quad h^{z'} = \Psi(z'),
\end{equation}{where $h^{z_i}$ and $h^{z'}$ are the low-dimensional embeddings of the training and test examples, respectively. To preserve the encoder's semantic generalizability, we keep it frozen during the simulator's training.} 

 The multiplicative and additive influence factors are then derived by passing the embeddings through the corresponding linear projections, which are subsequently integrated using a Frobenius product as follows:

\begin{align}
A_{i,j} &{}= \langle \mathbf{W}_{(j)}^\top h^{z_i}_{j}, \mathbf{U}_{(j)}^\top h^{z'} \rangle_F \\
B_{i} &{}= \langle \mathbf{W'}^\top h^{z_i}_{j}, \mathbf{U'}^\top h^{z'} \rangle_F
\end{align}

where $\mathbf{W_{(j)}}, \mathbf{U_{(j)}}, \mathbf{W'}, \mathbf{U'}$ are learnable weights, $\langle \cdot, \cdot \rangle_F$ represents the Frobenius inner product between the hidden representations of the training and test examples, yielding a refined estimation of the multiplicative influence exerted by each training example $z_i$ on the test example's performance trajectory. Our approach offers a granular and comprehensive analysis of training dynamics through this intricate data-driven simulation.

To learn our featurized simulator $\Theta$, we optimize the following L2-regularized regression objective:
\begin{equation}
    \Theta^{\star} = \underset{\Theta}{\mathrm{argmin}}\sum_{t \in T}(y_t - \hat{\phi}_{t}(z') )^2 + \lambda(||\Theta||_2^2) 
\end{equation}{where $\lambda$ is the discounting factor dictating the degree of L2-regularization, $\hat{\phi}_t(\cdot)$ is the test score prediction at step $t$ using Eq.\eqref{eq:n-markov}.} Refer to Algorithm~\ref{algo:gptfluence} for the pseudo-code.




\begin{table*}[!ht]
\centering
\resizebox{\linewidth}{!}{
\begin{tabular}{llp{2cm}p{2cm}p{2.6cm}p{2cm}p{2cm}p{2.6cm}p{2cm}p{2cm}p{2.6cm}}
\hline
\multirow{2}{*}{\textbf{Method}} & \multirow{2}{*}{\textbf{\#Param}} & \multicolumn{3}{c}{\textbf{RTE}} & \multicolumn{3}{c}{\textbf{SST-2}} & \multicolumn{3}{c}{\textbf{BoolQ}} \\ \cline{3-11} 
 &  & All-Steps MSE ($\downarrow$) & All-Steps MAE ($\downarrow$) & \multicolumn{1}{p{2.6cm}|}{Final-Step Spearman’s $\rho$ 
 ($\uparrow$)} & All-Steps MSE ($\downarrow$) & All-Steps MAE ($\downarrow$) & \multicolumn{1}{p{2.6cm}|}{Final-Step Spearman’s $\rho$ 
 ($\uparrow$)} & All-Steps MSE ($\downarrow$) & All-Steps MAE ($\downarrow$) & Final-Step Spearman’s $\rho$ 
 ($\uparrow$) \\ \hline
TracIn-CP (10-steps) & \multirow{5}{*}{410M} & 1.156{\fontsize{9}{\baselineskip}\selectfont(0.838)} & 0.787{\fontsize{9}{\baselineskip}\selectfont(0.339)} & \multicolumn{1}{p{2.6cm}|}{0.460} & 0.551{\fontsize{9}{\baselineskip}\selectfont(0.560)} & 0.584{\fontsize{9}{\baselineskip}\selectfont(0.307)} & \multicolumn{1}{p{2.6cm}|}{-0.089} & 0.957{\fontsize{9}{\baselineskip}\selectfont(0.728)} & 0.735{\fontsize{9}{\baselineskip}\selectfont(0.332)} & -0.066 \\
TracIn-CP (all-steps) & & 0.757{\fontsize{9}{\baselineskip}\selectfont(0.591)} & 0.629{\fontsize{9}{\baselineskip}\selectfont(0.299)} & \multicolumn{1}{p{2.6cm}|}{0.460} & 0.446{\fontsize{9}{\baselineskip}\selectfont(0.555)} & 0.525{\fontsize{9}{\baselineskip}\selectfont(0.321)} & \multicolumn{1}{p{2.6cm}|}{-0.089} & 0.782{\fontsize{9}{\baselineskip}\selectfont(0.690)} & 0.680{\fontsize{9}{\baselineskip}\selectfont(0.339)} & -0.066 \\
 Grad-Dot & & 12.061{\fontsize{9}{\baselineskip}\selectfont(3.688)} & 2.906{\fontsize{9}{\baselineskip}\selectfont(0.410)} & \multicolumn{1}{p{2.6cm}|}{0.459} & 7.715{\fontsize{9}{\baselineskip}\selectfont(1.543)} & 1.918{\fontsize{9}{\baselineskip}\selectfont(0.205)} & \multicolumn{1}{p{2.6cm}|}{-0.084} & 12.527{\fontsize{9}{\baselineskip}\selectfont(3.617)} & 2.900{\fontsize{9}{\baselineskip}\selectfont(0.344)} & -0.071 \\
  Simfluence & & 1.477{\fontsize{9}{\baselineskip}\selectfont(0.274)} & 0.634{\fontsize{9}{\baselineskip}\selectfont(0.111)} & \multicolumn{1}{p{2.6cm}|}{0.426{\fontsize{9}{\baselineskip}\selectfont(0.340)}} & 1.133{\fontsize{9}{\baselineskip}\selectfont(0.287)} & 0.455{\fontsize{9}{\baselineskip}\selectfont(0.082)} & \multicolumn{1}{p{2.6cm}|}{0.696{\fontsize{9}{\baselineskip}\selectfont(0.156)}} & 1.189{\fontsize{9}{\baselineskip}\selectfont(0.362)} & 0.485{\fontsize{9}{\baselineskip}\selectfont(0.082)} & 0.793{\fontsize{9}{\baselineskip}\selectfont(0.201)} \\\colorrow
 Ours & & \textbf{0.220{\fontsize{9}{\baselineskip}\selectfont(0.184)}} & \textbf{0.334{\fontsize{9}{\baselineskip}\selectfont(0.140)}} & \multicolumn{1}{p{2.6cm}|}{\textbf{0.644{\fontsize{9}{\baselineskip}\selectfont(0.174)}}} & \textbf{0.111{\fontsize{9}{\baselineskip}\selectfont(0.045)}} & \textbf{0.224{\fontsize{9}{\baselineskip}\selectfont(0.047)}} & \multicolumn{1}{p{2.6cm}|}{\textbf{0.834{\fontsize{9}{\baselineskip}\selectfont(0.129)}}} & \textbf{0.132{\fontsize{9}{\baselineskip}\selectfont(0.073)}} & \textbf{0.251{\fontsize{9}{\baselineskip}\selectfont(0.075)}} & \textbf{0.828{\fontsize{9}{\baselineskip}\selectfont(0.154)}} \\ \hline
 TracIn-CP (10-steps) & \multirow{5}{*}{1B} & 1.225{\fontsize{9}{\baselineskip}\selectfont(0.744)} & 0.979{\fontsize{9}{\baselineskip}\selectfont(0.344)} & \multicolumn{1}{p{2.6cm}|}{-0.203} & 4.412{\fontsize{9}{\baselineskip}\selectfont(1.301)} & 1.697{\fontsize{9}{\baselineskip}\selectfont(0.170)} & \multicolumn{1}{p{2.6cm}|}{-0.058} & 0.999{\fontsize{9}{\baselineskip}\selectfont(1.034)} & 0.793{\fontsize{9}{\baselineskip}\selectfont(0.400)} & 0.649 \\
 TracIn-CP (all-steps) &  & 1.137{\fontsize{9}{\baselineskip}\selectfont(0.740)} & 0.939{\fontsize{9}{\baselineskip}\selectfont(0.343)} & \multicolumn{1}{p{2.6cm}|}{-0.203} & 2.158{\fontsize{9}{\baselineskip}\selectfont(0.782)} & 1.218{\fontsize{9}{\baselineskip}\selectfont(0.187)} & \multicolumn{1}{p{2.6cm}|}{-0.058} & 0.858{\fontsize{9}{\baselineskip}\selectfont(1.043)} & 0.731{\fontsize{9}{\baselineskip}\selectfont(0.416)} & 0.649 \\
 Grad-Dot & & 21.928{\fontsize{9}{\baselineskip}\selectfont(7.871)} & 4.332 {\fontsize{9}{\baselineskip}\selectfont(0.874)} & \multicolumn{1}{p{2.6cm}|}{-0.198} & 6.601{\fontsize{9}{\baselineskip}\selectfont(1.927)} & 2.077{\fontsize{9}{\baselineskip}\selectfont(0.193)} & \multicolumn{1}{p{2.6cm}|}{-0.057} & 18.270{\fontsize{9}{\baselineskip}\selectfont(5.630)} & 3.563{\fontsize{9}{\baselineskip}\selectfont(0.711)} & 0.650 \\
  Simfluence & & 0.889{\fontsize{9}{\baselineskip}\selectfont(0.551)} & 0.523{\fontsize{9}{\baselineskip}\selectfont(0.197)} & \multicolumn{1}{p{2.6cm}|}{0.360{\fontsize{9}{\baselineskip}\selectfont(0.207)}} & 0.582{\fontsize{9}{\baselineskip}\selectfont(0.253)} & 0.410{\fontsize{9}{\baselineskip}\selectfont(0.084)} & \multicolumn{1}{p{2.6cm}|}{0.712{\fontsize{9}{\baselineskip}\selectfont(0.148)}} & 0.876{\fontsize{9}{\baselineskip}\selectfont(0.470)} & 0.469{\fontsize{9}{\baselineskip}\selectfont(0.198)} & 0.862{\fontsize{9}{\baselineskip}\selectfont(0.050)} \\ \colorrow
 Ours & & \textbf{0.099{\fontsize{9}{\baselineskip}\selectfont(0.078)}} & \textbf{0.227{\fontsize{9}{\baselineskip}\selectfont(0.097)}} & \multicolumn{1}{p{2.6cm}|}{\textbf{0.757{\fontsize{9}{\baselineskip}\selectfont(0.123)}}} & \textbf{0.096{\fontsize{9}{\baselineskip}\selectfont(0.075)}} & \textbf{0.221{\fontsize{9}{\baselineskip}\selectfont(0.084)}} & \multicolumn{1}{p{2.6cm}|}{\textbf{0.807{\fontsize{9}{\baselineskip}\selectfont(0.175)}}} & \textbf{0.068{\fontsize{9}{\baselineskip}\selectfont(0.058)}} & \textbf{0.187{\fontsize{9}{\baselineskip}\selectfont(0.070)}} & \textbf{0.953{\fontsize{9}{\baselineskip}\selectfont(0.034)}} \\ \hline
 
TracInCP (10-steps) & \multirow{5}{*}{2.8B} & 8.869{\fontsize{9}{\baselineskip}\selectfont(3.673)} & 2.700{\fontsize{9}{\baselineskip}\selectfont(0.650)} & \multicolumn{1}{p{2.6cm}|}{0.573} & 0.294{\fontsize{9}{\baselineskip}\selectfont(0.235)} & 0.447{\fontsize{9}{\baselineskip}\selectfont(0.176)} & \multicolumn{1}{p{2.6cm}|}{0.801} & 1.185{\fontsize{9}{\baselineskip}\selectfont(1.271)} & 0.804{\fontsize{9}{\baselineskip}\selectfont(0.436)} & 0.184 \\
TracInCP (all-steps) &  & 10.256{\fontsize{9}{\baselineskip}\selectfont(4.396)} & 2.967{\fontsize{9}{\baselineskip}\selectfont(0.652)} & \multicolumn{1}{p{2.6cm}|}{0.573} & 0.265{\fontsize{9}{\baselineskip}\selectfont(0.228)} & 0.419{\fontsize{9}{\baselineskip}\selectfont(0.178)} & \multicolumn{1}{p{2.6cm}|}{0.801} & 1.183{\fontsize{9}{\baselineskip}\selectfont(1.260)} & 0.800{\fontsize{9}{\baselineskip}\selectfont(0.434)} & 0.184 \\
Grad-Dot &  & 10.101{\fontsize{9}{\baselineskip}\selectfont(9.212)} & 2.580{\fontsize{9}{\baselineskip}\selectfont(1.327)} & \multicolumn{1}{p{2.6cm}|}{0.573} & 1.216{\fontsize{9}{\baselineskip}\selectfont(0.411)} & 0.935{\fontsize{9}{\baselineskip}\selectfont(0.175)} & \multicolumn{1}{p{2.6cm}|}{-0.801} & 1.990{\fontsize{9}{\baselineskip}\selectfont(1.082)} & 1.219{\fontsize{9}{\baselineskip}\selectfont(0.321)} & 0.184 \\
Simfluence-linear &  & 2.032{\fontsize{9}{\baselineskip}\selectfont(1.214)} & 0.996{\fontsize{9}{\baselineskip}\selectfont(0.360)} & \multicolumn{1}{p{2.6cm}|}{0.845{\fontsize{9}{\baselineskip}\selectfont(0.061)}} & 0.921{\fontsize{9}{\baselineskip}\selectfont(0.435)} & 0.634{\fontsize{9}{\baselineskip}\selectfont(0.194)} & \multicolumn{1}{p{2.6cm}|}{0.912{\fontsize{9}{\baselineskip}\selectfont(0.018)}} & 1.545{\fontsize{9}{\baselineskip}\selectfont(1.293)} & 0.849{\fontsize{9}{\baselineskip}\selectfont(0.412)} & 0.681{\fontsize{9}{\baselineskip}\selectfont(0.087)} \\ \colorrow
Ours &  & \textbf{0.132{\fontsize{9}{\baselineskip}\selectfont(0.172)}} & \textbf{0.273{\fontsize{9}{\baselineskip}\selectfont(0.129)}} & \multicolumn{1}{p{2.6cm}|}{\textbf{0.969{\fontsize{9}{\baselineskip}\selectfont(0.009)}}} & \textbf{0.023{\fontsize{9}{\baselineskip}\selectfont(0.015)}} & \textbf{0.123{\fontsize{9}{\baselineskip}\selectfont(0.040)}} & \multicolumn{1}{p{2.6cm}|}{\textbf{0.979{\fontsize{9}{\baselineskip}\selectfont(0.006)}}} & \textbf{0.175{\fontsize{9}{\baselineskip}\selectfont(0.232)}} & \textbf{0.305{\fontsize{9}{\baselineskip}\selectfont(0.165)}} & \textbf{0.963{\fontsize{9}{\baselineskip}\selectfont(0.018)}} \\ \hline \hline

 \multirow{2}{*}{\textbf{Method}} & \multirow{2}{*}{\textbf{\#Param}} & \multicolumn{3}{c}{\textbf{WebNLG}} & \multicolumn{3}{c}{\textbf{WMT-16 DE/EN}} & \multicolumn{3}{c}{ \color{brown} \textbf{Average}} \\ \cline{3-11} 
 &  & All-Steps MSE ($\downarrow$) & All-Steps MAE ($\downarrow$) & Final-Step Spearman’s $\rho$ 
 ($\uparrow$) & All-Steps MSE ($\downarrow$) & All-Steps MAE ($\downarrow$) & Final-Step Spearman’s $\rho$ 
 ($\uparrow$) & All-Steps MSE ($\downarrow$) & All-Steps MAE ($\downarrow$) & Final-Step Spearman’s $\rho$ 
 ($\uparrow$) \\ \hline
TracIn-CP (10-steps) & \multirow{5}{*}{410M} & 0.048{\fontsize{9}{\baselineskip}\selectfont(0.072)} & 0.168{\fontsize{9}{\baselineskip}\selectfont(0.115)} & \multicolumn{1}{p{2.6cm}|}{0.836} & 0.030{\fontsize{9}{\baselineskip}\selectfont(0.071)} & 0.122{\fontsize{9}{\baselineskip}\selectfont(0.107)} & \multicolumn{1}{p{2.6cm}|}{0.963} & 0.548 & 0.479 & 0.421 \\
 TracIn-CP (all-steps) & & 0.050{\fontsize{9}{\baselineskip}\selectfont(0.073)} & 0.173{\fontsize{9}{\baselineskip}\selectfont(0.113)} & \multicolumn{1}{p{2.6cm}|}{0.836} & 0.030{\fontsize{9}{\baselineskip}\selectfont(0.071)} & 0.123{\fontsize{9}{\baselineskip}\selectfont(0.107)} & \multicolumn{1}{p{2.6cm}|}{0.963} & 0.413 & 0.426 & 0.421 \\
 Grad-Dot & & 0.062{\fontsize{9}{\baselineskip}\selectfont(0.080)} & 0.187{\fontsize{9}{\baselineskip}\selectfont(0.113)} &\multicolumn{1}{p{2.6cm}|}{0.837} & 0.033{\fontsize{9}{\baselineskip}\selectfont(0.073)} & 0.127{\fontsize{9}{\baselineskip}\selectfont(0.109)} & \multicolumn{1}{p{2.6cm}|}{0.963} & 6.479 & 1.608 & 0.421 \\
  Simfluence & & 0.036{\fontsize{9}{\baselineskip}\selectfont(0.029)} & 0.130{\fontsize{9}{\baselineskip}\selectfont(0.049)} & \multicolumn{1}{p{2.6cm}|}{0.986{\fontsize{9}{\baselineskip}\selectfont(0.002)}} & 0.016{\fontsize{9}{\baselineskip}\selectfont(0.013)} & 0.101{\fontsize{9}{\baselineskip}\selectfont(0.034)} & \multicolumn{1}{p{2.6cm}|}{0.997{\fontsize{9}{\baselineskip}\selectfont(0.001)}} & 0.770 & 0.361 & 0.779 \\ \colorrow
 Ours & & \textbf{0.002{\fontsize{9}{\baselineskip}\selectfont(0.002)}} & \textbf{0.033{\fontsize{9}{\baselineskip}\selectfont(0.017)}} & \multicolumn{1}{p{2.6cm}|}{\textbf{0.994{\fontsize{9}{\baselineskip}\selectfont(0.001)}}} & \textbf{0.002{\fontsize{9}{\baselineskip}\selectfont(0.004)}} & \textbf{0.033{\fontsize{9}{\baselineskip}\selectfont(0.023)}} & \multicolumn{1}{p{2.6cm}|}{\textbf{0.998{\fontsize{9}{\baselineskip}\selectfont(0.000)}}} & \textbf{0.093} & \textbf{0.175} & \textbf{0.860} \\ \hline
TracIn-CP (10-steps) & \multirow{5}{*}{1B} & 0.032{\fontsize{9}{\baselineskip}\selectfont(0.053)} & 0.132{\fontsize{9}{\baselineskip}\selectfont(0.095)} & \multicolumn{1}{p{2.6cm}|}{0.885} & 0.012{\fontsize{9}{\baselineskip}\selectfont(0.032)} & 0.075{\fontsize{9}{\baselineskip}\selectfont(0.069)} & \multicolumn{1}{p{2.6cm}|}{0.981} & 1.336 & 0.735 & 0.451 \\
 TracIn-CP (all-steps) & & 0.033{\fontsize{9}{\baselineskip}\selectfont(0.053)} & 0.135{\fontsize{9}{\baselineskip}\selectfont(0.094)} & \multicolumn{1}{p{2.6cm}|}{0.885} & 0.012{\fontsize{9}{\baselineskip}\selectfont(0.032)} & 0.076{\fontsize{9}{\baselineskip}\selectfont(0.069)} & \multicolumn{1}{p{2.6cm}|}{0.981} & 0.840 & 0.620 & 0.451 \\
 Grad-Dot & & 0.044{\fontsize{9}{\baselineskip}\selectfont(0.061)} & 0.154{\fontsize{9}{\baselineskip}\selectfont(0.097)} & \multicolumn{1}{p{2.6cm}|}{0.881} & 0.013{\fontsize{9}{\baselineskip}\selectfont(0.033)} & 0.075{\fontsize{9}{\baselineskip}\selectfont(0.071)} & \multicolumn{1}{p{2.6cm}|}{0.981} & 9.371 & 2.040 & 0.451 \\
  Simfluence & & 0.167{\fontsize{9}{\baselineskip}\selectfont(0.127)} & 0.323{\fontsize{9}{\baselineskip}\selectfont(0.112)} & \multicolumn{1}{p{2.6cm}|}{0.823{\fontsize{9}{\baselineskip}\selectfont(0.030)}} & 0.171{\fontsize{9}{\baselineskip}\selectfont(0.269)} & 0.309{\fontsize{9}{\baselineskip}\selectfont(0.168)} & \multicolumn{1}{p{2.6cm}|}{0.925{\fontsize{9}{\baselineskip}\selectfont(0.007)}} & 0.537 & 0.407 & 0.737 \\ 
 \colorrow Ours & & \textbf{0.007{\fontsize{9}{\baselineskip}\selectfont(0.005)}} & \textbf{0.068{\fontsize{9}{\baselineskip}\selectfont(0.022)}} & \multicolumn{1}{p{2.6cm}|}{\textbf{0.984{\fontsize{9}{\baselineskip}\selectfont(0.005)}}} & \textbf{0.004{\fontsize{9}{\baselineskip}\selectfont(0.004)}} & \textbf{0.049{\fontsize{9}{\baselineskip}\selectfont(0.020)}} & \multicolumn{1}{p{2.6cm}|}{\textbf{0.997{\fontsize{9}{\baselineskip}\selectfont(0.001)}}} & \textbf{0.055} & \textbf{0.150} & \textbf{0.900} \\ \hline

 TracInCP (10-steps) & \multirow{5}{*}{2.8B} & 0.005{\fontsize{9}{\baselineskip}\selectfont(0.008)} & 0.051{\fontsize{9}{\baselineskip}\selectfont(0.035)} & \multicolumn{1}{p{2.6cm}|}{0.978} & \textbf{0.001{\fontsize{9}{\baselineskip}\selectfont(0.002)}} & \textbf{0.020{\fontsize{9}{\baselineskip}\selectfont(0.019)}} & \multicolumn{1}{p{2.6cm}|}{0.997} & 2.071 & 0.804 & 0.707 \\
TracInCP (all-steps) &  & 0.005{\fontsize{9}{\baselineskip}\selectfont(0.008)} & 0.051{\fontsize{9}{\baselineskip}\selectfont(0.035)} & \multicolumn{1}{p{2.6cm}|}{0.978} & \textbf{0.001{\fontsize{9}{\baselineskip}\selectfont(0.002)}} & \textbf{0.020{\fontsize{9}{\baselineskip}\selectfont(0.019)}} & \multicolumn{1}{p{2.6cm}|}{0.997} & 2.342 & 0.851 & 0.707 \\
Grad-Dot &  & 0.015{\fontsize{9}{\baselineskip}\selectfont(0.020)} & 0.089{\fontsize{9}{\baselineskip}\selectfont(0.061)} & \multicolumn{1}{p{2.6cm}|}{0.978} & \textbf{0.001{\fontsize{9}{\baselineskip}\selectfont(0.002)}} & 0.021{\fontsize{9}{\baselineskip}\selectfont(0.019)} & \multicolumn{1}{p{2.6cm}|}{0.997} & 2.665 & 0.969 & 0.386 \\
Simfluence-linear &  & 0.102{\fontsize{9}{\baselineskip}\selectfont(0.065)} & 0.283{\fontsize{9}{\baselineskip}\selectfont(0.091)} & \multicolumn{1}{p{2.6cm}|}{0.971{\fontsize{9}{\baselineskip}\selectfont(0.004)}} & 0.063{\fontsize{9}{\baselineskip}\selectfont(0.085)} & 0.203{\fontsize{9}{\baselineskip}\selectfont(0.119)} & \multicolumn{1}{p{2.6cm}|}{0.991{\fontsize{9}{\baselineskip}\selectfont(0.001)}} & 0.933 & 0.593 & 0.880 \\ \colorrow
Ours &  & \textbf{0.001{\fontsize{9}{\baselineskip}\selectfont(0.001)}} & \textbf{0.024{\fontsize{9}{\baselineskip}\selectfont(0.016)}} & \multicolumn{1}{p{2.6cm}|}{\textbf{0.997{\fontsize{9}{\baselineskip}\selectfont(0.000)}}} & \textbf{0.001{\fontsize{9}{\baselineskip}\selectfont(0.002)}} & \textbf{0.020{\fontsize{9}{\baselineskip}\selectfont(0.016)}} & \multicolumn{1}{p{2.6cm}|}{\textbf{0.999{\fontsize{9}{\baselineskip}\selectfont(0.000)}}} & \textbf{0.066} & \textbf{0.149} & \textbf{0.981} \\ \hline
\end{tabular}
}
\caption{Results of test loss estimation for \textit{instruction tuning}. Results are averaged over 5 held-out test runs. 
}
\vskip -2mm
\label{tab:main}
\end{table*}

\subsection{Connection to Previous Approaches}
Our approach offers a flexible framework that, under specific conditions, aligns with established models in the TDA literature. Specifically, when the focus narrows down to the overall influence of per-step dynamics, our approach converges to the datamodels~\cite{ilyas2022datamodels, engstrom2024dsdm}. Moreover, in scenarios where the Markov order $n$ is set to 1 and the input encoder is configured to process sample indices, our method reduces to Simfluence~\cite{guu2023simfluence}. 


\section{Experiments}
\label{sec:exp}
\subsection{Experimental Settings}

\subsubsection{\textit{GPTDynamics} Data Collection }

\vspace{0.1em}\noindent \textbf{Datasets and GPT Training Scenarios} \quad
In subsequent experiments, we refer to the comprehensive training process that employs the aggregated FLAN datasets along with task-specific instructions as \textit{instruction tuning}. Conversely, the term \textit{fine-tuning} is reserved to describe the process of individually optimizing models on separate tasks without the use of instructional prompts. Both instruction tuning and fine-tuning processes are encapsulated within our \textit{GPTDynamics} dataset. We refer to Appendix~\S\ref{ap:dataset} for detailed information.


\vspace{0.1em}\noindent \textbf{GPT Backbone} \quad We employed Pythia~\cite{biderman2023pythia}, a model suite recently made available to the public, as our foundational architecture. Within this suite, we selected five distinct models based on their sizes, encompassing 14M, 70M, 160M, 410M, 1B, and 2.8B, to ensure a broad range of computational capacities were represented.


\subsubsection{Experiment Setup for Simulators}
\vspace{0.1em}\noindent \textbf{Baselines} \quad We select TracIn~\cite{pruthi2020tracin}, Grad-Dot~\cite{charpiat2019input}, and Simfluence~\cite{guu2023simfluence} as our baselines. Refer to Appendix \S\ref{ap:baseline} for detailed information.


\vspace{0.1em}\noindent \textbf{Evaluation Metrics} \quad
We utilize a comprehensive set of metrics, including the Mean Squared Error (MSE) and Mean Absolute Error (MAE) calculated across all training steps, alongside the Spearman correlation coefficient ($\rho$) at the final step, to thoroughly assess performance.


\subsection{Test Loss Estimation}
\label{sub:setting}
\vspace{0.1em}\noindent \textbf{Instruction Tuning} \quad
Table~\ref{tab:main} presents a comparison between our approach and traditional TDA methods for instruction tuning. {\name} demonstrated a distinct edge over Simfluence and other gradient-based TDA techniques across a set of five natural language understanding (NLU) and natural language generation (NLG) tasks, as evidenced by the MSE and MAE metrics for the entire trajectory, alongside the Spearman correlation coefficients at the final time step across various test samples. Examples are shown in Fig.~\ref{fig:loss_a} and~\ref{fig:loss_b}.  Additionally, we observed that while the effectiveness of all evaluated TDA methods in predicting loss trajectories varied with changes in GPT sizes, {\name} maintained optimal performance, independent of the GPT scale.

\begin{table}[]
 \vspace{-2mm}
 \centering
\resizebox{\linewidth}{!}{
\large
\begin{tabular}{llp{2cm}p{2cm}p{2.9cm}}
\hline
\textbf{Dataset} & \textbf{Method} & All-Steps MSE ($\downarrow$) & All-Steps MAE ($\downarrow$) & Final-Step Spearman’s $\rho$ 
 ($\uparrow$) \\ \hline
RTE &  Simfluence & \textbf{0.035{\fontsize{9}{\baselineskip}\selectfont(0.022)}} & \textbf{0.151{\fontsize{9}{\baselineskip}\selectfont(0.054)}} & 0.743{\fontsize{9}{\baselineskip}\selectfont(0.094)} \\
 & Ours & 0.036{\fontsize{9}{\baselineskip}\selectfont(0.029)} & \textbf{0.151{\fontsize{9}{\baselineskip}\selectfont(0.060)}} & \textbf{0.746{\fontsize{9}{\baselineskip}\selectfont(0.095)}} \\ \hline
SST-2 &  Simfluence & 0.037{\fontsize{9}{\baselineskip}\selectfont(0.017)} & 0.128{\fontsize{9}{\baselineskip}\selectfont(0.030)} & 0.938{\fontsize{9}{\baselineskip}\selectfont(0.074)} \\
 & Ours & \textbf{0.014{\fontsize{9}{\baselineskip}\selectfont(0.006)}} & \textbf{0.081{\fontsize{9}{\baselineskip}\selectfont(0.018)}} & \textbf{0.943{\fontsize{9}{\baselineskip}\selectfont(0.073)}} \\ \hline
BoolQ &  Simfluence & 0.032{\fontsize{9}{\baselineskip}\selectfont(0.019)} & 0.140{\fontsize{9}{\baselineskip}\selectfont(0.038)} & 0.992{\fontsize{9}{\baselineskip}\selectfont(0.002)} \\
 & Ours & \textbf{0.011{\fontsize{9}{\baselineskip}\selectfont(0.011)}} & \textbf{0.082{\fontsize{9}{\baselineskip}\selectfont(0.049)}} & \textbf{0.994{\fontsize{9}{\baselineskip}\selectfont(0.002)}} \\ \hline
WebNLG &  Simfluence & 0.016{\fontsize{9}{\baselineskip}\selectfont(0.012)} & 0.094{\fontsize{9}{\baselineskip}\selectfont(0.036)} & 0.984{\fontsize{9}{\baselineskip}\selectfont(0.002)} \\
 & Ours & \textbf{0.011{\fontsize{9}{\baselineskip}\selectfont(0.014)}} & \textbf{0.078{\fontsize{9}{\baselineskip}\selectfont(0.043)}} & \textbf{0.985{\fontsize{9}{\baselineskip}\selectfont(0.002)}} \\ \hline
WMT-16 &  Simfluence & 0.010{\fontsize{9}{\baselineskip}\selectfont(0.008)} & 0.067{\fontsize{9}{\baselineskip}\selectfont(0.029)} & 0.998{\fontsize{9}{\baselineskip}\selectfont(0.003)} \\
\,\,DE/EN & Ours & \textbf{0.002{\fontsize{9}{\baselineskip}\selectfont(0.002)}} & \textbf{0.031{\fontsize{9}{\baselineskip}\selectfont(0.018)}} & \textbf{0.999{\fontsize{9}{\baselineskip}\selectfont(0.000)}} \\ \hline
\color{brown} \textbf{Average} &  Simfluence & 0.026 & 0.116 & 0.931 \\
 & Ours & \colorcell \textbf{0.015} & \colorcell\textbf{0.084} & \colorcell \textbf{0.933} \\ \hline
\end{tabular}
}
\caption{Results of test loss estimation for \textit{fine-tuning}.}
\label{tabel:loss-ft}
\end{table}

\paragraph{Fine-tuning} 
In Table~\ref{tabel:loss-ft}, it is evident that our approach consistently outperforms Simfluence when it comes to fine-tuning GPT models. On average, our method reduces the MSE and MAE across all training steps by $42\%$ and $28\%$, respectively, when compared to Simfluence. This implies that our method is more robust and adaptable in simulating training dynamics.

\subsection{Generalizing to Test Metric Estimation}

We have expanded the evaluation of our model beyond the mere prediction of test loss, now including vital measures such as ROUGE and BLEU scores. We have not reported the performance of TracIn and Grad-Dot baselines due to its inability on such metric predictions. 

\begin{figure*}[t]
    \centering
    \subfigure[Loss simulation on \textit{RTE}]
    {
        \label{fig:loss_a}
        \includegraphics[width=\columnwidth]{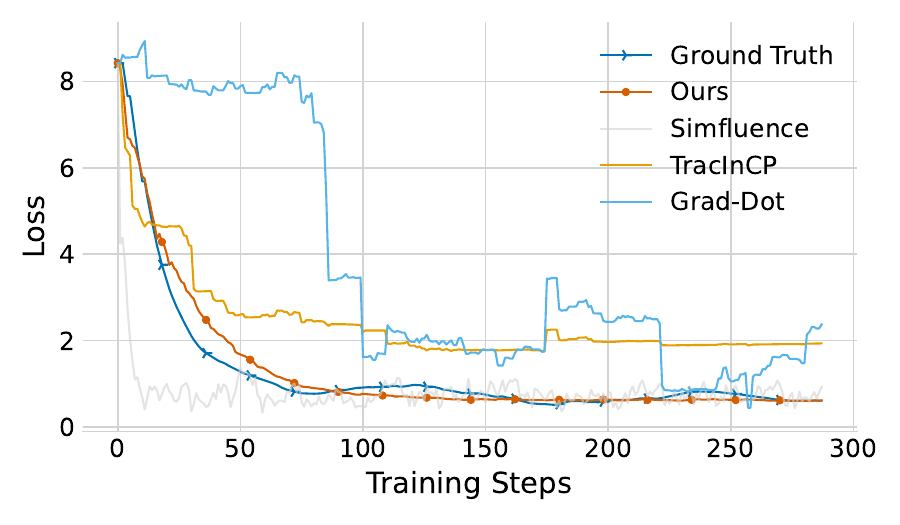}
    }
    \subfigure[Loss simulation on \textit{WMT16 DE/EN}]{
        \label{fig:loss_b}
        \includegraphics[width=\columnwidth]{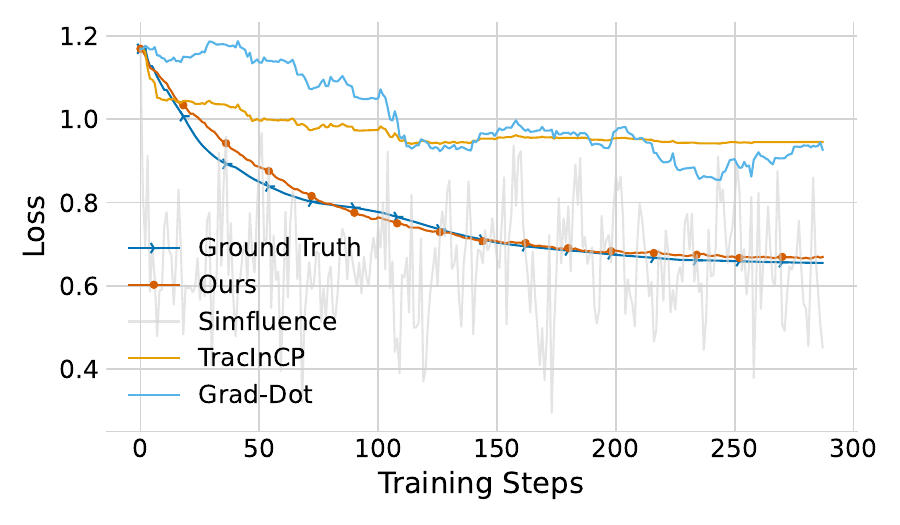}
    }
    \subfigure[BLEU simulation on \textit{WebNLG}]
    {
        \label{fig:bleu_c}
        \includegraphics[width=\columnwidth]{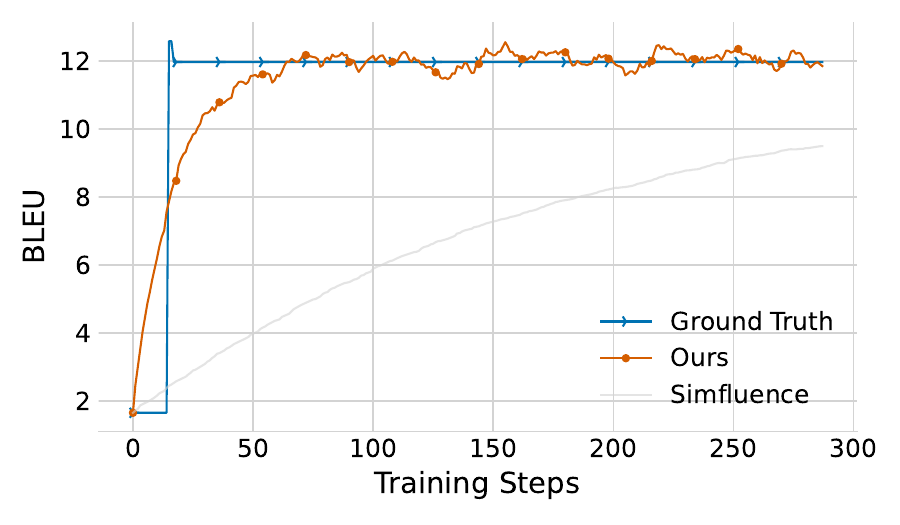}
    }
    \subfigure[ROUGE-L simulation on \textit{WMT16 DE/EN}]
    {
        \label{fig:rougeL_d}   \includegraphics[width=\columnwidth]{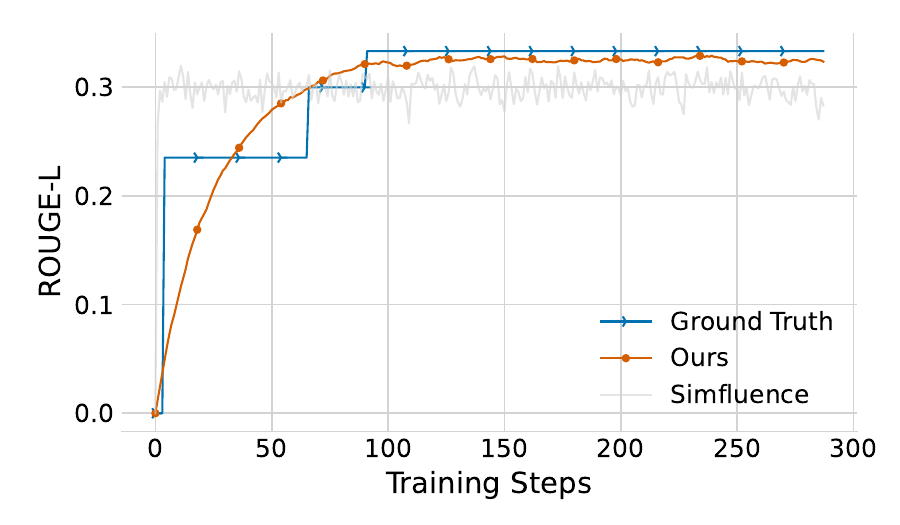}
    }
    \caption{Illustration of \textit{loss} and \textit{metric} simulation on \textbf{NLU} and \textbf{NLG} tasks with different TDA methods for \textit{instruction tuning}. See the \S\ref{ap:examples} for more examples.}
\end{figure*}

\paragraph{Instruction Tuning} 
As for instruction tuning, our findings, displayed in Table~\ref{table:metric-it}, demonstrate a superior performance of our method over Simfluence in predicting both BLEU and ROUGE-L scores and for GPTs of varying sizes. Intuitively, We draw some qualitative examples in the Fig.~\ref{fig:bleu_c} and~\ref{fig:rougeL_d}. Notably, for BLEU simulation on the WMT-16 DE/EN task, as the size of GPT increases, all steps MSE of Simfluence increases, whereas our method maintains a more stable performance, even exhibiting slight improvements from 0.92 to 0.93 in loss prediction accuracy at the final step. This suggests that our model is better equipped to manage more challenging tasks and larger model sizes, leveraging the pre-trained representations and instance interactions.

\paragraph{Fine-tuning}
Our method’s superiority remains evident in the fine-tuning scenario, as depicted in Table~\ref{table:metric-ft}, underscoring the robustness of our feature-based simulation approach. It’s worth noting that the margin by which {\name} outperforms Simfluence in BLEU metric simulation is not as pronounced in fine-tuning contexts as it is in instruction tuning settings. This discrepancy is likely due to the richer and more diverse data available in instruction tuning, which accentuates Simfluence’s relative inefficiency, given its independent parameter learning for \textit{each training instance} and a distinct simulator for each test instance.

\begin{table}[t]
    \centering
        \begin{minipage}{\columnwidth}
        \resizebox{\linewidth}{!}{
\begin{tabular}{@{}llp{2cm}p{2cm}p{2.6cm}p{2cm}p{2cm}p{2.6cm}@{}}
\toprule
 \multirow{3}{*}{\textbf{Method}} & \multirow{3}{*}{\textbf{\#Param}} & \multicolumn{6}{c}{\textbf{WebNLG}} \\ \cline{3-8} 
 &  & \multicolumn{3}{c|}{\textbf{BLEU}} & \multicolumn{3}{c}{\textbf{ROUGE-L}} \\ \cline{3-8} 
 &  & All-steps MSE ($\downarrow$) & All-steps MAE ($\downarrow$) & \multicolumn{1}{p{2.6cm}|}{Final-step Spearman’s $\rho$ ($\uparrow$)} & All-steps MSE ($\downarrow$) & All-steps MAE ($\downarrow$) & Final-step Spearman’s $\rho$ ($\uparrow$) \\ \hline
 Simfluence  & \multirow{2}{*}{410M} & 23.47{\fontsize{9}{\baselineskip}\selectfont(63.52)} & 2.34{\fontsize{9}{\baselineskip}\selectfont(3.26)} & \multicolumn{1}{p{2.6cm}|}{0.81{\fontsize{9}{\baselineskip}\selectfont(0.02)}} & 0.007{\fontsize{9}{\baselineskip}\selectfont(0.008)} & 0.055{\fontsize{9}{\baselineskip}\selectfont(0.038)} & 0.708{\fontsize{9}{\baselineskip}\selectfont(0.067)} \\
 Ours & & \textbf{9.11{\fontsize{9}{\baselineskip}\selectfont(18.41)}} & \textbf{1.73{\fontsize{9}{\baselineskip}\selectfont(1.82)}} & \multicolumn{1}{p{2.6cm}|}{\textbf{0.90{\fontsize{9}{\baselineskip}\selectfont(0.03)}}} & \textbf{0.005{\fontsize{9}{\baselineskip}\selectfont(0.006)}} & \textbf{0.045{\fontsize{9}{\baselineskip}\selectfont(0.034)}} & \textbf{0.796{\fontsize{9}{\baselineskip}\selectfont(0.047)}} \\ \cline{1-8} 
  Simfluence & \multirow{2}{*}{1B} & 20.58{\fontsize{9}{\baselineskip}\selectfont(60.80)} & 2.01{\fontsize{9}{\baselineskip}\selectfont(3.03)} & \multicolumn{1}{p{2.6cm}|}{\textbf{0.87{\fontsize{9}{\baselineskip}\selectfont(0.03)}}} & 0.006{\fontsize{9}{\baselineskip}\selectfont(0.006)} & 0.052{\fontsize{9}{\baselineskip}\selectfont(0.031)} & 0.878{\fontsize{9}{\baselineskip}\selectfont(0.035)} \\
 Ours & & \textbf{9.72{\fontsize{9}{\baselineskip}\selectfont(23.70)}} & \textbf{1.63{\fontsize{9}{\baselineskip}\selectfont(2.02)}} & \multicolumn{1}{p{2.6cm}|}{0.86{\fontsize{9}{\baselineskip}\selectfont(0.03)}} & \textbf{0.004{\fontsize{9}{\baselineskip}\selectfont(0.005)}} & \textbf{0.043{\fontsize{9}{\baselineskip}\selectfont(0.029)}} & \textbf{0.903{\fontsize{9}{\baselineskip}\selectfont(0.020)}} \\ \hline
 
 Simfluence & 2.8B & 15.08{\fontsize{9}{\baselineskip}\selectfont(51.72)} & 1.52{\fontsize{9}{\baselineskip}\selectfont(2.90)} & \multicolumn{1}{p{2.6cm}|}{0.80{\fontsize{9}{\baselineskip}\selectfont(0.08)}} & 0.005{\fontsize{9}{\baselineskip}\selectfont(0.006)} & 0.050{\fontsize{9}{\baselineskip}\selectfont(0.036)} & 0.817{\fontsize{9}{\baselineskip}\selectfont(0.063)} \\
 Ours &  & \textbf{5.56{\fontsize{9}{\baselineskip}\selectfont(17.26)}} & \textbf{1.15{\fontsize{9}{\baselineskip}\selectfont(1.42)}} & \multicolumn{1}{p{2.6cm}|}{\textbf{0.86{\fontsize{9}{\baselineskip}\selectfont(0.05)}}} & \textbf{0.003{\fontsize{9}{\baselineskip}\selectfont(0.003)}} & \textbf{0.035{\fontsize{9}{\baselineskip}\selectfont(0.026)}} & \textbf{0.911{\fontsize{9}{\baselineskip}\selectfont(0.050)}} \\

 \hline
 \multirow{3}{*}{\textbf{Method}} & \multirow{3}{*}{\textbf{\#Param}} & \multicolumn{6}{c}{\textbf{WMT-16 DE/EN}} \\ \cline{3-8} 
 &  & \multicolumn{3}{c|}{\textbf{BLEU}} & \multicolumn{3}{c}{\textbf{ROUGE-L}} \\ \cline{3-8} 
 &  & All-steps MSE ($\downarrow$) & All-steps MAE ($\downarrow$) & \multicolumn{1}{p{2.6cm}|}{Final-Step Spearman’s $\rho$ ($\uparrow$)} & All-steps MSE ($\downarrow$) & All-steps MAE ($\downarrow$) & Final-Step Spearman’s $\rho$ ($\uparrow$) \\ \hline
  Simfluence & \multirow{2}{*}{410M} & 32.15{\fontsize{9}{\baselineskip}\selectfont(116.17)} & 2.25{\fontsize{9}{\baselineskip}\selectfont(4.08)} & \multicolumn{1}{p{2.6cm}|}{0.83{\fontsize{9}{\baselineskip}\selectfont(0.03)}} & 0.007{\fontsize{9}{\baselineskip}\selectfont(0.017)} & 0.039{\fontsize{9}{\baselineskip}\selectfont(0.055)} & 0.931{\fontsize{9}{\baselineskip}\selectfont(0.014)} \\
 Ours & & \textbf{7.71{\fontsize{9}{\baselineskip}\selectfont(28.05)}} & \textbf{1.14{\fontsize{9}{\baselineskip}\selectfont(1.92)}} & \multicolumn{1}{p{2.6cm}|}{\textbf{0.92{\fontsize{9}{\baselineskip}\selectfont(0.02)}}} & \textbf{0.004{\fontsize{9}{\baselineskip}\selectfont(0.009)}} & \textbf{0.030{\fontsize{9}{\baselineskip}\selectfont(0.041)}} & \textbf{0.964{\fontsize{9}{\baselineskip}\selectfont(0.012)}} \\ \cline{1-8} 
  Simfluence & \multirow{2}{*}{1B} & 162.94{\fontsize{9}{\baselineskip}\selectfont(466.30)} & 5.71{\fontsize{9}{\baselineskip}\selectfont(9.03)} & \multicolumn{1}{p{2.6cm}|}{0.76{\fontsize{9}{\baselineskip}\selectfont(0.03)}} & 0.025{\fontsize{9}{\baselineskip}\selectfont(0.038)} & 0.094{\fontsize{9}{\baselineskip}\selectfont(0.098)} & 0.833{\fontsize{9}{\baselineskip}\selectfont(0.031)} \\
 Ours & & \textbf{46.33{\fontsize{9}{\baselineskip}\selectfont(122.50)}} & \textbf{3.34{\fontsize{9}{\baselineskip}\selectfont(4.68)}} & \multicolumn{1}{p{2.6cm}|}{\textbf{0.93{\fontsize{9}{\baselineskip}\selectfont(0.01)}}} & \textbf{0.013{\fontsize{9}{\baselineskip}\selectfont(0.020)}} & \textbf{0.066{\fontsize{9}{\baselineskip}\selectfont(0.069)}} & \textbf{0.910{\fontsize{9}{\baselineskip}\selectfont(0.011)}} \\ \hline
 
Simfluence & 2.8B & 64.07{\fontsize{9}{\baselineskip}\selectfont(319.93)} & 2.59{\fontsize{9}{\baselineskip}\selectfont(5.84)} & \multicolumn{1}{p{2.6cm}|}{0.90{\fontsize{9}{\baselineskip}\selectfont(0.05)}} & 0.008{\fontsize{9}{\baselineskip}\selectfont(0.022)} & 0.040{\fontsize{9}{\baselineskip}\selectfont(0.059)} & 0.912{\fontsize{9}{\baselineskip}\selectfont(0.045)} \\
Ours &  & \textbf{24.27{\fontsize{9}{\baselineskip}\selectfont(93.41)}} & \textbf{1.94{\fontsize{9}{\baselineskip}\selectfont(3.36)}} & \multicolumn{1}{p{2.6cm}|}{\textbf{0.93{\fontsize{9}{\baselineskip}\selectfont(0.05)}}} & \textbf{0.005{\fontsize{9}{\baselineskip}\selectfont(0.018)}} & \textbf{0.030{\fontsize{9}{\baselineskip}\selectfont(0.051)}} & \textbf{0.936{\fontsize{9}{\baselineskip}\selectfont(0.037)}} \\
 
 \hline
\multirow{3}{*}{\textbf{Method}} & \multirow{3}{*}{\textbf{\#Param}} & \multicolumn{6}{c}{\color{brown} \textbf{Average}} \\ \cline{3-8} 
 &  & \multicolumn{3}{c|}{\textbf{BLEU}} & \multicolumn{3}{c}{\textbf{ROUGE-L}} \\ \cline{3-8} 
 &  & All-steps MSE ($\downarrow$) & All-steps MAE ($\downarrow$) & \multicolumn{1}{p{2.6cm}|}{Final-step Spearman’s $\rho$ ($\uparrow$)} & All-steps MSE ($\downarrow$) & All-steps MAE ($\downarrow$) & Final-step Spearman’s $\rho$ ($\uparrow$) \\ \hline
Simfluence & \multirow{2}{*}{410M} & 27.81 & 2.29 & \multicolumn{1}{l|}{0.82} & 0.007 & 0.047 & 0.820 \\
Ours &  & \colorcell \textbf{8.41} & \colorcell \textbf{1.43} & \multicolumn{1}{l|}{\colorcell \textbf{0.91}} & \colorcell \textbf{0.004} & \colorcell \textbf{0.037} & \colorcell \textbf{0.880} \\ \hline
Simfluence & \multirow{2}{*}{1B} & 91.76 & 3.86 & \multicolumn{1}{l|}{0.81} & 0.015 & 0.073 & 0.855 \\
Ours &  & \colorcell \textbf{28.02} & \colorcell \textbf{2.51} & \multicolumn{1}{l|}{\colorcell \textbf{0.90}} & \colorcell \textbf{0.008} &  \colorcell\textbf{0.055} & 
 \colorcell \textbf{0.907} \\ \hline
 
 Simfluence & 2.8B & 39.58 & 2.06 & \multicolumn{1}{p{2.6cm}|}{0.85} & 0.007 & 0.045 & 0.865 \\
 Ours &  & \colorcell \textbf{14.92} & \colorcell \textbf{1.55} & \multicolumn{1}{p{2.6cm}|}{\colorcell \textbf{0.89}} & \colorcell \textbf{0.004} & \colorcell \textbf{0.033} & \colorcell \textbf{0.924} \\
 \bottomrule
\end{tabular}
}
\caption{Results of test metric estimation on NLG datasets for \textit{instruction-tuning}.} 

\label{table:metric-it}
    \end{minipage}
    \\[12pt]
    \begin{minipage}{\columnwidth}
        \resizebox{\linewidth}{!}{
        \begin{tabular}{@{}p{1.5cm}llp{2.2cm}p{2.2cm}p{2.6cm}@{}}
        \toprule
        \textbf{Dataset} & \textbf{Metric} & \textbf{Method} & All-steps MSE ($\downarrow$) & All-steps MAE ($\downarrow$) & Final-Step Spearman’s $\rho$ ($\uparrow$) \\ \midrule
        \multirow{4}{*}{WebNLG} & \multirow{2}{*}{BLEU} & Simfluence & \textbf{43.33{ \small (77.34)}} & \textbf{4.23{ \small (3.52)}} & 0.78{ \small (0.02)} \\
         &  & Ours & 43.98{ \small (81.40)} & 4.28{ \small (3.57)} & \textbf{0.80{ \small (0.01)}} \\ \cmidrule(l){2-6} 
         & \multirow{2}{*}{ROUGE-L} & Simfluence & 0.008{ \small (0.007)} & 0.066{ \small (0.031)} & 0.706{ \small (0.038)} \\
        &  & Ours & \textbf{0.007{ \small (0.006)}} & \textbf{0.060{ \small (0.029)}} & \textbf{0.765{ \small (0.040)}} \\ \midrule
        \multirow{4}{*}{\parbox{1.5cm}{WMT-16 DE/EN}} & \multirow{2}{*}{BLEU} & Simfluence & 32.11{ \small (89.13)} & \textbf{2.76{ \small (3.75)}} & \textbf{0.82{ \small (0.02)}} \\
         &  & Ours & \textbf{30.26{ \small (77.23)}} & 2.91{ \small (3.69)} & 0.81{ \small (0.02)} \\ \cmidrule(l){2-6} 
         & \multirow{2}{*}{ROUGE-L} & Simfluence & 0.018{ \small (0.025)} & 0.091{ \small (0.075)} & 0.796{ \small (0.032)} \\
         &  & Ours & \textbf{0.012{ \small (0.016)}} & \textbf{0.075{ \small (0.057)}} & \textbf{0.843{ \small (0.010)}} \\ \midrule
        \multirow{4}{*}{ \textbf{Average}} & \multirow{2}{*}{BLEU} & Simfluence & 37.72 & \textbf{3.49} & 0.80 \\
         &  & Ours & \colorcell\textbf{37.12} & 3.59 & \colorcell\textbf{0.81} \\ \cmidrule(l){2-6} 
         & \multirow{2}{*}{ROUGE-L} & Simfluence & 0.013 & 0.079 & 0.751 \\
        &  & Ours & \colorcell \textbf{0.009} & \colorcell\textbf{0.068} & \colorcell\textbf{0.805} \\ \bottomrule
        \end{tabular}
        }
        \vspace{-2mm}
        \caption{Results of test metric estimation on NLG datasets for \textit{fine-tuning}.}
         \vspace{-4mm}
        \label{table:metric-ft}
    \end{minipage}
\end{table}

\subsection{Ablation Study}


%
\paragraph{Practical Influence via Checkpoints} 
Our featured simulator is adept at learning from past training dynamics. However, monitoring the training dynamics at every step can be expensive, especially when dealing with large-sized GPTs. Therefore, we conduct experiments to choose training checkpoints at specific intervals to approximate the reality of the neighboring points with the training state of that particular point. Then, we trained our simulator on the approximate training dynamics to 
find the balance between the cost of collecting training dynamics and the simulator performance.

\begin{figure}[t]
    \centering
    \begin{minipage}{\columnwidth}
            \includegraphics[width=\linewidth]{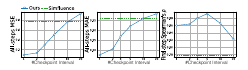}
            \caption{Variation curves of the average performance of {\name} for loss simulation in five datasets when different checkpoint intervals are selected.}
            \label{fig:ckpt_interval_flan_loss}
    \end{minipage}
    \\[12pt]
    \subfigure[RTE]{
    \begin{minipage}[t]{\linewidth}
        \centering
        \label{fig:a}
        \includegraphics[width=\columnwidth]{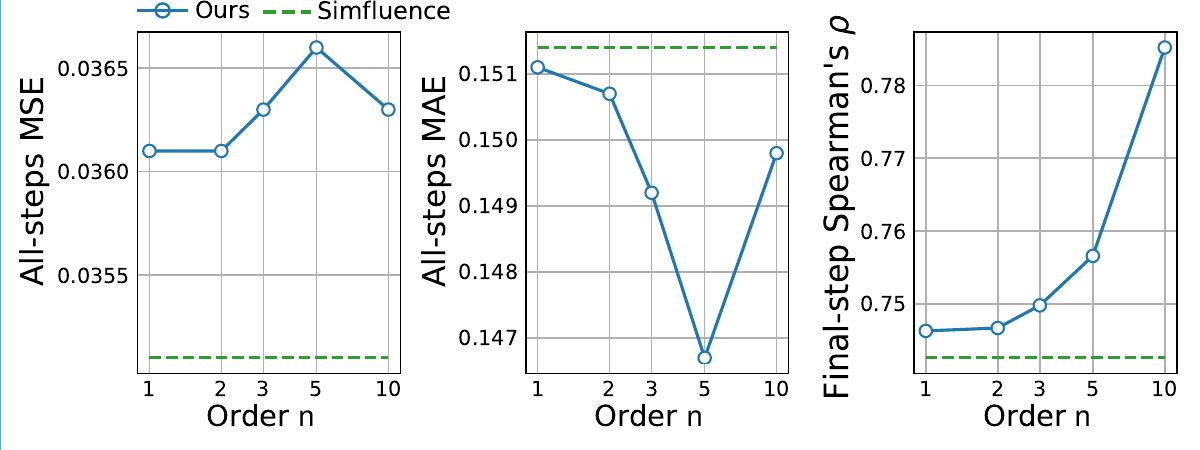}
    \end{minipage}
    } 
    \subfigure[WebNLG]{
    \begin{minipage}[t]{\linewidth}
        \centering
        \label{fig:b}
        \includegraphics[width=\columnwidth]{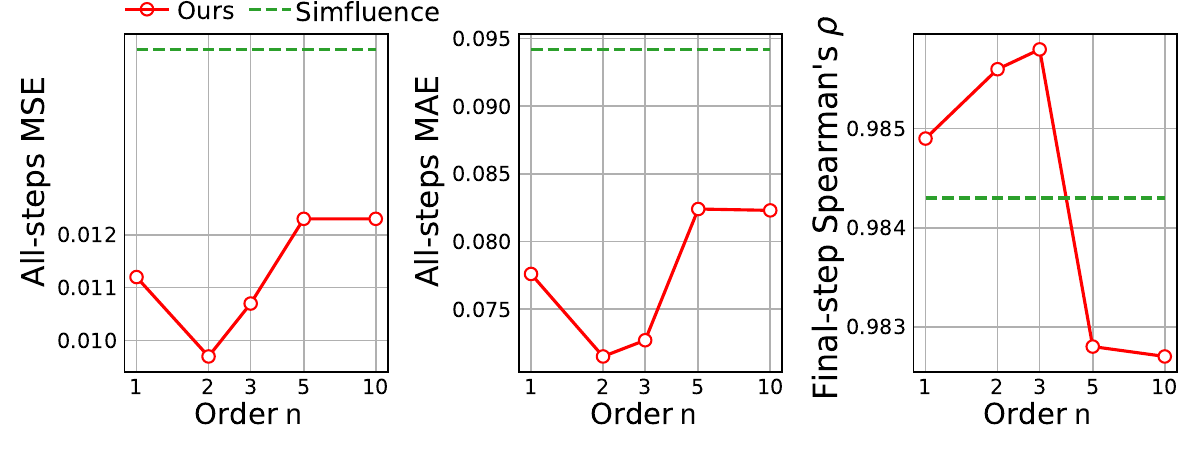}
    \end{minipage}
    }
    \caption{Analysis on the impact of \textit{$n$-th order Markov process} on language understanding (RTE) and generation (WebNLG) tasks, varying $n$ from 1 to 10.}
    \label{fig:order}
\end{figure}

Results are shown in Fig.~\ref{fig:ckpt_interval_flan_loss}. Unless otherwise specified, we instruction tuning the Pythia-410M for further analysis. In general, the performance of our simulator deteriorates as the number of checkpoint intervals increases. This is manifested by a rise in MSE and MAE at all steps and a drop in Spearman's $\rho$ when the checkpoint interval is large. However, even when the number of checkpoint intervals is equal to 10, which means that we will use the training state of one point to approximate the training state of the previous ten points and the training dynamics collection time will be shortened by almost $90\%$, our method still has comparable prediction error at all steps and better Spearman coefficient than Simfluence.

\paragraph{Empirical Analysis of Markov Order Dependency} 
Using the first-order Markov process to predict future states based on the prior step, potentially oversimplifies GPT training dynamics. Therefore, we consider the training dynamics as an $n$-th order Markov process ($n=2,3,5,10$) and experiment on both language understanding (RTE) and generative (WebNLG) tasks.

The result can be seen in Fig.~\ref{fig:order}. Overall, when considering more preceding training information, the simulation error initially increases and decreases for both datasets, as indicated by the all-steps MSE metric. It suggests that a high order $n$ might introduce noise, leading to a degraded simulator's performance. Moreover, the final-step Spearman's $\rho$ shows a significant increase from 0.746 to 0.785 for RTE with the increase of order $n$, but not the same for WebNLG. We guess considering more past training information could improve the prediction accuracy for NLU tasks.


\begin{figure}
    \centering
    \includegraphics[width=\linewidth]{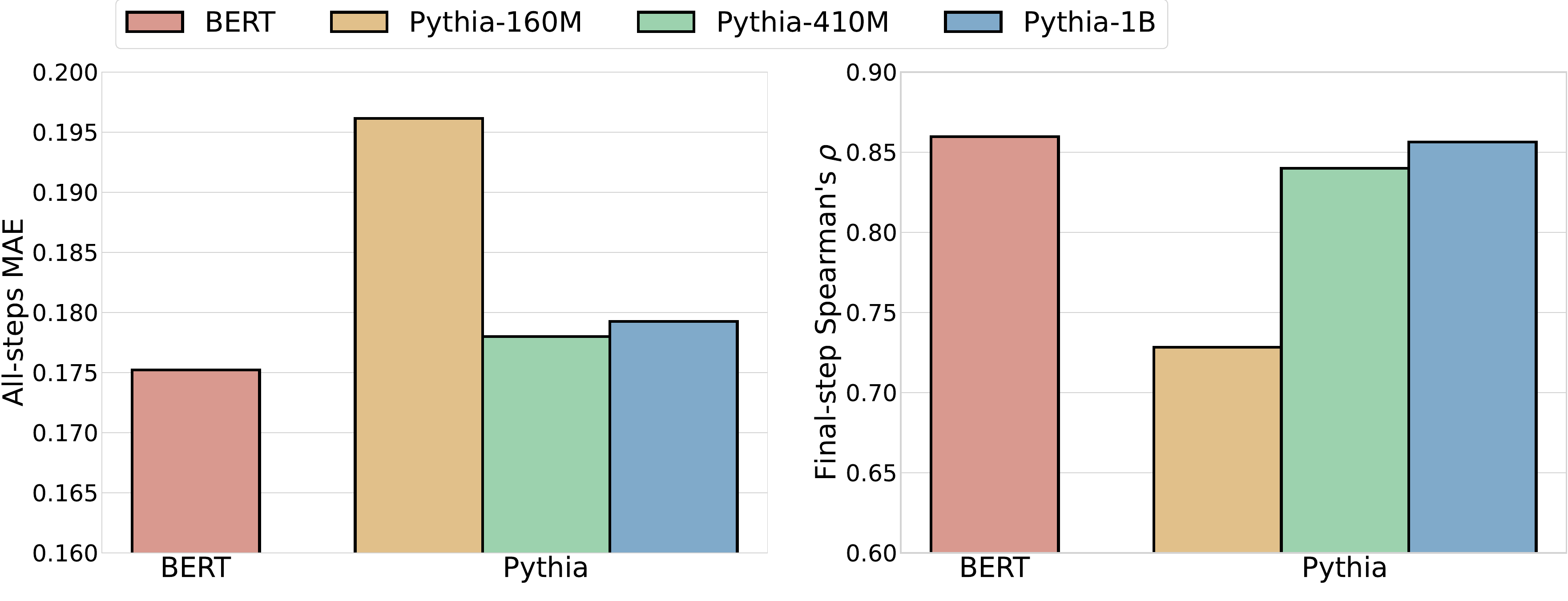}
    \caption{Impact of feature representation of different pre-trained encoders on loss simulation.}
    \label{fig:feature}
\end{figure}

\paragraph{Impact of Different Feature Representations}
To further explore the impact of various feature representations, we conducted experiments on two types of pre-trained encoders: BERT~\footnote{\url{https://huggingface.co/sentence-transformers/all-MiniLM-L6-v2}} and Pythia~\footnote{\url{https://github.com/EleutherAI/pythia}} with different sizes. Results are shown in Fig.~\ref{fig:feature}. In general, BERT's feature representations produce better simulation results than the Pythia encoder. This could be due to its ability to encode context information in both directions. Interestingly, we also found that increasing the parameters of the Pythia encoder does not always lead to better performance of the performance simulator.

\subsection{Analysis}
\begin{figure}
    \centering
    \includegraphics[width=\linewidth]{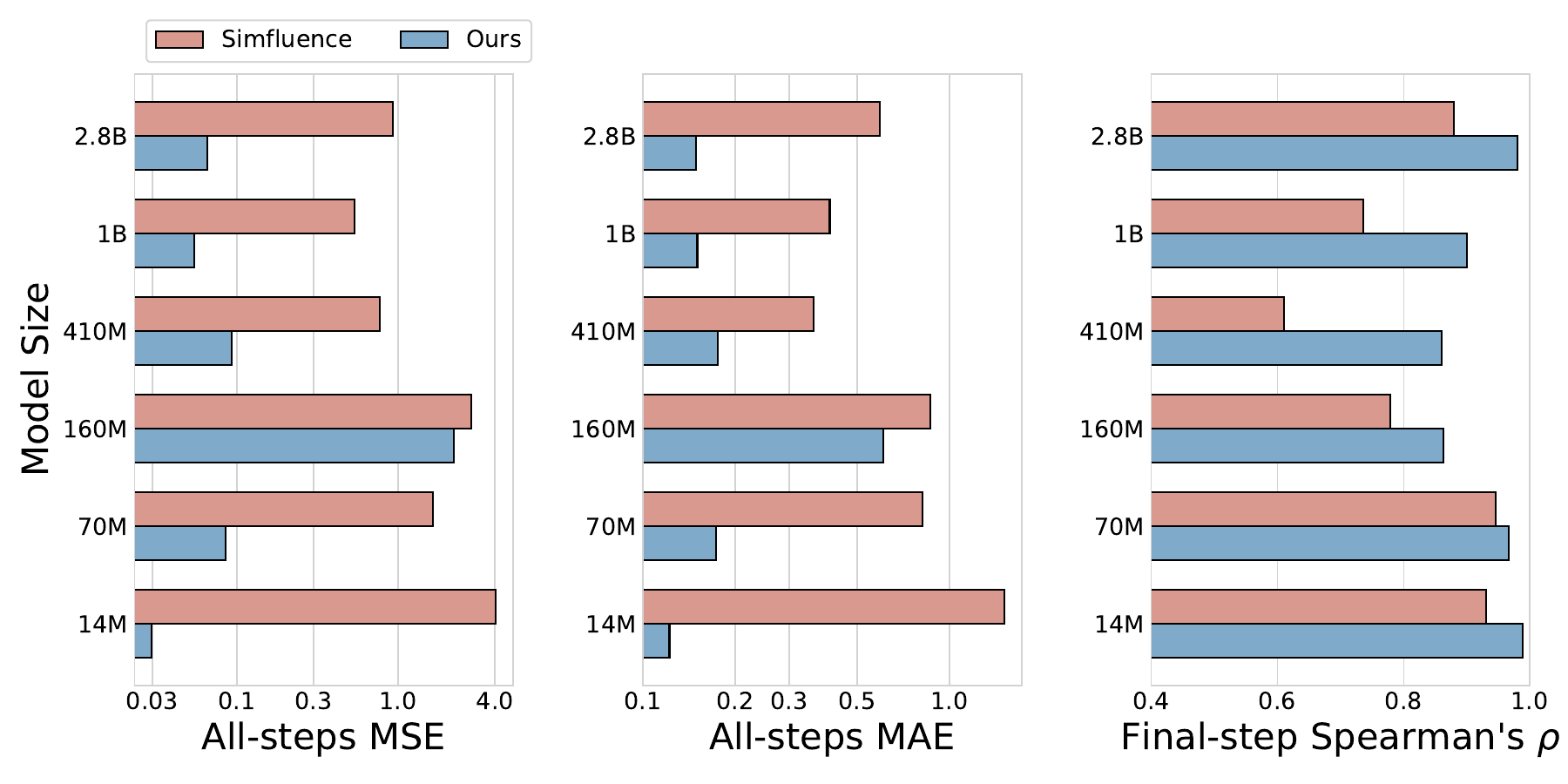}
    \caption{Comparison of the loss simulation between {\name} and Simfluence on instruction tuning Pythia model series, ranging from 14M to 2.8B.}
    \label{fig:sizes}
\end{figure}

\paragraph{Robustness across Varying Model Sizes} 
We conducted experiments to validate how our simulator handles the complexity of GPTs of different sizes, ranging from 14M to 2.8B, specifically focusing on instruction tuning scenarios. Results are presented in Fig.~\ref{fig:sizes}. Our loss simulation experiments revealed that despite the inconsistent simulation performance trend with increasing GPT size, our featurized model consistently surpassed Simfluence. These findings demonstrate the superiority of our model in effectively capturing and managing model complexity.

\paragraph{Unseen Data Generalization} 
Unlike Simfluence, which restricts the parameters only indexed by seen samples of past training runs, our {\name} can handle unseen samples via sample parameterization. We conducted experiments on RTE and WebNLG tasks in fine-tuning scenarios to further verify the unseen data generalization. For a future training run, we experiment in three different unseen data scenarios: 1) Examples in the training curriculum are unseen; 2) Test examples are unseen; 3) Both examples in training the curriculum and test examples are unseen.

We defer the results in Table~\ref{tab:unseen_ft} in Appendix. Overall, {\name} can generalize to unseen data, which includes simulating loss and performance metrics. What's more, we find that {\name} is better at generalizing to unseen training data to simulate the impact of test samples that have been seen in the past. To illustrate this more visually, we show the effect of GPTfluence's simulation of the unseen training data setting with loss and performance metrics, respectively. As shown in Fig.~\ref{fig:unsen_ft}, the generalization performance of {\name} is mostly satisfactory.


\subsection{Use Case: Mislabelled Data Identification}
\label{ap:mislabelled}
Following previous studies~\cite{yeh2018representer,pruthi2020tracin},
we present a mislabeled data identification use case to evaluate our TDA-based method.

\paragraph{Experimental Setup} We employ the Pythia-410M model as our classifier and utilize a subset of the SST-2 dataset. The methods compared include the following: \textbf{Random}, where we bypass influence calculation and apply random shuffling\footnote{Random shuffling is performed ten times with varying seeds, and the average result is reported.}. \textbf{TracIn-CP}, which uses self-influence as the metric by computing the gradient dot-product between a sample and itself. Similarly, \textbf{GPTfluence} calculates the influence by simulating the multiplicative factor $\alpha$ on the sample itself.

\begin{figure}
    \centering
    \includegraphics[width=\linewidth]{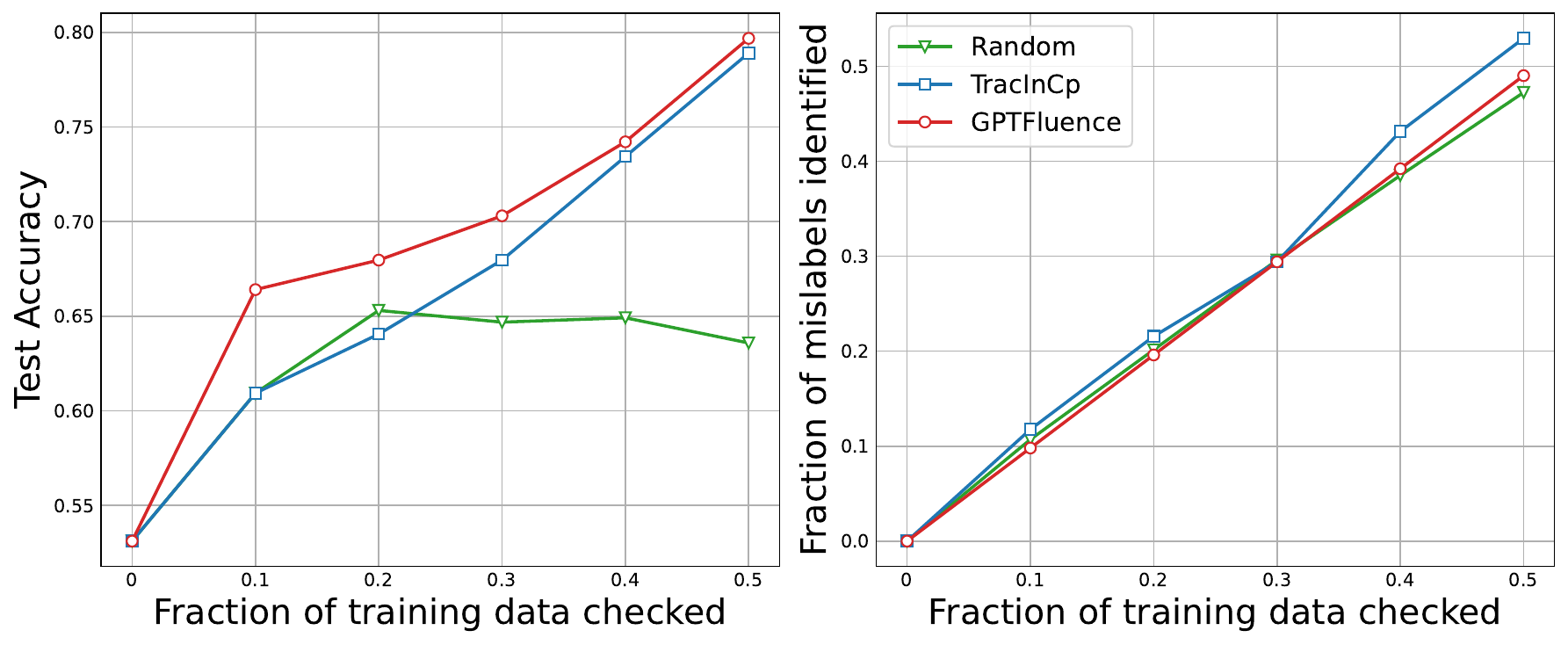}
    \caption{SST-2 Mislabelled Data Identification with GPTfluence, TracIn-CP and Random Selection.}
    \label{fig:mislabel}
\end{figure}

\paragraph{Results} The results are depicted in Fig.~\ref{fig:mislabel}. When examining the fraction of mislabelled data identified, GPTfluence demonstrates comparable performance to random selection, albeit slightly underperforming compared to TracIn-CP. However, the marginal difference in mislabel detection is offset by the notable improvement in test accuracy achieved with GPTfluence. Our method outperforms both TracIn-CP and random selection, particularly excelling in the early stages of mislabel detection, which is crucial when reviewing a small fraction of data. In scenarios where precision is key, especially with limited data available for review, GPTfluence proves its efficacy.

To simulate mislabeled data, we corrupted 40\% of the training set by flipping the labels, resulting in an initial classification accuracy of 0.53. We then sequentially corrected mislabelled samples by inspecting fractions of the dataset ranked by our influence metric, computed via the TDA method. After correcting the mislabels, we retrained the classifier and reported the test accuracy on the cleaned dataset.


\section{Related Work}
\label{sec:bg}
Our methodology extends the frontier of TDA techniques, which are instrumental in understanding the influence of individual training instances on model predictions. This body of work bifurcates into two main strands: gradient-based approximation methods and simulation-based approaches.

\paragraph{Gradient-Based Approximation Methods} This strand of research capitalizes on gradient information to infer the influence of training instances on model predictions, providing a quantifiable measure of individual data points' contributions~\cite{koh2017influence-function,yeh2018representer,k2021revisiting}. Influence Functions, a pioneering method in this domain, leverages the mathematical framework of influence functions for estimating the impact of dataset perturbations on model predictions. Complementing this, TracIn~\cite{pruthi2020tracin} employs gradient-based approximations to trace the influence of training data on test predictions. Similarly, Grad-Dot~\cite{charpiat2019input} uses gradient dot products to approximate the influence of training examples. A contemporary work~\cite{xia2024less} that adapts the TracIn framework for models optimized with Adam. LESS incorporates LoRA~\cite{hu2021lora} and random projection~\cite{park2023trak} techniques to enhance data selection processes.
These methods primarily rely on gradients to quantify data influence, offering tractable solutions with varying degrees of approximation accuracy.

\paragraph{Simulation-Based Approaches} An alternative research vein adopts model-based simulations to represent training dynamics~\cite{ilyas2022datamodels, guu2023simfluence}. Simfluence~\cite{guu2023simfluence} pioneers the simulation-based category by learning a linear model that predicts the influence of training examples through multiplicative and additive factors, as detailed in \S\ref{sec:pre}. Recent efforts~\cite{engstrom2024dsdm} have focused on simulating the overall influence of training examples, aiming at predicting the cumulative influence of training data for refined data selection.

Our contribution distinctly advances the simulation-based direction by forecasting the end-point influence and modeling the entire trajectory of training dynamics using featurized representations. This approach provides a more in-depth understanding of training data influence, facilitating dynamic adjustments and insights into the model training curricula.




\section{Conclusion and Future Work}
\label{sec:concl}

In this paper, we explore the data attribution analysis for GPT models through \name, a novel featurized simulator approach. This methodology not only surpasses the predictive capabilities of traditional test loss metrics but forecasts essential task performance metrics across a broad spectrum of GPT model sizes, ranging from 14M to 2.8B parameters. Our comprehensive evaluations across diverse downstream tasks and fine-tuning scenarios substantiate the superior efficacy of our approach. In the future, extending this approach to other tasks and training regime presents a promising avenue for future research.


\section*{Acknowledgements}
We would like to thank all anonymous reviewers for their insightful and constructive feedback. Qiwei Peng is supported by DisAI - Improving scientific excellence and creativity in combating disinformation with artificial intelligence and language technologies, a project funded by European Union under the Horizon Europe, GA No. 101079164. 

\section*{Ethical Consideration}

While our study focuses on predicting the influence of training data on GPT models, we recognize the broader ethical implications that our research may entail, especially as it contributes to the advancement of large language models (LLMs) that are increasingly integrated into societal functions.

\paragraph{Data Use and Privacy} Our research utilizes publicly available datasets and respects privacy concerns by anonymizing any potentially identifiable information. We ensure that our data handling practices comply with all relevant data protection regulations and ethical guidelines, safeguarding against misuse.


\paragraph{Potential Misuse} We are cognizant of the potential misuse of predictive models in manipulating or unfairly influencing AI systems. Our research aims to contribute to the understanding and mitigation of such risks by providing tools to analyze and adjust the influence of training data. We encourage the application of our findings in ethical ways that promote fairness and transparency in AI.

\paragraph{Broader Impact} 
This study advances understanding of data influence on LLMs, offering a methodological approach for detailed impact analysis. This work not only enhances the interpretability and transparency of LLMs but also lays the groundwork for more informed and ethical decisions in data curation and model training. 

\section*{Limitations}
This work introduces a novel feature-based approach within the simulation-based framework for predicting the influence of training data on GPT models. While our methodology represents a significant advancement in the field, it is not without its limitations, which we discuss below:

\paragraph{Dependence on Extensive Training Dynamics} A fundamental constraint of our approach is its reliance on a comprehensive set of training dynamics to train the simulator effectively. This requirement, while crucial for the accuracy of our predictions, necessitates considerable computational resources and time. The efficiency of data influence simulators remains an area ripe for further exploration, with the aim of reducing the computational overhead without compromising on performance.

\paragraph{Limited Dataset Scope} Our experimental validation is confined to a subset of the FLAN datasets, constrained by the logistical and computational costs associated with collecting a large-scale training dynamics dataset. Despite this limitation, we have conducted\textbf{ over 352 training experiments} across six different GPT model sizes (ranging from 14M to 2.8B parameters) to amass the \textit{GPTDynamics} dataset. This dataset, which we are making publicly available, is a step towards mitigating the data scarcity in this research area, yet the need for more expansive datasets encompassing a broader range of tasks and languages remains.

\paragraph{Model Size Constraints} The high computational costs involved in executing multiple runs on larger language models, such as those with 13B or even 72B parameters, have limited the scale of the models we could feasibly include in our study. While our findings are robust across the examined model sizes, extending our analysis to larger models with hundreds of billions of parameters would likely yield additional insights into the scalability and generalizability of our approach.

\paragraph{Generalization to Other Domains} While our study focuses on GPT models and a specific subset of datasets, the generalizability of our approach to other model architectures and domains is not fully explored. Future work could extend our methodology to different types of language models and beyond, including vision and multimodal systems, to assess the applicability and adaptability of our featurized simulation-based approach.

\bibliography{custom}

\clearpage
\appendix

\section{Implementation Details}
\label{ap:hyperparam}

\subsection{Tasks and Datasets for \textit{GPTDynamics}}
\label{ap:dataset}

We conduct experiments on a subset of \textbf{FLAN}~\cite{wei2022flan}, a diverse array of datasets for \textit{instruction tuning}, to conduct a thorough evaluation of TDA methods. Our dataset selection spans both NLU and NLG tasks, thereby offering a broad spectrum of challenges for TDA methods to tackle.

The NLU tasks selected include \textbf{RTE} (Natural Language Inference), \textbf{SST-2} (Sentiment Classification), and \textbf{BoolQ} (Reading Comprehension). For NLG, we delve into \textbf{WebNLG} (Struct-to-Text) and \textbf{WMT-16 DE/EN} (Machine Translation) tasks.

To exploit the superior generalization benefits that instruction tuning brings to language models, we have assembled a specialized subset for instruction fine-tuning. This subset amalgamates the previously mentioned five tasks with \textbf{CNN-DM} (Summarization), crafting an extensive testing environment of FLAN data. We sourced task-specific instructions directly from the original FLAN paper.



\subsection{Comparison Baselines}
\label{ap:baseline}

\quad \textbf{TracIn}~\cite{pruthi2020tracin} is a gradient-based used to calculate the influence through a first-order gradient approximation. It considers the influence of the training example $z$ on the test example $z'$ as a loss change in $z'$, which is provided by each gradient step of the training example $z$. In practice, TracInCP was proposed as an alternative approximation that considers specific checkpoints during training. TracInCP calculates the gradient dot product of $z$ and $z'$ at these checkpoints. In our experiments, we used TracInCP with 10 checkpoints and all steps' checkpoints to estimate the influence.

\textbf{Grad-Dot}~\cite{charpiat2019input} is a heuristic gradient-based TDA method. They also compute the effect of a training sample on a test sample by the dot product of the gradients but computed on top of the final trained model.

\textbf{Simfluence}~\cite{guu2023simfluence} is a novel framework for TDA. It characterizes the loss variation of test samples during training by modeling it as a Markov process. Then, it learns a unique multiplicative and additive influence parameter for each training example. It is worth noting that in the original paper, the framework that considers both multiplicative and additive influences is referred to as \textit{Simfluence-linear}. However, for simplicity in this paper, we use the term \textit{Simfluence} to refer to the same model.

\subsection{Implementation Details of Instruction Tuning}
\paragraph{\textit{GPTDynamics} Collection for Instruction Tuning}
We instruction tuned Pythia from 14M to 2.8B (\emph{i.e.}, 14M, 70M, 160M, 410M, 1B, and 2.8B) on the instruction tuning dataset referenced in Appendix~\ref{ap:dataset}. We collect a total of randomly sampled 768 instances from aforementioned five tasks, with each samples 128 of 200 data points in one training run for instruction tuning. The data division followed the same protocol as in the fine-tuning scenarios. All Pythia models underwent comprehensive fine-tuning, with the exception of the Pythia-2.8B model, which was fine-tuned using the parameter-efficient LoRA technique \cite{hu2021lora}. The LoRA module was implemented within the query, key, and value projection matrices of the self-attention module, with a LoRA rank of 8, alpha set to 4, and a dropout probability of 0.05. We evaluated the Pythia models using the identical datasets as those in the fine-tuning experiments.  For the WebNLG and WMT16 DE/EN datasets, we evaluated BLEU and ROUGE-L scores in addition to test loss, employing a top-$p$ sampling strategy for generation with a temperature of 0.2 and top-$p$ probability of 0.95. Detailed instruction-tuning hyperparameters are reported in Table~\ref{tab:hyperparam_it}.

\begin{table*}[th]
\centering
\resizebox{\linewidth}{!}{
\begin{tabular}{@{}lp{2.5cm}<{\centering}p{2.5cm}<{\centering}p{2.5cm}<{\centering}p{2.5cm}<{\centering}p{2.5cm}<{\centering}p{2.5cm}<{\centering}@{}}
\toprule
Instruction-Tuning Hyperparameters & Pythia-14M & Pythia-70M & Pythia-160M & Pythia-410M & Pythia-1B & Pythia-2.8B \\ \midrule
Optimizer &  &  &  & AdamW &  &  \\
Adam’s $\beta$ &  &  &  & (0.9, 0.999) &  &  \\
Adam’s $\epsilon$ &  &  &  & 1e-6 &  &  \\
Weight decay &  &  &  & 0.001 &  &  \\
Learning rate & 5e-7 & 5e-7 & 5e-7 & 2e-7 & 2e-7 & 1e-5 \\
Learning rate schedule &  &  &  & Linear decay &  &  \\
Warmup steps &  &  &  & 0 &  &  \\
Batch size &  &  &  & 8 &  &  \\
Max sequence length & 2048 & 2048 & 2048 & 2048 & 2048 & 1024 \\
Training epochs & 3 & 3 & 3 & 3 & 2 & 2 \\
Training steps & 288 & 288 & 288 & 288 & 192 & 192 \\
Precision &  &  &  &  fp32 &  &  \\ \bottomrule
\end{tabular}
}
\caption{Hyper-parameter settings for instruction tuning \textit{GPTDynamics} data across Pythia models, ranging in size from 14M to 2.8B.}
\label{tab:hyperparam_it}
\end{table*}

\paragraph{{\name} Training Setup}

The architecture of our simulator is a pre-trained sentence encoder followed by parallel weight-sharing fully-connected layers for predicting influence factors. The trainable model size of the simulator is 11.4M excluding pre-trained embeddings (frozen). Unless specified, we use the sentence transformer\footnote{\url{https://huggingface.co/sentence-transformers/all-MiniLM-L6-v2}} as our pre-trained encoder.
For the simulator training, we combine all five FLAN datasets and train our simulator in a multi-task manner, each dataset has 27 training runs. All reported results are averaged over 5 held-out runs.
We set the order $n$ of Markov process assumptions equal to 1 for instruction tuning. Detailed training hyperparameters of {\name} are shown in Table~\ref{tab:hyperparameters_it_training}.


\begin{table*}[th]
\centering
\resizebox{\linewidth}{!}{
\begin{tabular}{@{}lp{2.5cm}<{\centering}p{2.5cm}<{\centering}p{2.5cm}<{\centering}p{2.5cm}<{\centering}p{2.5cm}<{\centering}p{2.5cm}<{\centering}@{}}
\toprule
Hyperparameters & Pythia-14M & Pythia-70M & Pythia-160M & Pythia-410M & Pythia-1B & Pythia-2.8B \\ \midrule
L2 regularizaiton $\lambda$ &  &  &  & 1e-5 &  &  \\
Optimizer &  &  &  & AdamW &  &  \\
Adam’s $\beta$ &  &  &  & (0.9, 0.999) &  &  \\
Adam’s $\epsilon$ &  &  &  & 1e-8 &  &  \\
Learning rate & 1e-6 & 1e-6 & 1e-6 & 1e-5 & 1e-5 & 1e-5 \\
Learning rate schedule &  &  &  & Linear decay &  &  \\
Warmup steps &  &  &  & 200 &  &  \\
Batch size &  &  &  & 128 &  &  \\
Max training epochs & 50 & 50 & 50 & 50 & 50 & 50 \\
Pre-trained encoder &  &  &  & MiniLM-L6-v2 &  &  \\
Max sequence length & 512 & 512 & 512 & 512 & 512 & 512 \\
Early stopping &  &  &  & \ding{51} &  &  \\
Precision &  &  &  & fp32 &  &  \\
Seed &  &  &  & 42 &  &  \\ \bottomrule
\end{tabular}
}
\caption{Hyperparameters of training our featurized \textbf{simulator} for \textit{instruction tuning} on Pythia models of size from 14M to 2.8B. We use the same training hyperparameters as in the loss simulation for the BLEU and ROUGE-L score simulation on WebNLG and WMT16 DE/EN datasets.}
\label{tab:hyperparameters_it_training}
\end{table*}



\subsection{Implementation Details of Fine-Tuning}
\paragraph{\textit{GPTDynamics} Collection for Fine-Tuning}
All the experiments are conducted on the NVIDIA Tesla V100 GPUs unless specified. We fine-tune Pythia-410M on five datasets: SST-2, BoolQ, RTE, WebNLG, and WMT16 DE/EN. For each dataset, we perform a total of 32 training runs, with each sample 128 of 200 data points from the original training set for GPT training. The split of training runs is divided into 25 for training, 2 for validation, and 5 for test. All reported results are averaged over 5 held-out runs. For NLG datasets, we measure BLEU, ROUGE-L scores besides the test loss, using a top-$p$ sampling strategy for generation with a temperature setting of 0.2 and a top-$p$ probability of 0.95. Note that we collect ROUGE-L scores on a scale from 0 to 1. The fine-tuning hyperparameters are shown in Table~\ref{ap:hyperparam_ft}.

\begin{table*}[t]
\centering
\resizebox{0.9\linewidth}{!}{
\begin{tabular}{@{}lp{2cm}<{\centering}p{2cm}<{\centering}p{2.3cm}<{\centering}p{2cm}<{\centering}p{2.5cm}<{\centering}@{}}
\toprule
Fine-Tuning Hyperparameters & SST-2 & RTE & BoolQ & WebNLG & WMT16 DE/EN \\ \midrule
Optimizer &  &  & AdamW &  &  \\
Adam $\beta$ &  &  & (0.9, 0.999) &  &  \\
Adam $\epsilon$ &  &  & 1e-6 &  &  \\
Weight decay &  &  & 0.001 &  &  \\
Learning rate & 5e-7 & 5e-7 & 5e-7 & 1e-6 & 5e-7 \\
Learning rate schedule &  &  & Linear decay &  &  \\
Warmup steps &  &  & 0 &  &  \\
Batch size &  &  & 4 &  &  \\
Max sequence length &  &  & 2048 &  &  \\
Training epochs &  &  & 3 &  &  \\
Training steps &  &  & 96 &  &  \\
Precision &  &  & fp32 &  &  \\ \bottomrule
\end{tabular}
}
\caption{Fine-tuning hyper-parameter settings of \textit{GPTDynamcis} for various tasks.}
\label{ap:hyperparam_ft}
\end{table*}

\paragraph{{\name} Training Setup}

We train a single featurized simulator on training runs for each dataset with the L2-regularized regression objective as defined in section~\ref{sec:featurized}. We freeze the parameters of the pre-trained encoder during training for better generalization. We set the order $n$ of Markov process assumptions equal to 1 for fine-tuning. Detailed training hyperparameters are shown in Table~\ref{ap:hyperparam_ft_training}.

\begin{table*}[]
\centering
\resizebox{0.9\linewidth}{!}{
\begin{tabular}{@{}lp{2.5cm}<{\centering}p{2.5cm}<{\centering}p{2.5cm}<{\centering}p{2.5cm}<{\centering}p{2.5cm}<{\centering}@{}}
\toprule
Hyperparameters & SST-2 & RTE & BoolQ & WebNLG & WMT16 DE/EN \\ \midrule
L2-regularizaiton's $\lambda$ &  &  & 1e-5 &  &  \\
Optimizer &  &  & AdamW &  &  \\
Adam’s $\beta$ &  &  & (0.9, 0.999) &  &  \\
Adam’s $\epsilon$ &  &  & 1e-8 &  &  \\
Learning rate & 1e-4 & 1e-4 & 1e-4 & 1e-4 & 1e-4 \\
Learning rate schedule &  &  & Linear decay &  &  \\
Warmup steps &  &  & 200 &  &  \\
Batch size &  &  & 128 &  &  \\
Max training epochs & 300 & 300 & 300 & 300 & 300 \\
Pre-trained encoder & MiniLM-L6-v2 & MiniLM-L6-v2 & MiniLM-L6-v2 & MiniLM-L6-v2 & MiniLM-L6-v2 \\
Max sequence length & 512 & 512 & 512 & 512 & 512 \\
Early stopping &  &  & \ding{51} &  &  \\
Precision &  &  & fp32 &  &  \\
Seed &  &  & 42 &  &  \\ \bottomrule
\end{tabular}
}
\caption{Hyperparameters of training our featurized \textbf{simulator} for each dataset for \textit{fine-tuning}. We use the same training hyperparameters as in the loss simulation for the BLEU and ROUGE-L score simulation on WebNLG and WMT16 DE/EN datasets.}
\label{ap:hyperparam_ft_training}
\end{table*}


\begin{table}[]
\tiny
\centering
\resizebox{0.8\linewidth}{!}{
\begin{tabular}{@{}lp{1.5cm}<{\centering}@{}}
\toprule
Hyperparameters &  \\ \midrule
L2 regularizaiton $\lambda$ & 1e-5  \\
Optimizer & AdamW   \\
Adam’s $\beta$ & (0.9, 0.999)  \\
Adam’s $\epsilon$ & 1e-8  \\
Learning rate & 1e-3  \\
Learning rate schedule & Linear decay  \\
Warmup steps & 200  \\
Batch size & 128  \\
Max training epochs & 300  \\
Early stopping & \ding{51}  \\
Precision & fp32  \\
Seed & 42  \\ \bottomrule
\end{tabular}
}
\caption{Training hyperparameters of Simfluence for \textit{fine-tuning}. It is noted that we use the same hyperparameters for both loss and metric simulation, as we see that different hyperparameters has little effect on Simfluence's performance.}
\end{table}

\subsection{Implementing {\name}}

\paragraph{{\name} Training} 
To elucidate the intricate process of collecting training data dynamics and the training of the featurized simulator with {\name}, we present the pseudo-code in Algorithm~\ref{algo:gptfluence}. The execution of this algorithm yields a {\name} simulator, which is adept at simulating the target performance trajectory and assessing the impact of training examples on a given test point.

\begin{algorithm}[!th]
\caption{{\name} Training Procedure}
    {\bf Input:} Language modeling task $P$, pre-trained GPT $\theta$, Target sample $z'$, Dataset $\mathcal{D}$, Subset size $I$, Target metric $\phi$, Training dynamic $D_{run}$, Multiplicative factor function $\alpha(\cdot)$, Additive factor function $\beta(\cdot)$, $L2$ regularization weight, Featurized simulator $\Theta$, Markov order $n$-th
    
    {\bf Output:} Simulator $\hat{\Theta}$
    \begin{algorithmic}[1] 
    \STATE Initialize $\mathcal{D}_{run}$ with an empty set
    \FOR{$k = 1$ to $K$}
        \STATE Sample a subset $D' \subset D$ of size $I$
        \FOR{Sample batch $c_t \in D'$}
            \STATE Update $\theta_{t}$ using $P$ based on $c_t$
            \STATE Calculate target metric $y_t^k = $ \\ $\phi(\theta_{t},z')$
            \STATE Add $c_t$ and $y^k_t$ into $D_{run}$
        \ENDFOR
    \ENDFOR
    \STATE Initialize $g_{\theta}$ with pre-trained encoder
    \WHILE{not converged}
        \STATE Sample a mini-batch $B_{train}, B_{test}$ from $D_{run}$
        \FOR{each $z_i \in B_{train}$}
            \STATE Compute multiplicative and additive influences $A_{i,1:n}$, $B_i$
        \ENDFOR
        \STATE $\alpha = \{\alpha_j(B_{train}) | j = 1, 2, ..., n\}$
        \STATE $\beta = \beta(B_{train})$
        \STATE Update  $\Theta$ with $\alpha$, $\beta$, $\gamma$
    \ENDWHILE
    \STATE \textbf{return} $\hat{\Theta}$
    \end{algorithmic}
    \label{algo:gptfluence}
\end{algorithm}

\paragraph{{\name} Evaluation} For evaluation, The simulator \textit{autoregressively} forecasts upcoming test-set metrics, based on the previous $n$ observations. Specifically, it commences with the initial test metric recorded at the starting step, thereafter predicting the subsequent performance metrics across the training curriculum.

\section{Experiment Results}
In this section, we provide additional experimental results and detailed descriptions to complement the main findings.

\label{ap:results}

\subsection{Empirical Analysis on Markov Property}

Table~\ref{tabel:markov} presents a comprehensive results of how the order of the Markov process influences test loss, BLEU, and ROUGE-L metrics during instruction tuning simulations.

\begin{table}[]
\resizebox{\linewidth}{!}{
\begin{tabular}{llp{2cm}p{2cm}p{2.5cm}}
\hline
\textbf{Task} & \textbf{Order} & All-steps MSE ($\downarrow$) & All-steps MAE ($\downarrow$) & Final-step Spearman’s $\rho$ ($\uparrow$)\\ \hline
RTE & 1 & 0.036(0.029) & 0.151(0.060) & 0.746(0.095) \\
 & 2 & 0.036(0.029) & 0.151(0.060) & 0.747(0.094) \\
 & 3 & 0.036(0.030) & 0.149(0.062) & 0.750(0.094) \\
 & 5 & 0.037(0.032) & 0.147(0.067) & 0.757(0.093) \\
 & 10 & 0.036(0.032) & 0.150(0.071) & 0.785(0.088) \\ \hline
\textbf{Task} & \textbf{Order} & All-steps MSE ($\downarrow$) & All-steps MAE ($\downarrow$) & Final-step Spearman’s $\rho$ ($\uparrow$) \\ \hline
WEBNLG & 1 & 0.011(0.014) & 0.078(0.043) & 0.985(0.002) \\
 & 2 & 0.010(0.011) & 0.072(0.039) & 0.986(0.002) \\
 & 3 & 0.011(0.012) & 0.073(0.040) & 0.986(0.002) \\
 & 5 & 0.012(0.014) & 0.082(0.044) & 0.983(0.003) \\
 & 10 & 0.012(0.014) & 0.082(0.044) & 0.983(0.002) \\ \hline
\multirow{2}{*}{\textbf{Task}} & \multirow{2}{*}{\textbf{Order}} & \multicolumn{3}{c}{\textbf{BLEU}} \\ \cline{3-5} 
 &  & All-steps MSE ($\downarrow$) & All-steps MAE ($\downarrow$) & Final-step Spearman’s $\rho$ ($\uparrow$) \\ \hline
WEBNLG & 1 & 43.98(81.40) & 4.28(3.57) & 0.80(0.01) \\
 & 2 & 43.31(80.70) & 4.24(3.54) & 0.80(0.03) \\
 & 3 & 43.67(81.77) & 4.24(3.57) & 0.80(0.02) \\
 & 5 & 44.79(76.57) & 4.39(3.49) & 0.78(0.02) \\
 & 10 & 47.83(99.06) & 4.35(3.72) & 0.74(0.03) \\ \hline
\multirow{2}{*}{\textbf{Task}} & \multirow{2}{*}{\textbf{Order}} & \multicolumn{3}{c}{\textbf{ROUGE-L}} \\ \cline{3-5} 
 &  & All-steps MSE ($\downarrow$) & All-steps MAE ($\downarrow$) & Final-step Spearman’s $\rho$ ($\uparrow$) \\ \hline
WEBNLG & 1 & 0.01(0.01) & 0.06(0.03) & 0.77(0.04) \\
 & 2 & 0.01(0.01) & 0.06(0.03) & 0.76(0.04) \\
 & 3 & 0.01(0.01) & 0.06(0.03) & 0.76(0.03) \\
 & 5 & 0.01(0.01) & 0.06(0.03) & 0.77(0.04) \\
 & 10 & 0.01(0.01) & 0.06(0.03) & 0.78(0.03) \\ \hline
\end{tabular}
}
\caption{Impact of the Markov process order on test loss, BLEU, and ROUGE-L metrics in \textit{instruction tuning} simulations.}
\label{tabel:markov}
\end{table}

\subsection{Unseen Data Generalization}
\begin{figure}[ht]
    \centering
    \subfigure[RTE]{
        \includegraphics[width=0.9\columnwidth]{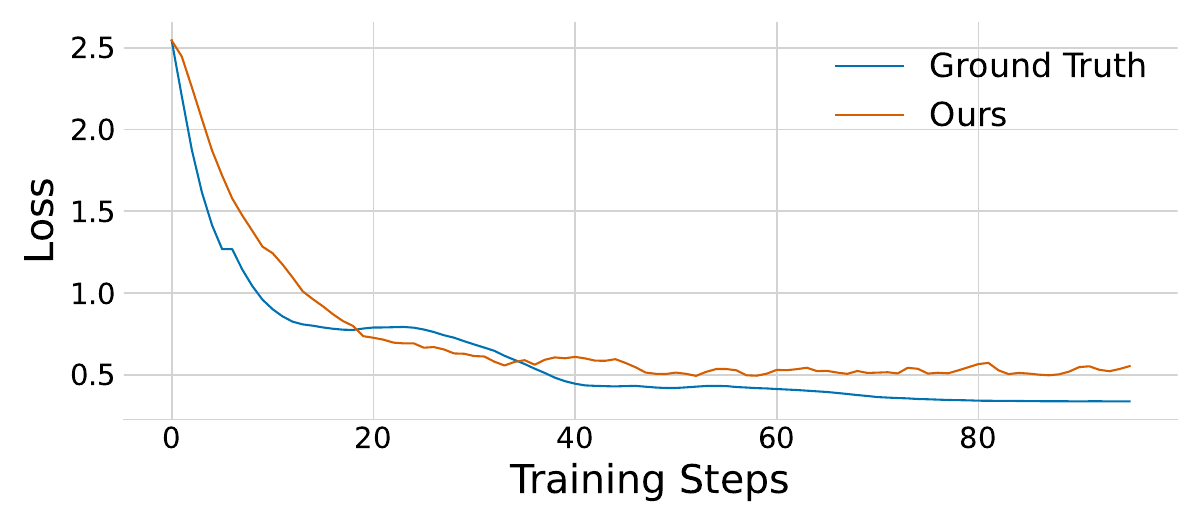}
    }
    \subfigure[WebNLG]{
        \includegraphics[width=0.9\columnwidth]{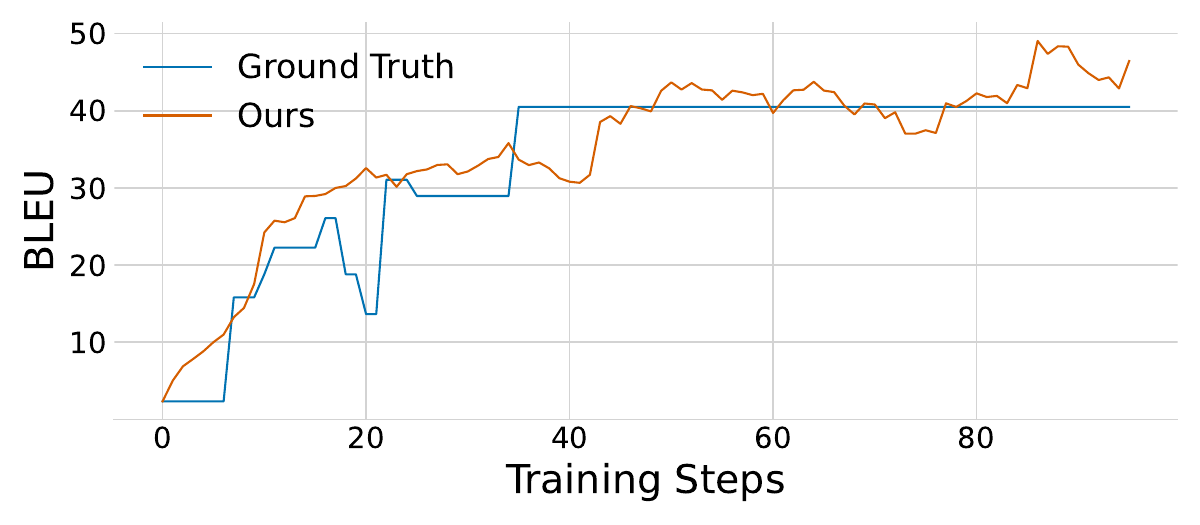}
    }
    \vspace{-4mm}
    \caption{Illustration of simulation results on \textbf{\textit{unseen training data}}. The \textit{top} shows the loss simulation on RTE, while the \textit{bottom} shows the BLEU metric simulation for WebNLG. Additional qualitative examples for different settings and metrics are provided in  \S~\ref{ap:examples_unseen}.}
    \label{fig:unsen_ft}
    \vspace{-3mm}
\end{figure}

We offer in-depth simulations of loss and performance metrics across scenarios involving unseen training data, unseen test data, and both unseen training and test data. Simulation results for fine-tuning are detailed in Table~\ref{tab:unseen_ft}, while those for instruction-tuning can be found in Table~\ref{tab:unseen_it}. Illustration examples are shown in the Fig.~\ref{fig:unsen_ft}.

\begin{table}[t]
    \begin{minipage}{\columnwidth}    
        \resizebox{\linewidth}{!}{
            \begin{tabular}{@{}llp{2.3cm}p{1.6cm}p{2.5cm}p{2cm}p{2.5cm}@{}}
            \toprule
            \textbf{Task} & \textbf{Metrics} & \textbf{Training Data Unseen} & \textbf{Test Data Unseen} & All-steps MSE & All-steps MAE & Final-Step Spearman’s $\rho$ \\ \midrule
            \multirow{3}{*}{RTE} & \multirow{3}{*}{\textbf{Loss}} & \multicolumn{1}{c}{\ding{51}} & \multicolumn{1}{c}{\ding{55}} & 0.346(0.281) & 0.513(0.211) & 0.913(0.052) \\
             &  & \multicolumn{1}{c}{\ding{55}} & \multicolumn{1}{c}{\ding{51}} & 0.351(0.489) & 0.444(0.325) & -0.024(0.050) \\
             &  & \multicolumn{1}{c}{\ding{51}} & \multicolumn{1}{c}{\ding{51}} & 0.984(4.569) & 0.568(0.728) & -0.048(0.045) \\ \midrule
            \multirow{9}{*}{WEBNLG} & \multirow{3}{*}{\textbf{Loss}} & \multicolumn{1}{c}{\ding{51}} & \multicolumn{1}{c}{\ding{55}} & 1.251(0.962) & 1.003(0.413) & 0.892(0.011) \\
             &  & \multicolumn{1}{c}{\ding{55}} & \multicolumn{1}{c}{\ding{51}} & 0.403(0.575) & 0.476(0.476) & 0.123(0.019) \\
             &  & \multicolumn{1}{c}{\ding{51}} & \multicolumn{1}{c}{\ding{51}} & 0.886(2.112) & 0.699(0.549) & 0.190(0.013) \\ \cmidrule(l){2-7} 
             & \multirow{3}{*}{\textbf{BLEU}} & \multicolumn{1}{c}{\ding{51}} & \multicolumn{1}{c}{\ding{55}} & 94.99(273.96) & 6.13(5.72) & 0.51(0.02) \\
             &  & \multicolumn{1}{c}{\ding{55}} & \multicolumn{1}{c}{\ding{51}} & 106.19(150.51) & 7.14(5.02) & 0.18(0.08) \\
             &  & \multicolumn{1}{c}{\ding{51}} & \multicolumn{1}{c}{\ding{51}} & 153.63(219.29) & 8.66(6.39) & 0.15(0.01) \\ \cmidrule(l){2-7} 
             & \multirow{3}{*}{\textbf{ROUGE-L}} & \multicolumn{1}{c}{\ding{51}} & \multicolumn{1}{c}{\ding{55}} & 0.008(0.009) & 0.069(0.036) & 0.578(0.062) \\
             &  & \multicolumn{1}{c}{\ding{55}}& \multicolumn{1}{c}{\ding{51}} & 0.009(0.008) & 0.073(0.034) & 0.288(0.049) \\
             &  & \multicolumn{1}{c}{\ding{51}} & \multicolumn{1}{c}{\ding{51}} & 0.010(0.010) & 0.075(0.039) & 0.168(0.091) \\ \bottomrule
            \end{tabular}
    }
    \caption{Results of loss and metric simulation on unseen data for RTE and WebNLG Datasets for \textit{fine-tuning}.}
    \label{tab:unseen_ft}
    \end{minipage}
    \\[12pt]
    \begin{minipage}{\columnwidth}
        \resizebox{\linewidth}{!}{
\begin{tabular}{llp{2cm}p{1.3cm}p{2cm}p{2cm}p{2.5cm}}
\hline
\textbf{Task} & \textbf{Metrics} & \textbf{Training Set OOD} & \textbf{Test Set OOD} & All-steps MSE ($\downarrow$) & All-steps MAE ($\downarrow$) & Final-step Spearman’s $\rho$ ($\uparrow$) \\ \hline
\multirow{3}{*}{RTE} & \multirow{3}{*}{\textbf{Loss}} & \multicolumn{1}{c}{\ding{51}} & \multicolumn{1}{c}{\ding{55}} & 0.781(0.793) & 0.730(0.419) & -0.082(0.214) \\
 &  & \multicolumn{1}{c}{\ding{55}} & \multicolumn{1}{c}{\ding{51}} & 1.137(2.927) & 0.725(0.619) & -0.011(0.033) \\
 &  & \multicolumn{1}{c}{\ding{51}} & \multicolumn{1}{c}{\ding{51}} & 1.110(1.057) & 0.888(0.482) & -0.047(0.062) \\ \hline
\multirow{9}{*}{WEBNLG} & \multirow{3}{*}{\textbf{Loss}} & \multicolumn{1}{c}{\ding{51}} & \multicolumn{1}{c}{\ding{55}} & 2.398(1.722) & 1.435(0.508) & 0.358(0.006) \\
 &  & \multicolumn{1}{c}{\ding{55}} & \multicolumn{1}{c}{\ding{51}} & 22.627(203.637) & 1.530(3.062) & 0.247(0.008) \\
 &  & \multicolumn{1}{c}{\ding{51}} & \multicolumn{1}{c}{\ding{51}} & 2.708(1.415) & 1.580(0.432) & 0.072(0.003) \\ \cline{2-7} 
 & \multirow{3}{*}{\textbf{BLEU}} & \multicolumn{1}{c}{\ding{51}} & \multicolumn{1}{c}{\ding{55}} & 200.50(270.92) & 10.82(7.08) & 0.34(0.06) \\
 &  & \multicolumn{1}{c}{\ding{55}} & \multicolumn{1}{c}{\ding{51}} & 115.19(188.30) & 7.33(5.69) & -0.03(0.03) \\
 &  & \multicolumn{1}{c}{\ding{51}} & \multicolumn{1}{c}{\ding{51}} & 329.61(369.14) & 14.20(8.11) & 0.10(0.04) \\ \cline{2-7} 
 & \multirow{3}{*}{\textbf{ROUGE-L}} & \multicolumn{1}{c}{\ding{51}} & \multicolumn{1}{c}{\ding{55}} & 0.12(0.06) & 0.33(0.08) & 0.35(0.02) \\
 &  & \multicolumn{1}{c}{\ding{55}} & \multicolumn{1}{c}{\ding{51}} & 0.01(0.02) & 0.09(0.05) & 0.06(0.06) \\
 &  & \multicolumn{1}{c}{\ding{51}} & \multicolumn{1}{c}{\ding{51}} & 0.13(0.05) & 0.35(0.06) & 0.10(0.01) \\ \hline
\end{tabular}
}   
\caption{Results of loss and metric simulation on unseen data for RTE and WebNLG Datasets for \textit{instruction tuning}.}
\label{tab:unseen_it}
    \end{minipage}

\end{table}

\section{Computational Complexity}
\begin{figure}[!ht]
    \centering
    \begin{minipage}[]{\columnwidth}
        \centering
        \tiny
        \resizebox{0.7\linewidth}{!}{
        \begin{tabular}{@{}lp{1cm}l@{}}
            \toprule
            \textbf{Method} & \textbf{Latency (sec/sample)} & \textbf{FLOPs}\\ \midrule
            TracIn-CP & \hfill 153.0 & 1.1$\times$10\textsuperscript{13} \\
            Simfluence & \hfill 0.1 & 
            1.6$\times$10\textsuperscript{1} \\ \midrule
            Ours & \hfill 0.2 & 5.3$\times$10\textsuperscript{6} \\ \bottomrule
        \end{tabular}
        }
        \captionof{table}{Inference latency and FLOPs of GPTfluence, Simfluence, and TracIn-CP.}
        \label{tab:latency}
        \vspace{1em}
    \end{minipage}
    \\[12pt]
    \begin{minipage}[]{0.6\textwidth}
        \includegraphics[width=0.4\linewidth]{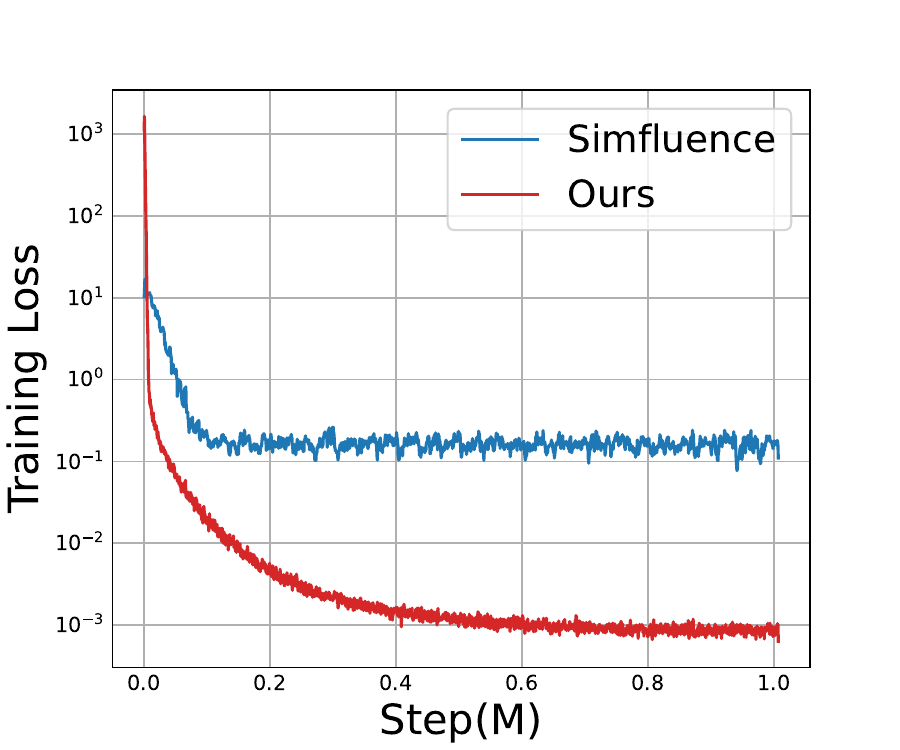}
        \includegraphics[width=0.4\linewidth]{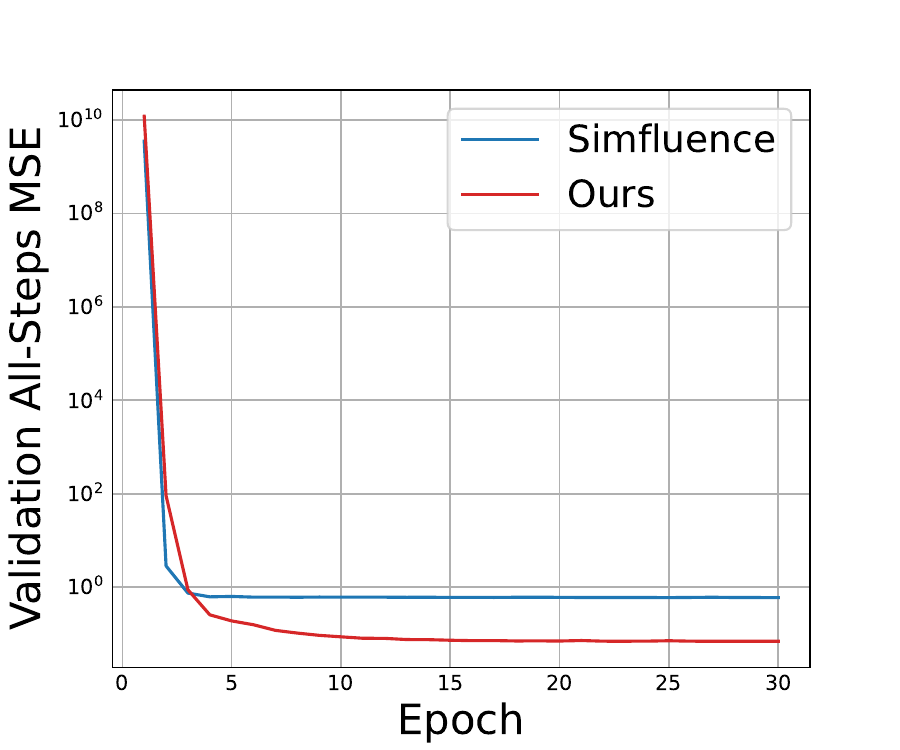}
    \end{minipage}
    \hfill
    \caption{Comparison of our method and Simfluence with respect to \textbf{training loss} (Left) and \textbf{validation all-steps MSE} (Right).}
    \label{fig:train}
\end{figure}

We conducted a comparison of inference latency and floating point operations (FLOPs) among various TDA methods. Results are presented in Table~\ref{tab:latency}. TracIn-CP, a representative of gradient-based methods, exhibited the highest inference latency and FLOPs. This is attributable to the need to do forward and backward operations directly on the GPTs. Conversely, {\name} solely depends on a considerably smaller simulator during inference.

Furthermore, we analyzed the convergence and validation performance of our {\name} in comparison with Simfluence. As shown in Fig.~\ref{fig:train}, {\name} exhibits a better convergence efficiency and also has lower validation all-steps MSE. This underscores the better training efficiency and model capacity of our featurized simulator.

\section{Qualitative Examples}
\label{ap:examples}
In this section, we provide additional quantitative examples, including loss and metric simulations, for a comparison. This includes experimental results across various training scenarios and the use of unseen data, among others.

\subsection{Simulation For Instruction-Tuning}
\label{ap:examples_it}

\begin{figure*}
    \centering
    \subfigure[BoolQ]{
        \includegraphics[width=0.3\linewidth]{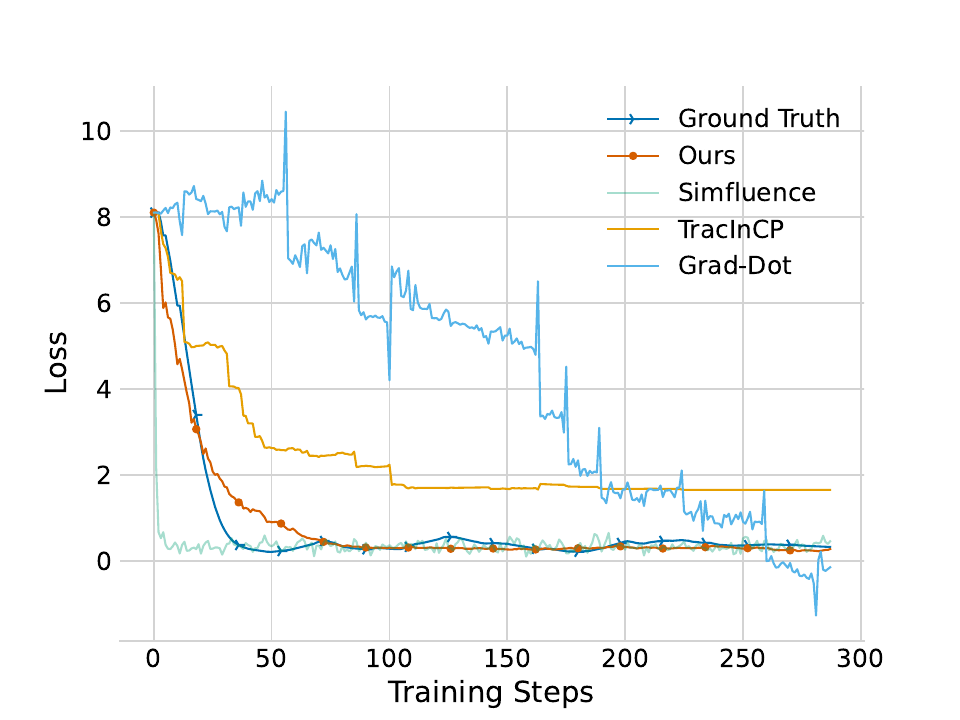}
        \includegraphics[width=0.3\linewidth]{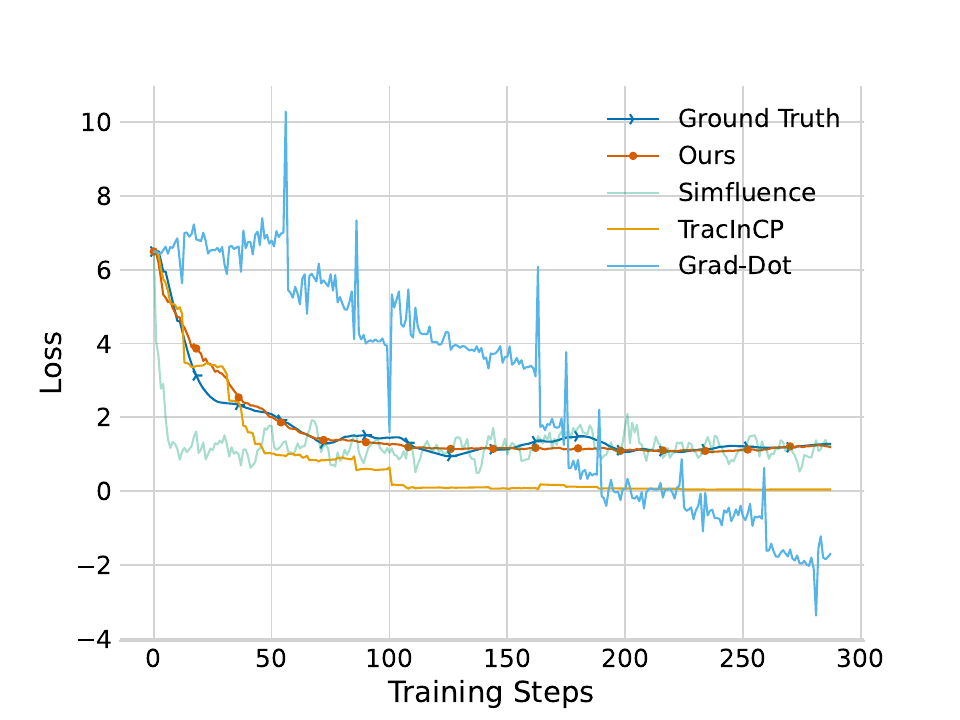}
        \includegraphics[width=0.3\linewidth]{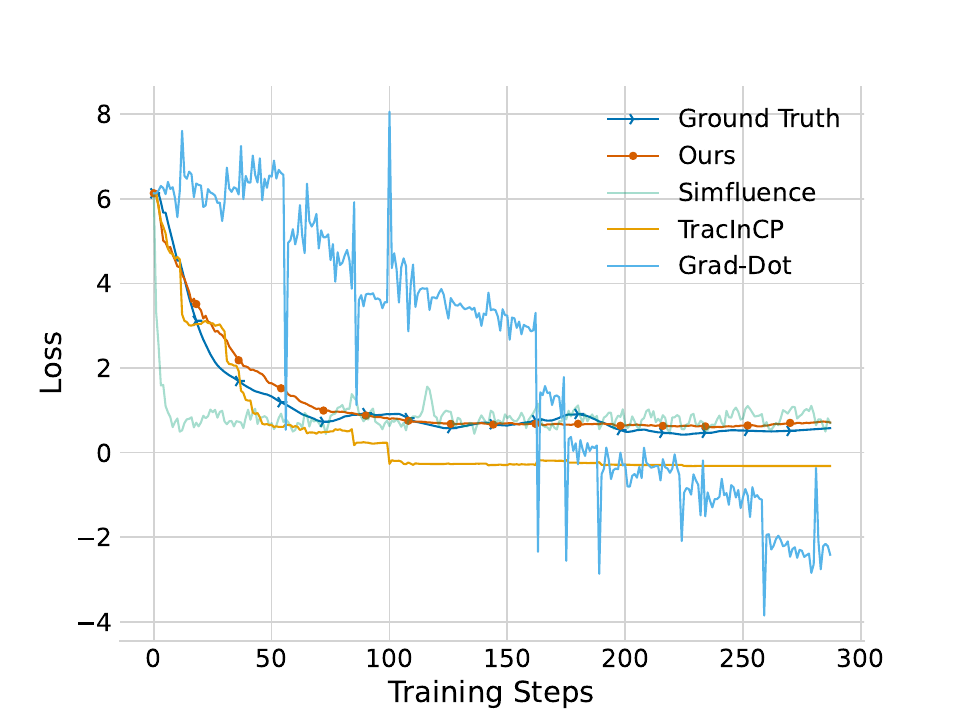}
    }
    
    \subfigure[RTE]{
        \includegraphics[width=0.3\linewidth]{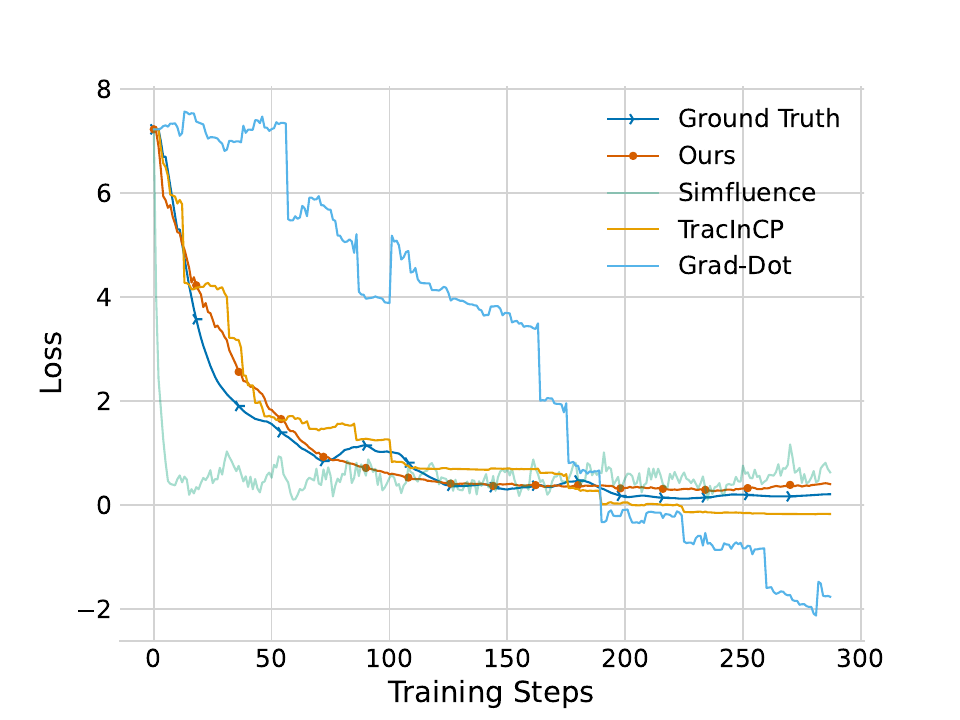}
        \includegraphics[width=0.3\linewidth]{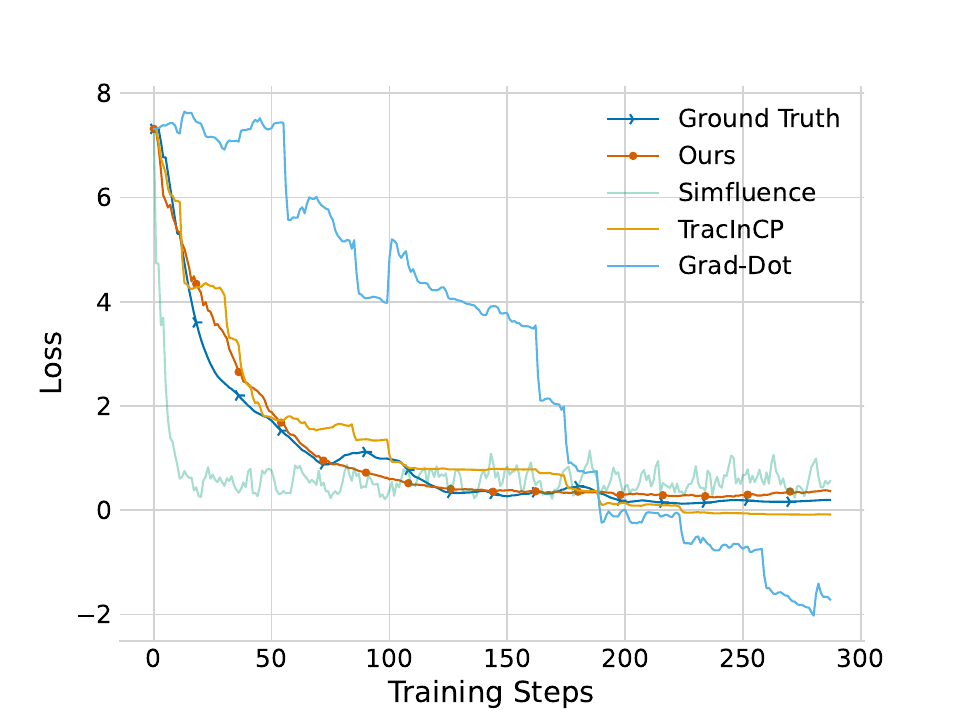}
        \includegraphics[width=0.3\linewidth]{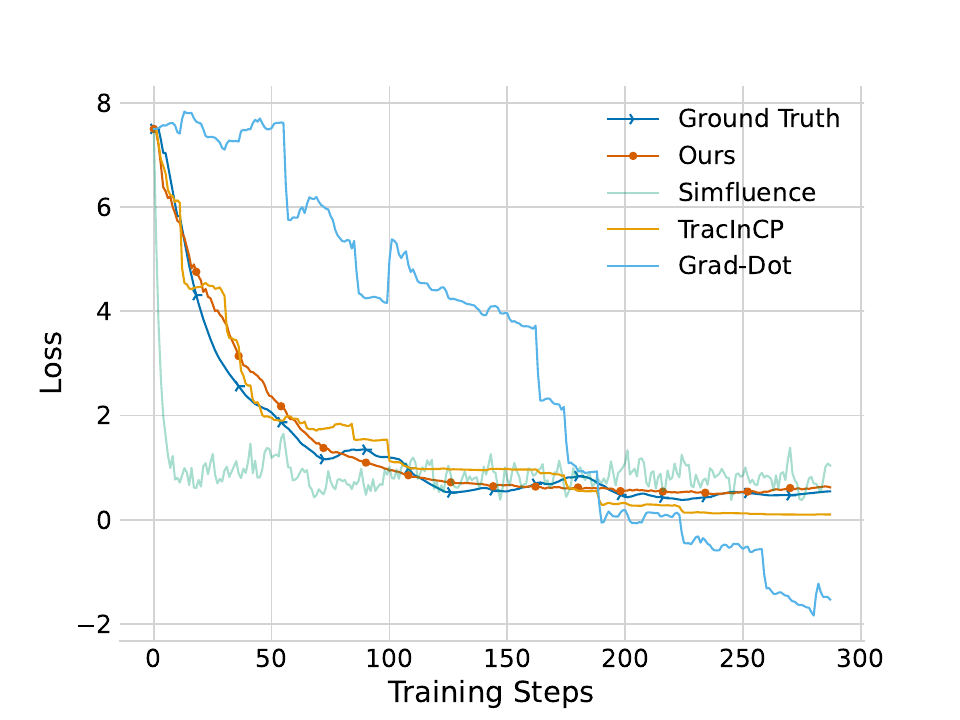}
    }
    
    \subfigure[SST-2]{
        \includegraphics[width=0.3\linewidth]{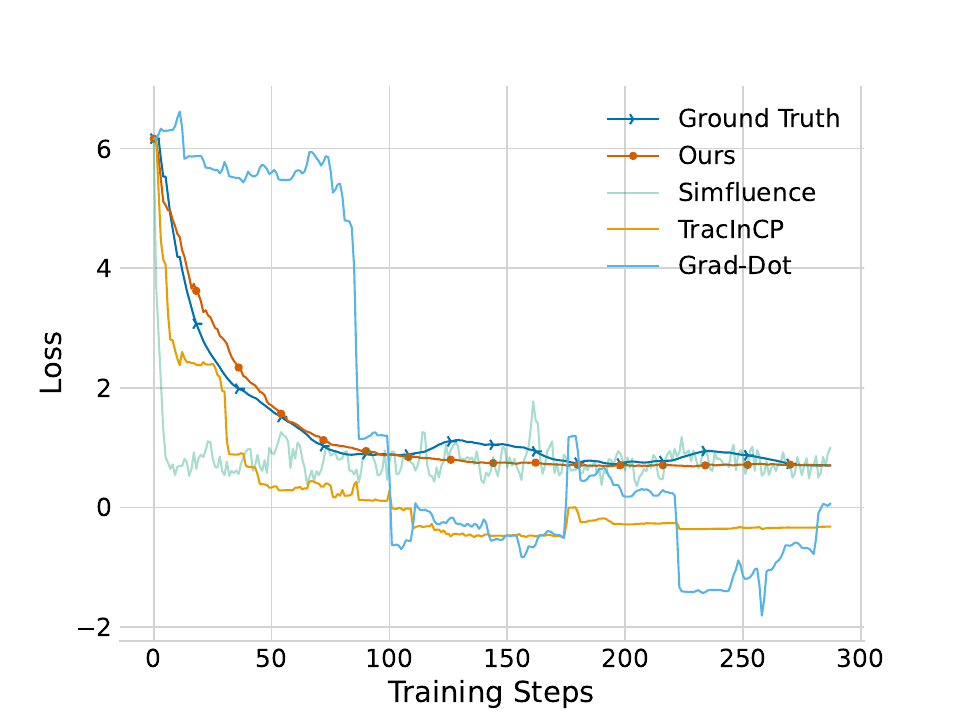}
        \includegraphics[width=0.3\linewidth]{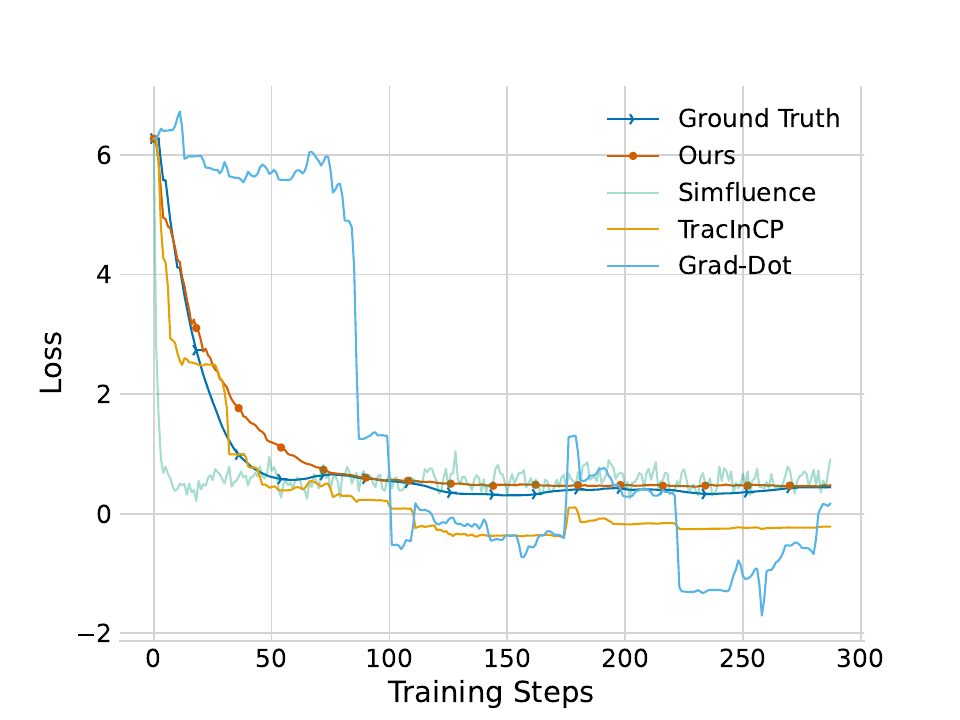}
        \includegraphics[width=0.3\linewidth]{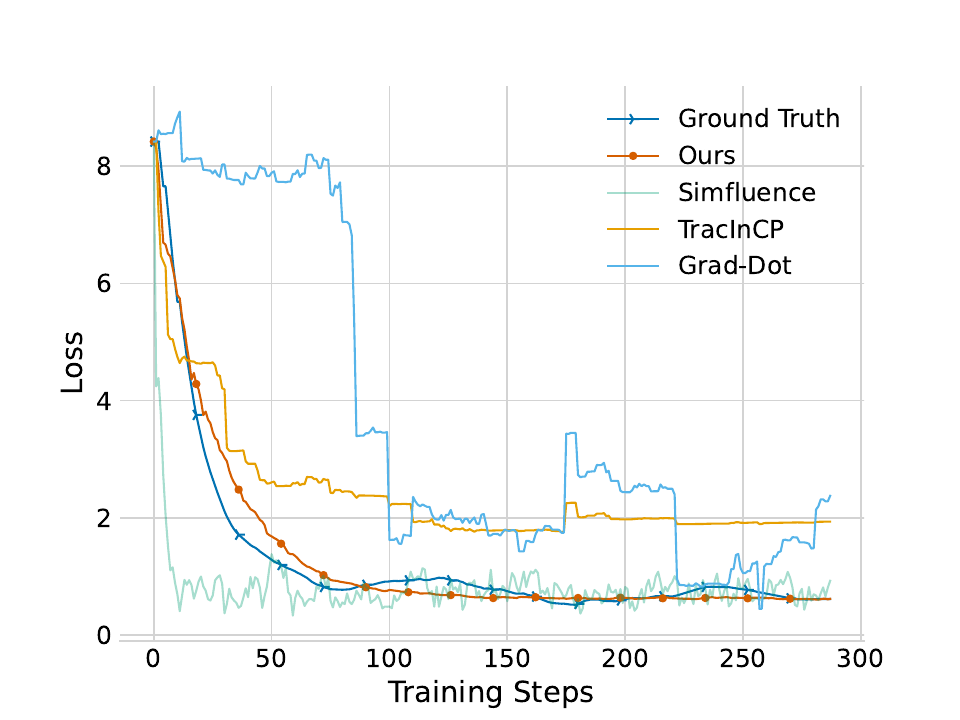}
    }

    \subfigure[WebNLG]{
        \includegraphics[width=0.3\linewidth]{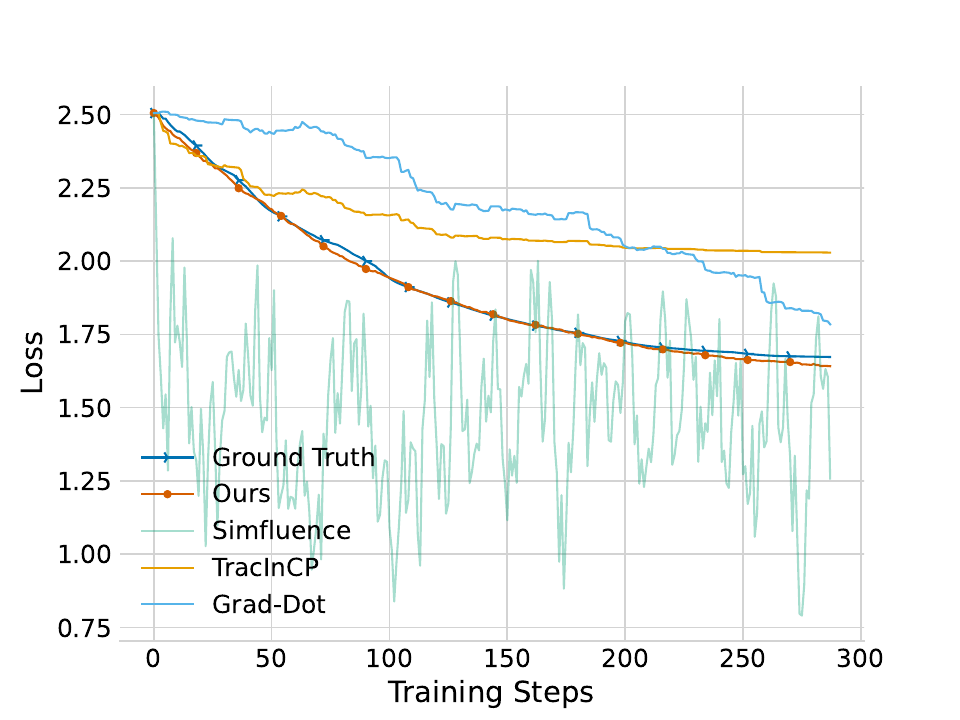}
        \includegraphics[width=0.3\linewidth]{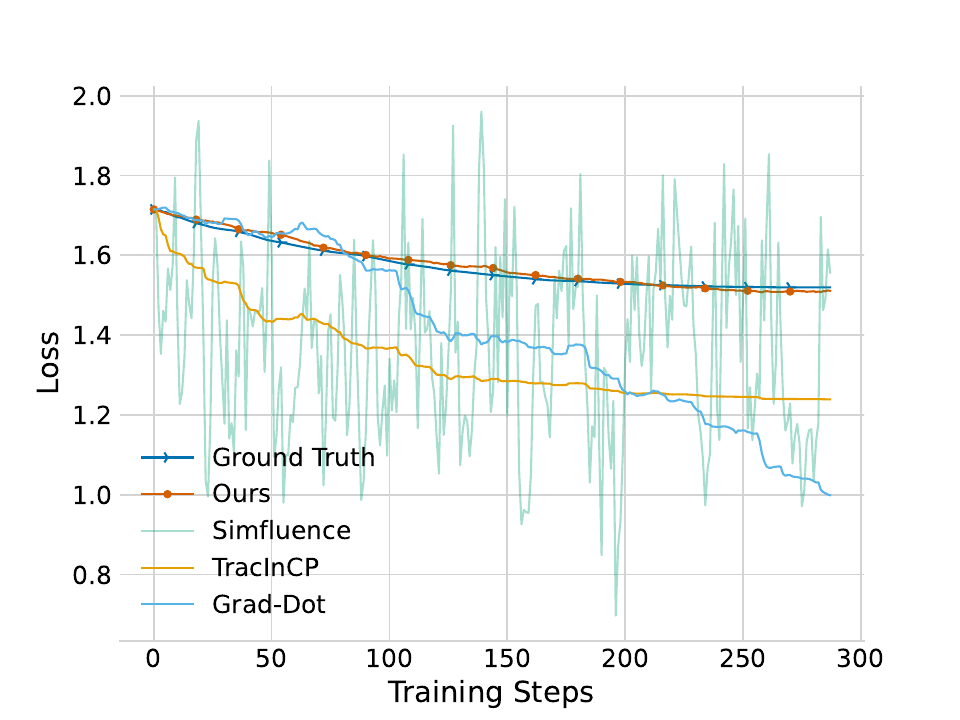}
        \includegraphics[width=0.3\linewidth]{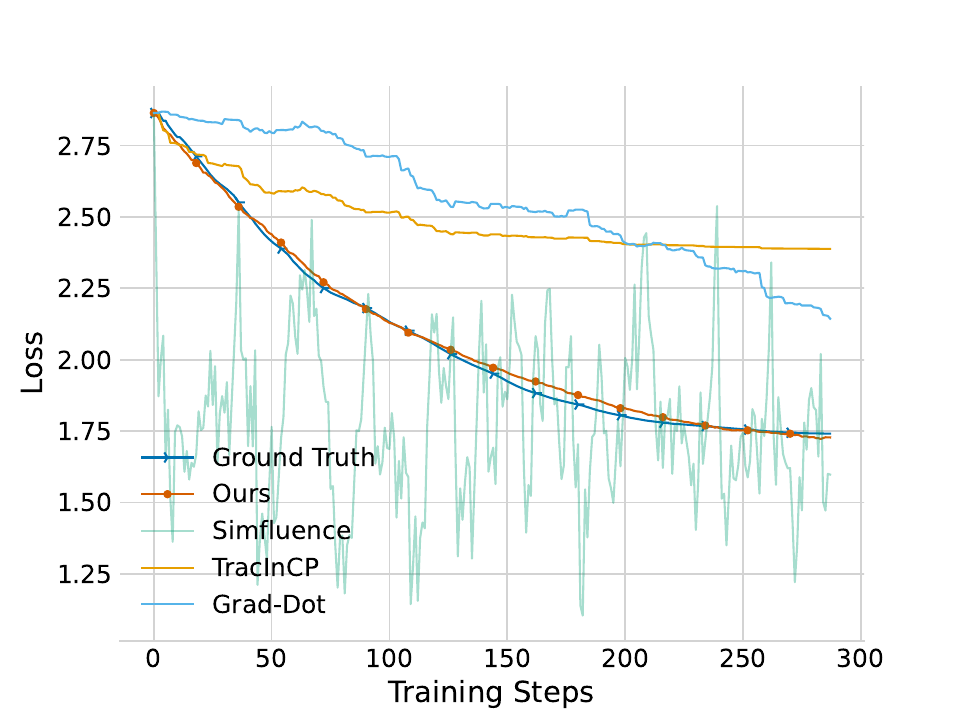}
    }

    \subfigure[WMT16 DE/EN]{
        \includegraphics[width=0.3\linewidth]{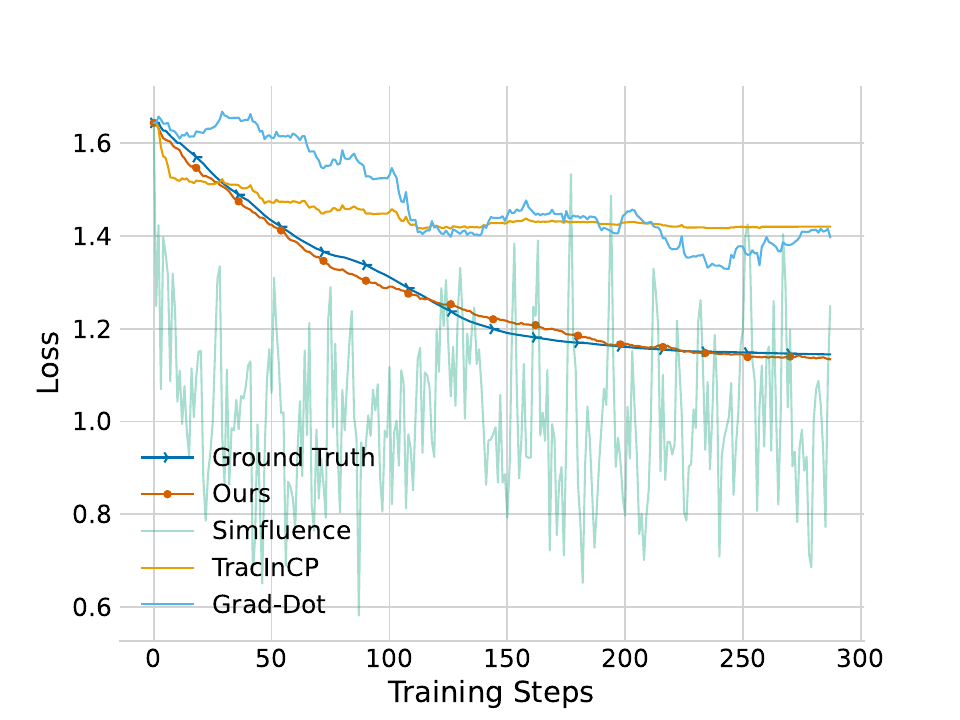}
        \includegraphics[width=0.3\linewidth]{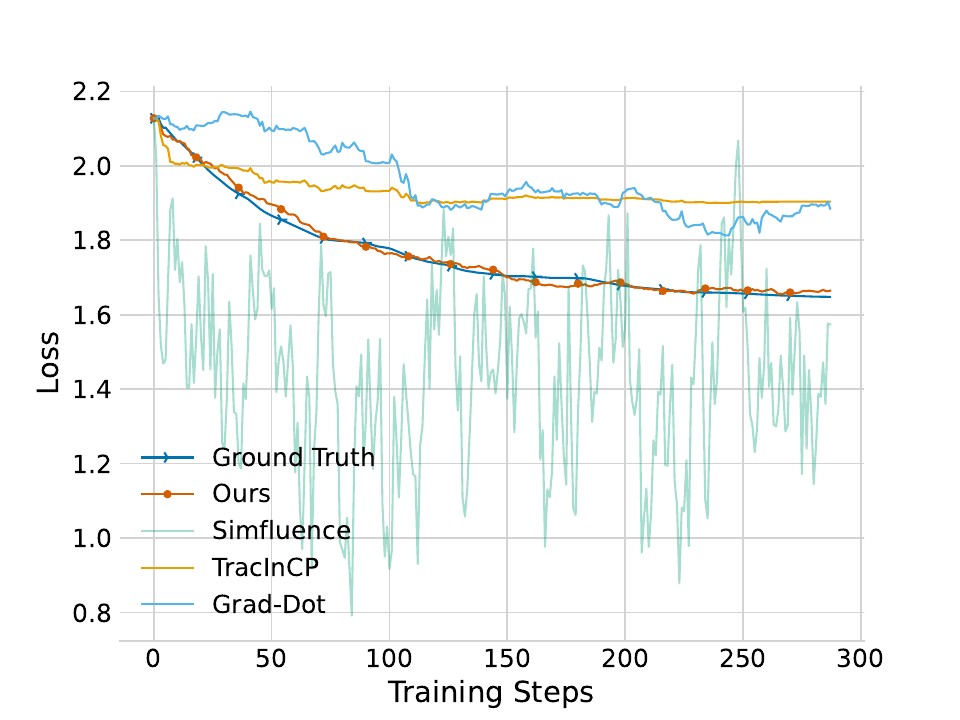}
        \includegraphics[width=0.3\linewidth]{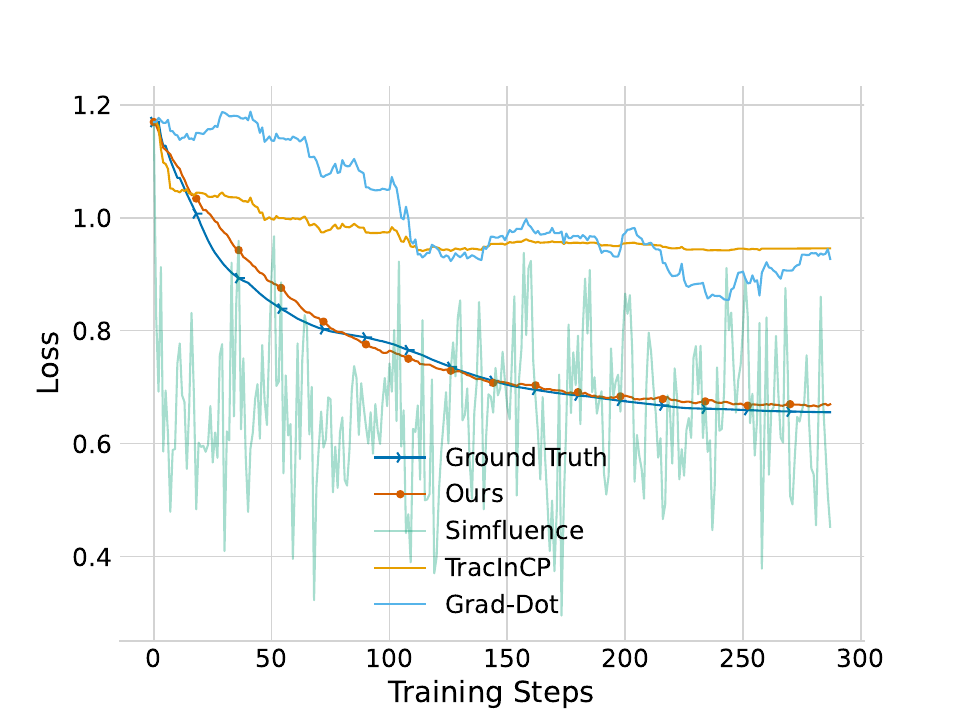}
    }
    \caption{Comparative loss simulations for \textit{instruction tuning} using {\name} versus other TDA methods on Pythia-410M across the BoolQ, RTE, SST-2, WebNLG, and WMT16 DE/EN datasets.}
    \label{fig:exm_it_loss_410}
\end{figure*}

\begin{figure*}
    \centering
    \subfigure[BoolQ]{
        \includegraphics[width=0.3\linewidth]{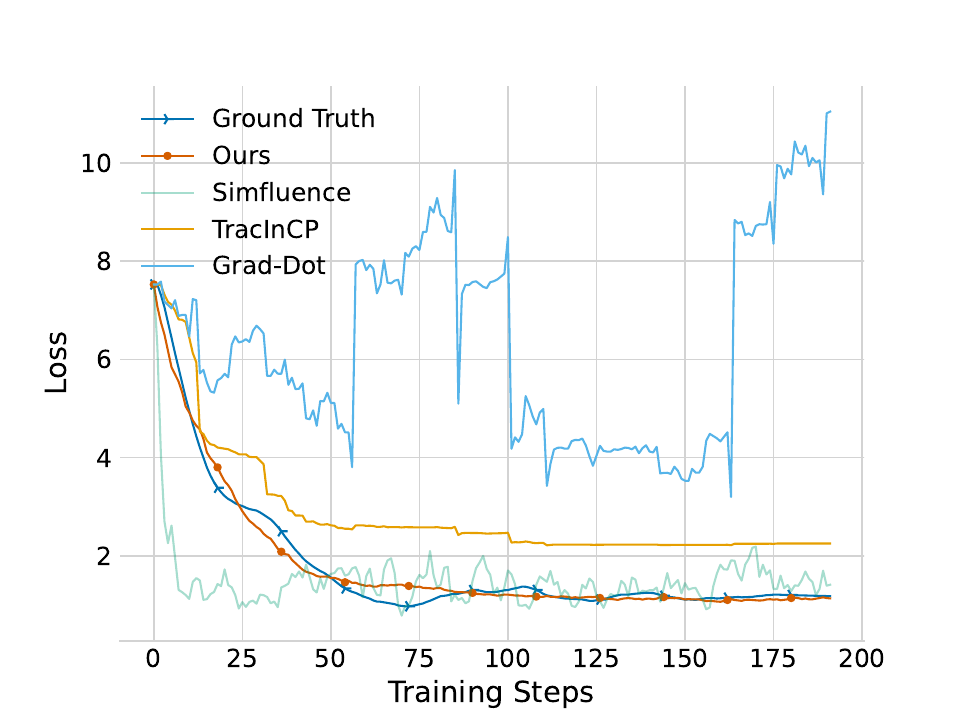}
        \includegraphics[width=0.3\linewidth]{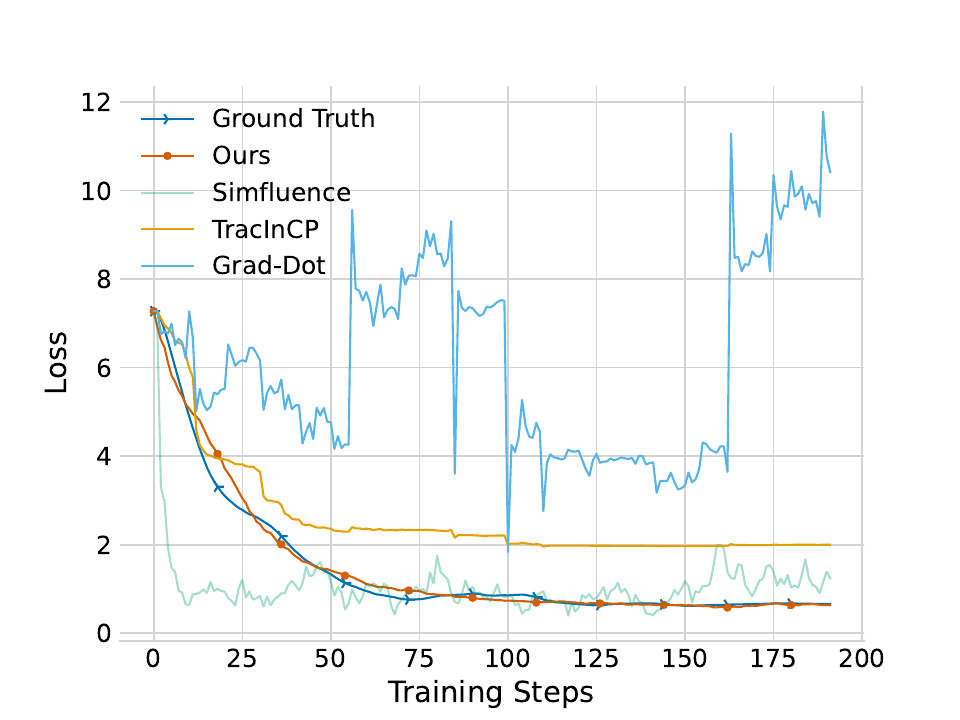}
        \includegraphics[width=0.3\linewidth]{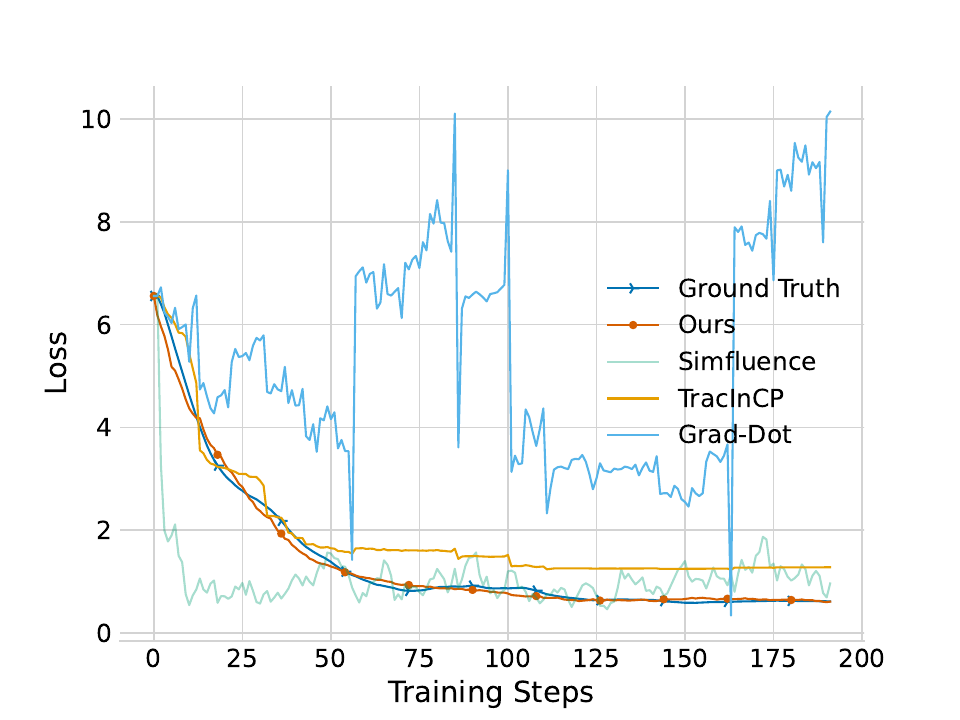}
    }
    
    \subfigure[RTE]{
        \includegraphics[width=0.3\linewidth]{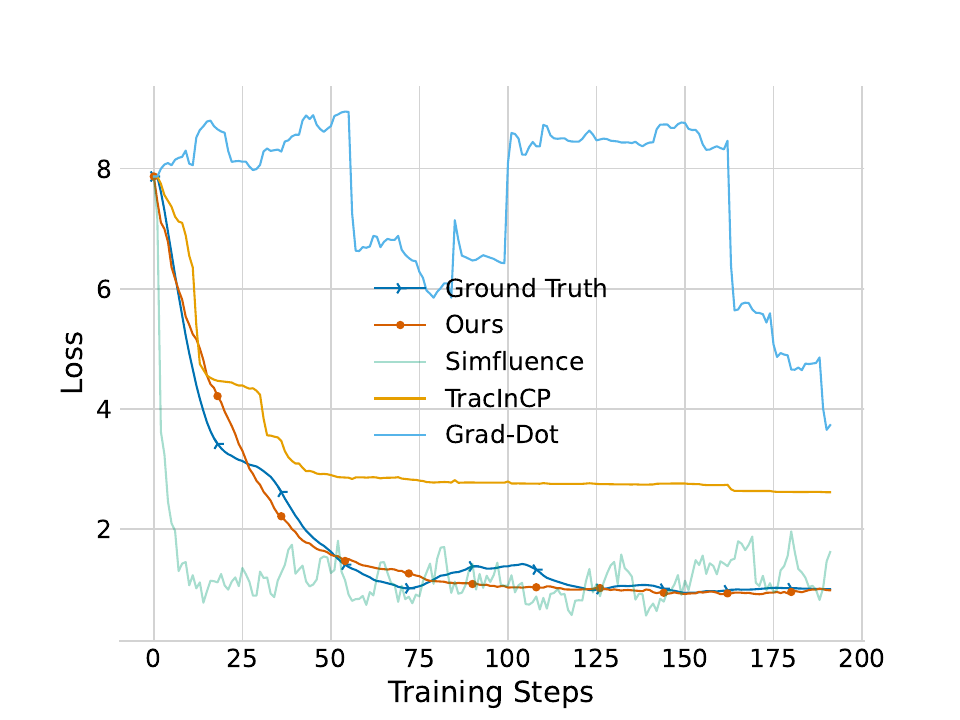}
        \includegraphics[width=0.3\linewidth]{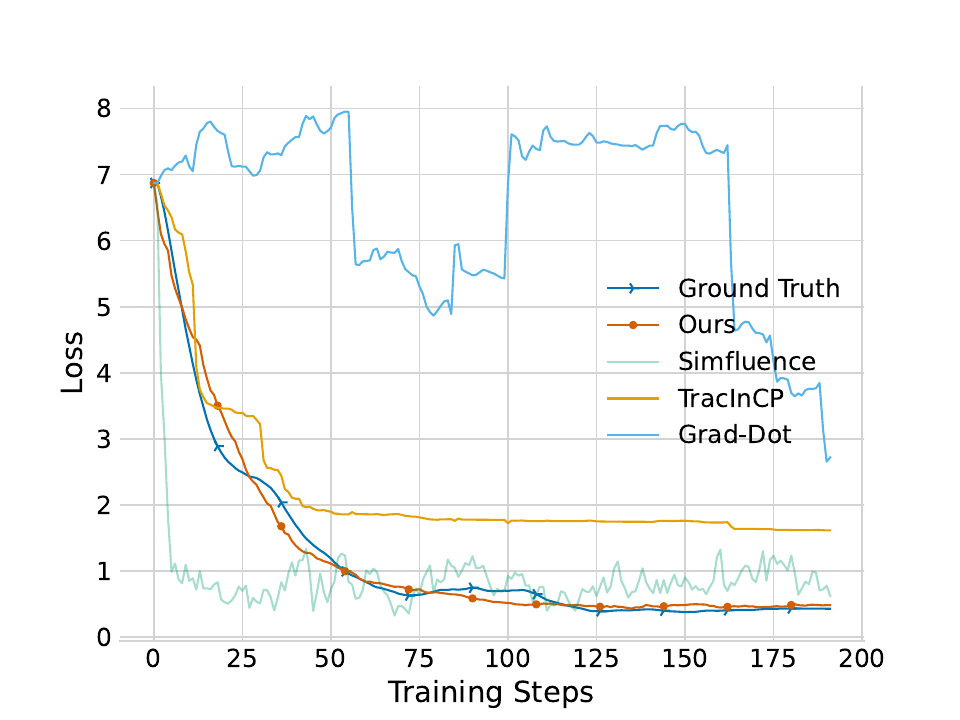}
        \includegraphics[width=0.3\linewidth]{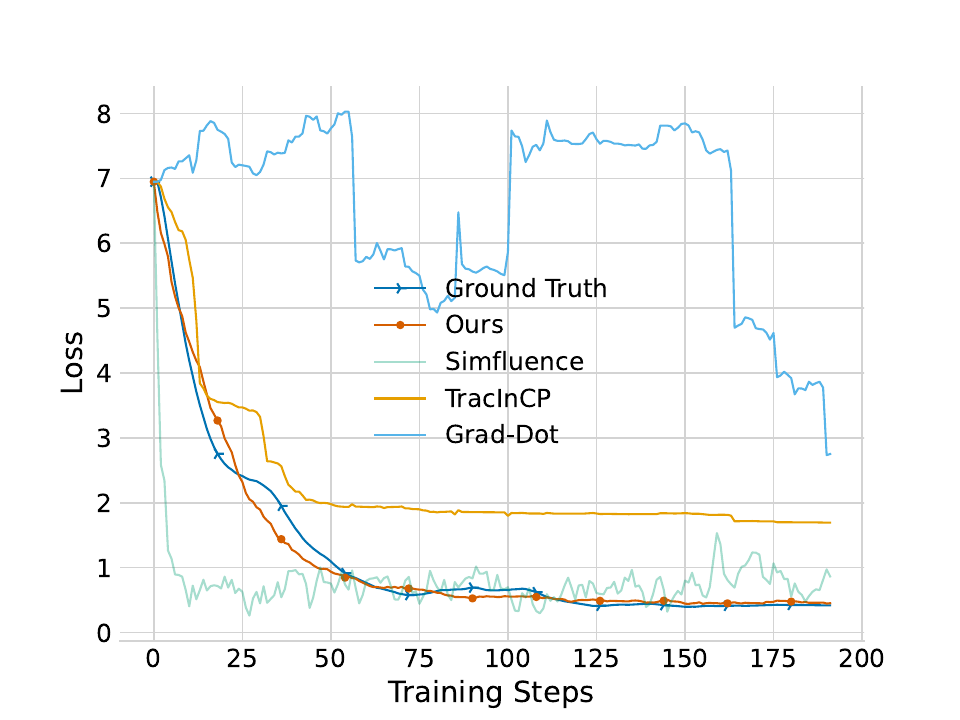}
    }
    
    \subfigure[SST-2]{
        \includegraphics[width=0.3\linewidth]{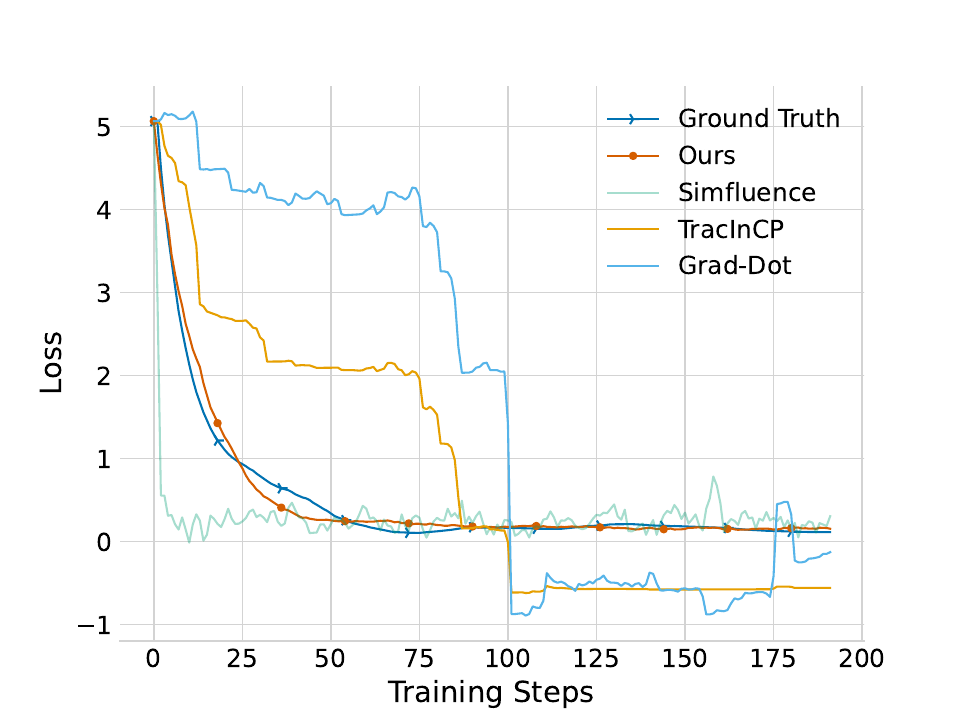}
        \includegraphics[width=0.3\linewidth]{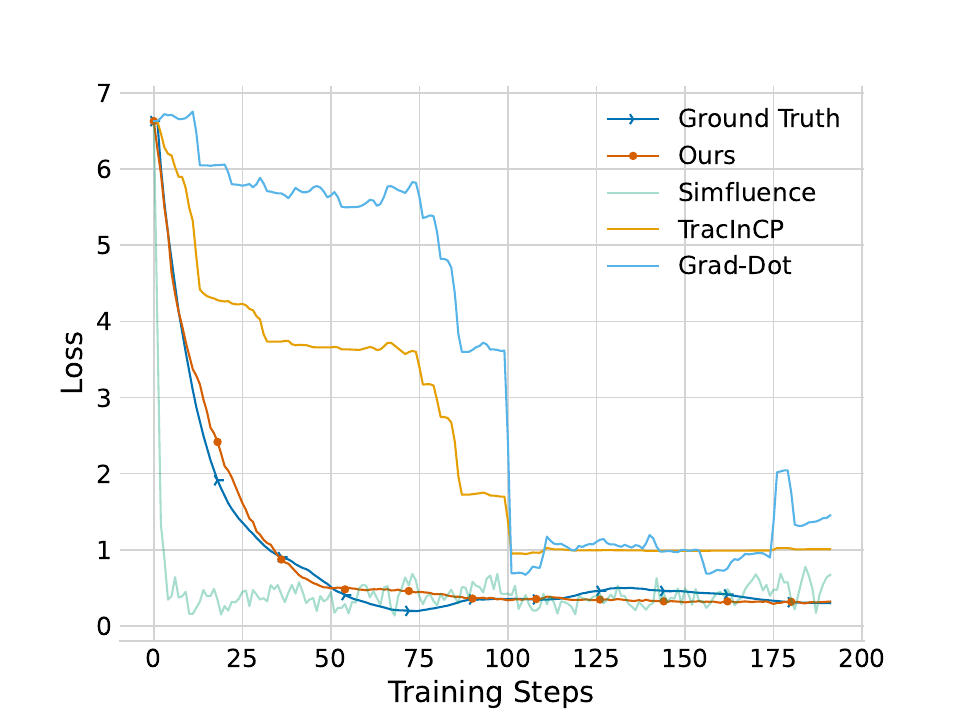}
        \includegraphics[width=0.3\linewidth]{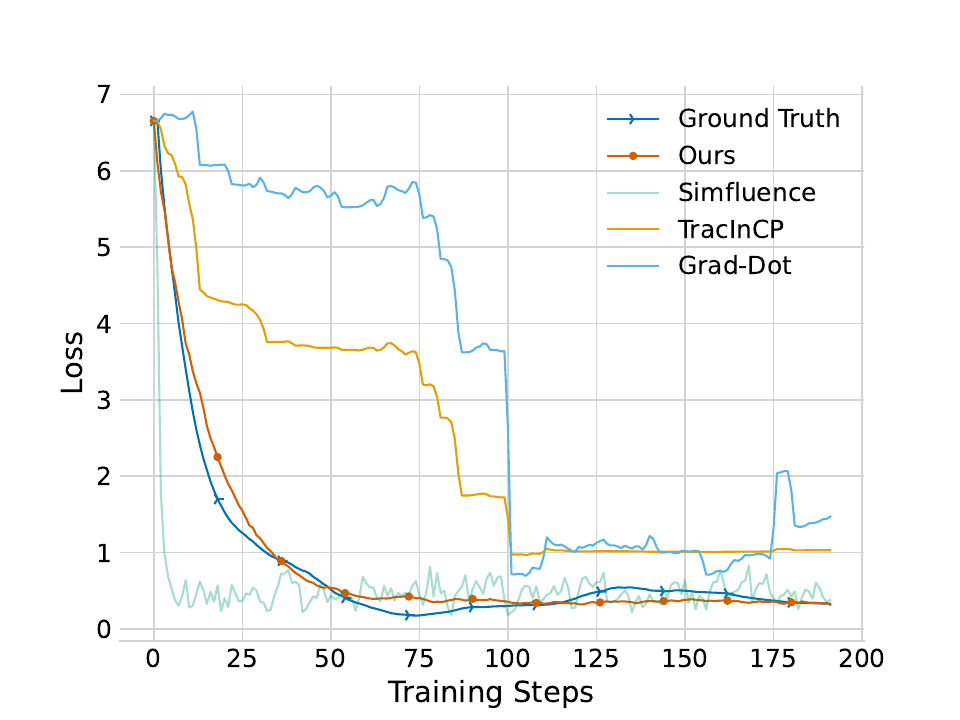}
    }

    \subfigure[WebNLG]{
        \includegraphics[width=0.3\linewidth]{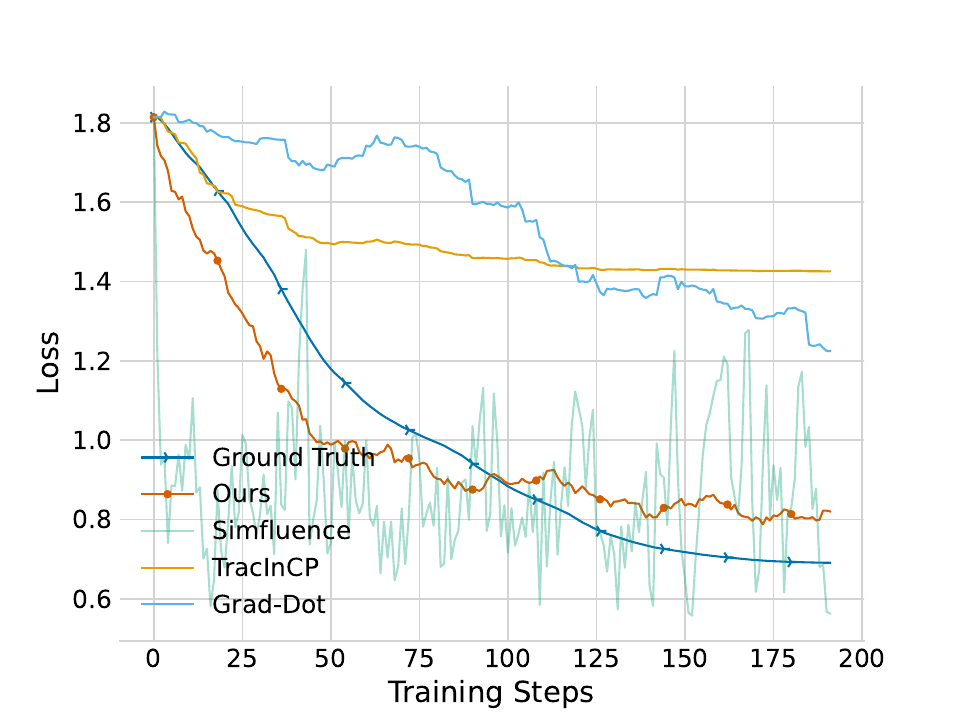}
        \includegraphics[width=0.3\linewidth]{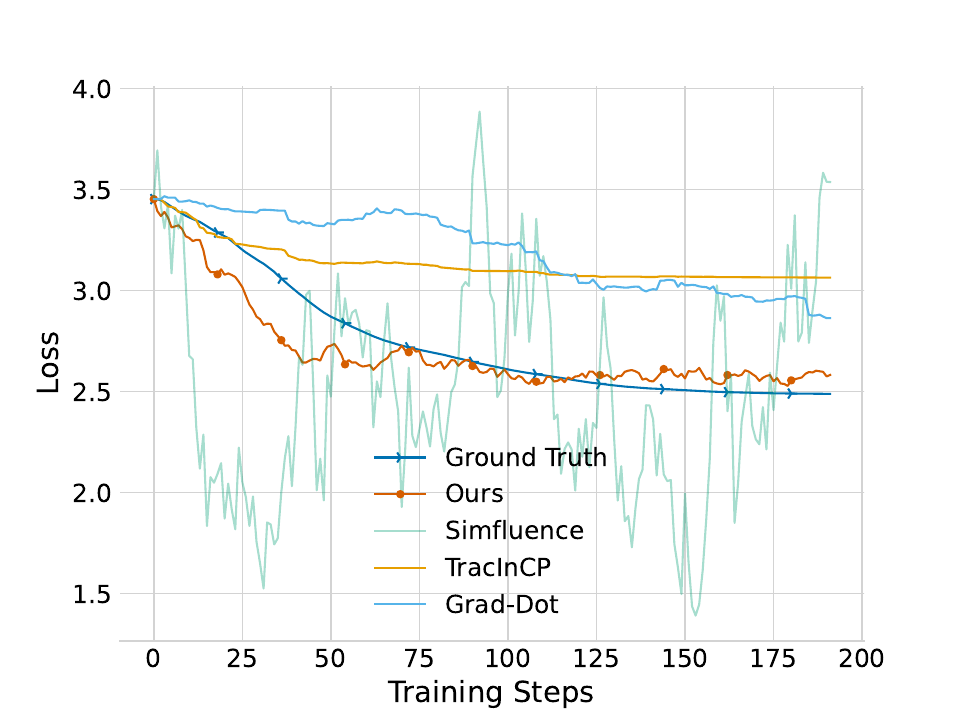}
        \includegraphics[width=0.3\linewidth]{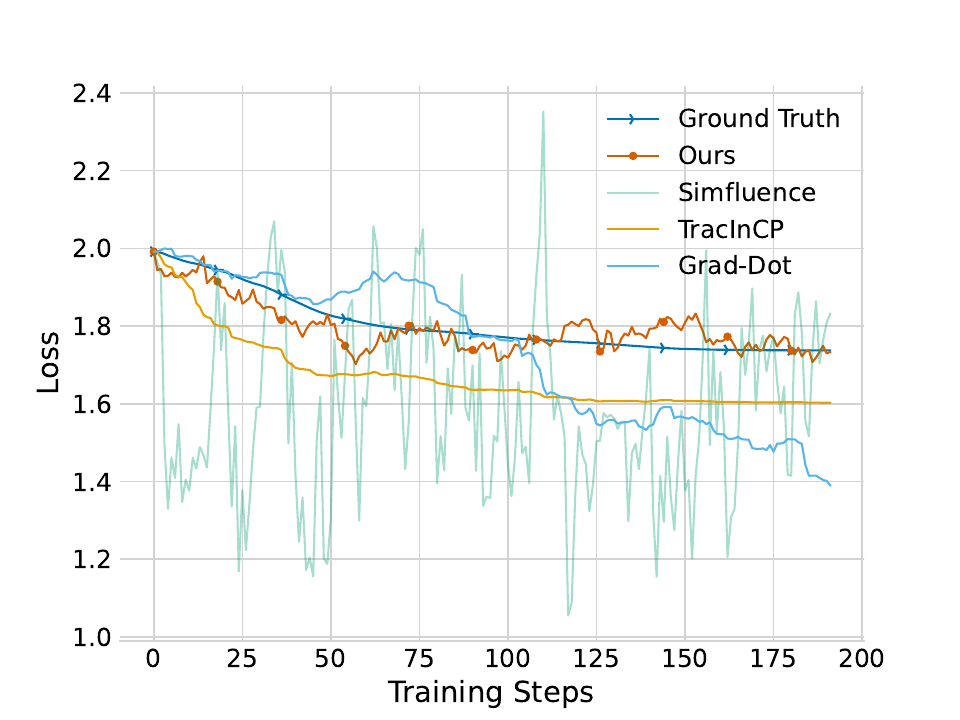}
    }

    \subfigure[WMT16 DE/EN]{
        \includegraphics[width=0.3\linewidth]{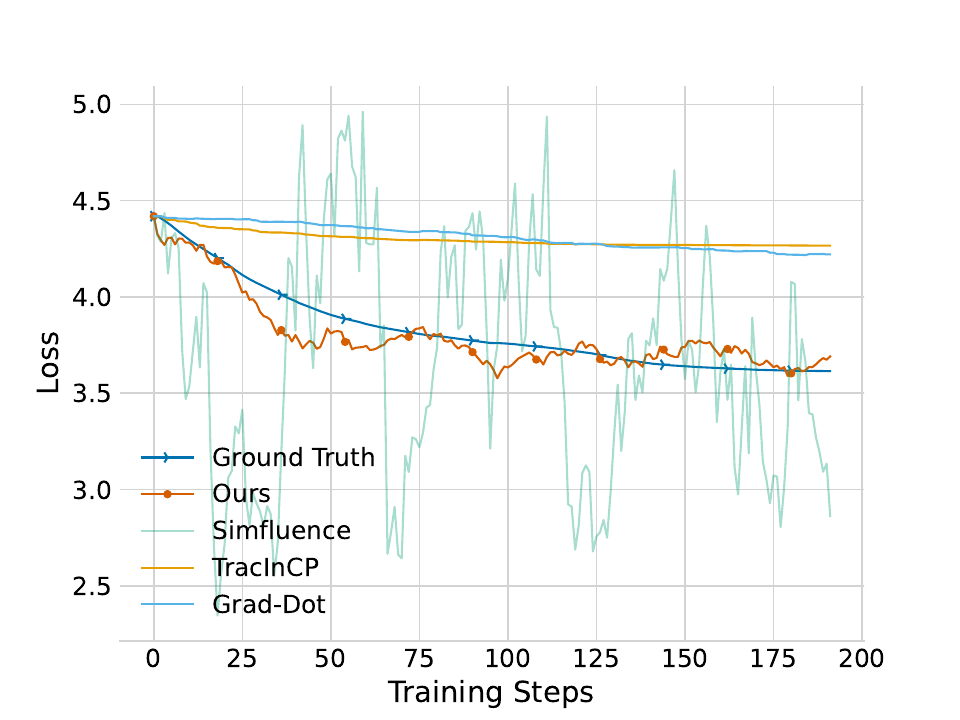}
        \includegraphics[width=0.3\linewidth]{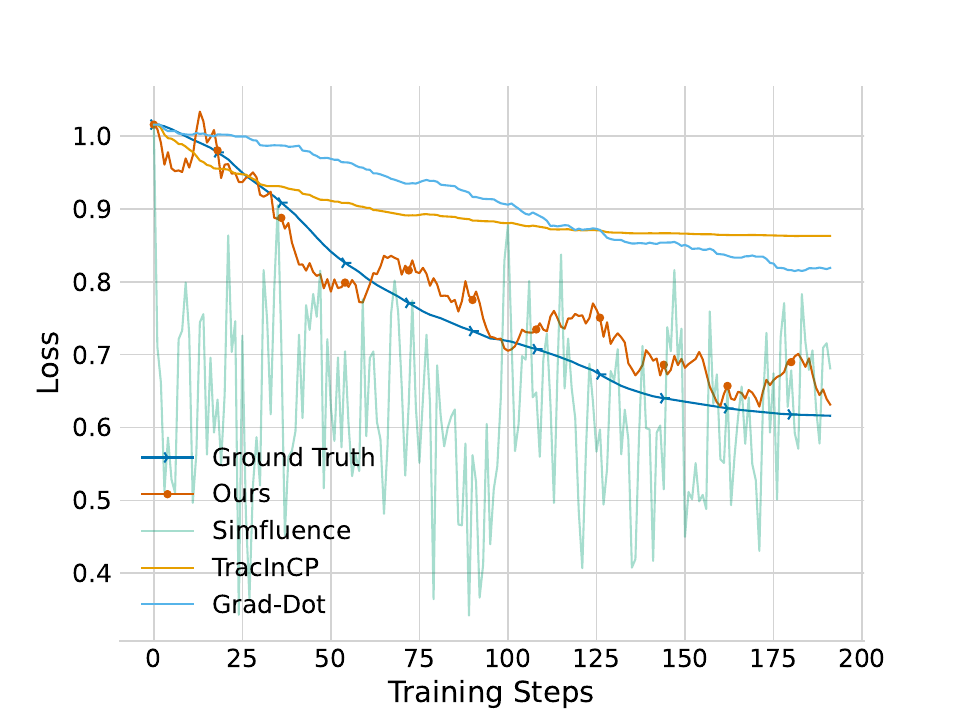}
        \includegraphics[width=0.3\linewidth]{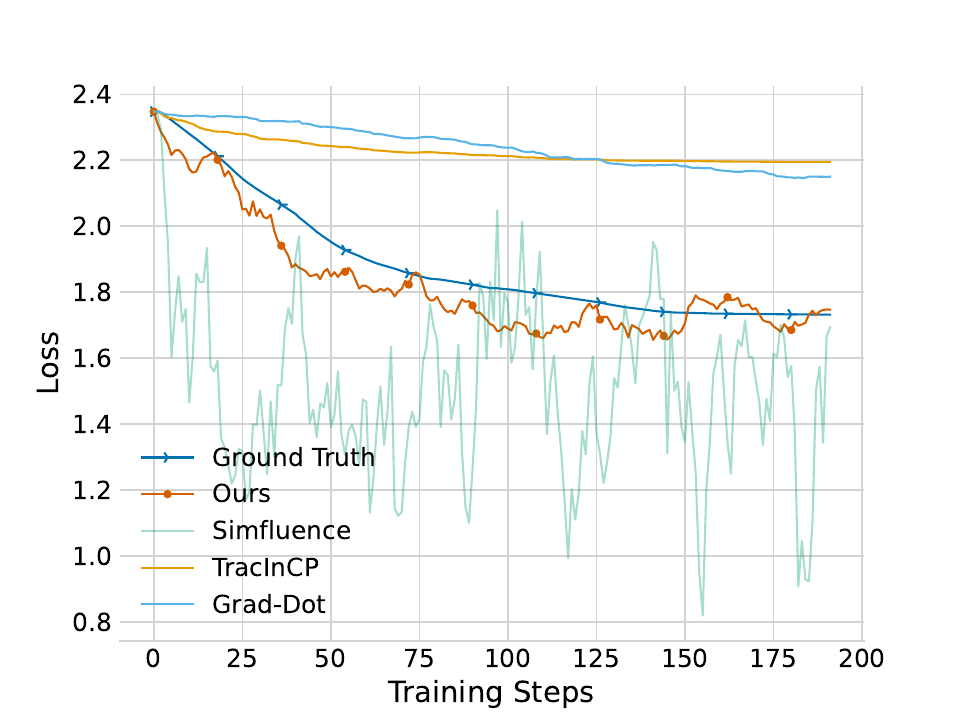}
    }
    \caption{Loss simulation comparisons between {\name} and alternative TDA methods for \textit{instruction tuning} on Pythia-1B, across the BoolQ, RTE, SST-2, WebNLG, and WMT16 DE/EN datasets.}
    \label{fig:fig:exm_it_loss_1b}
\end{figure*}

\begin{figure*}
    \centering
    \subfigure[WebNLG]{
        \includegraphics[width=0.3\linewidth]{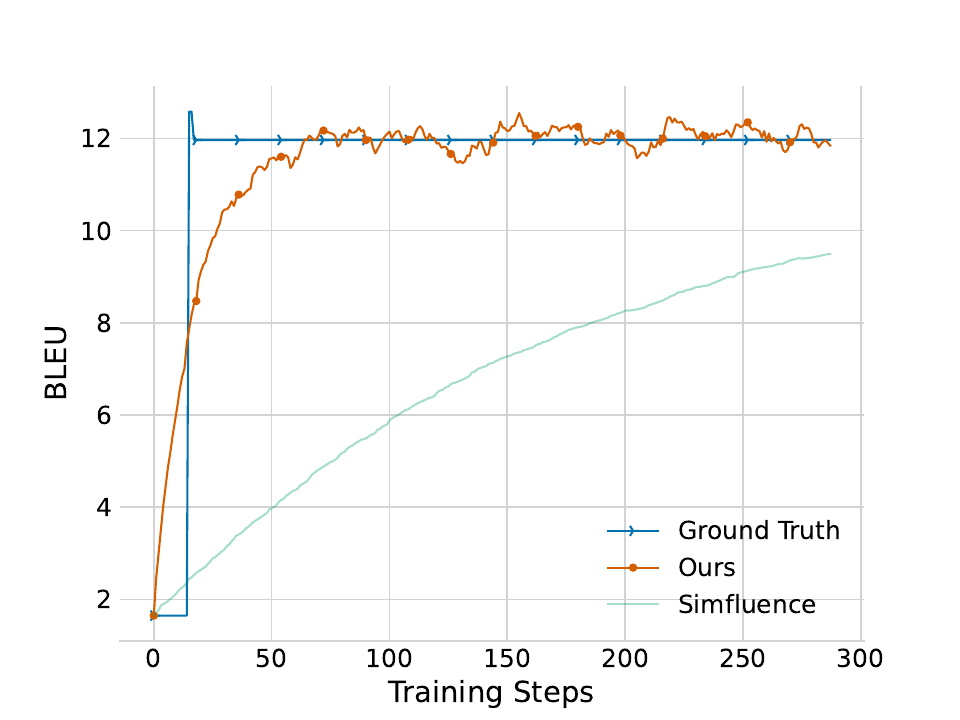}
        \includegraphics[width=0.3\linewidth]{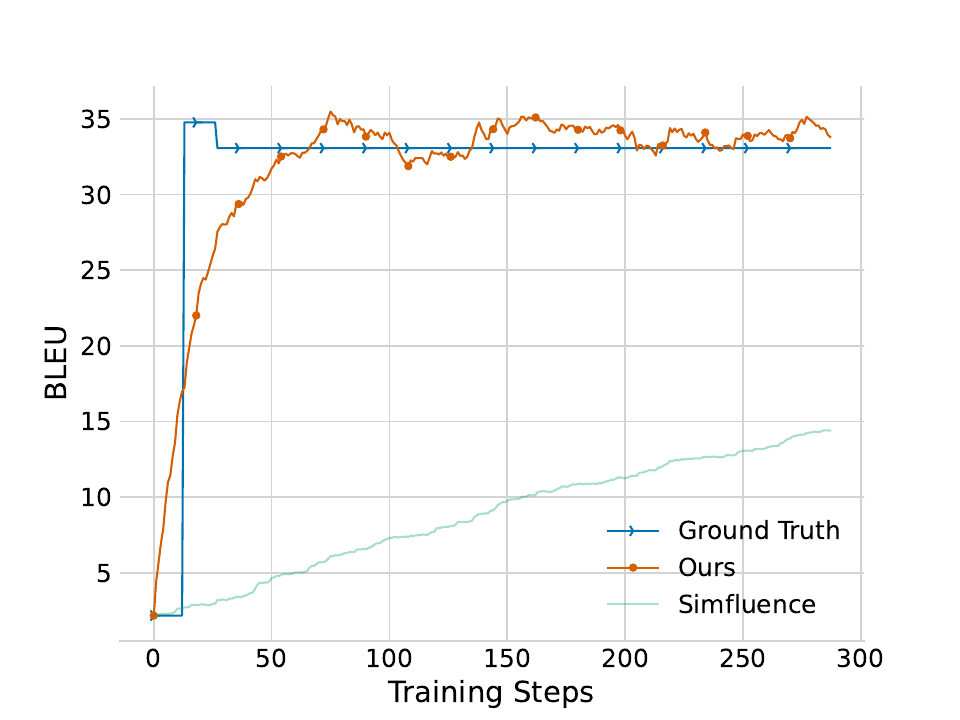}
        \includegraphics[width=0.3\linewidth]{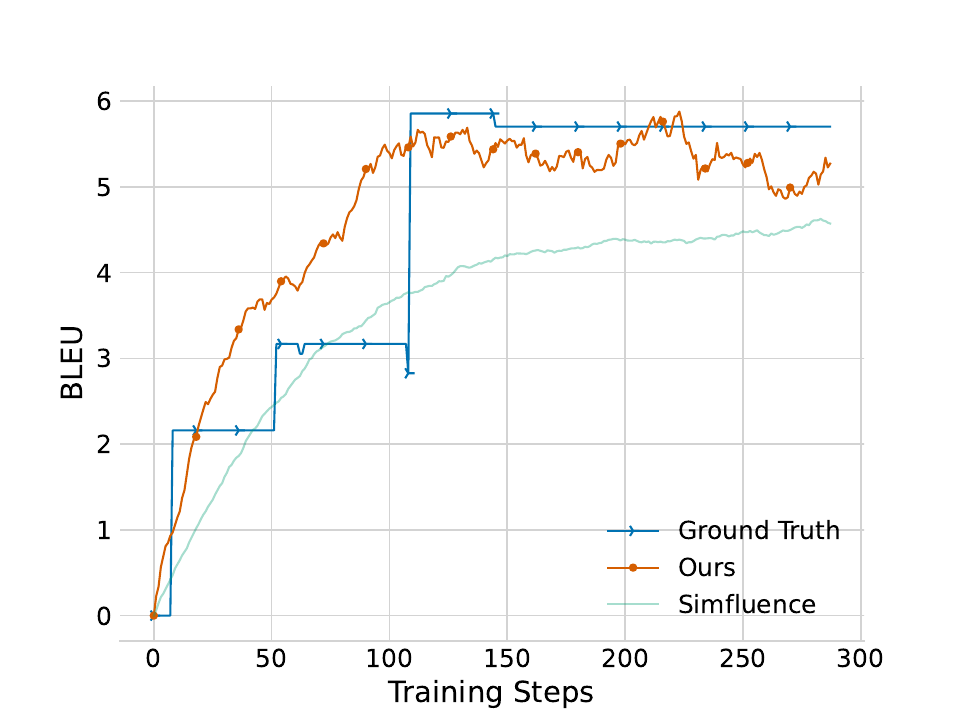}
    }
    
    \subfigure[WMT16 DE/EN]{
        \includegraphics[width=0.3\linewidth]{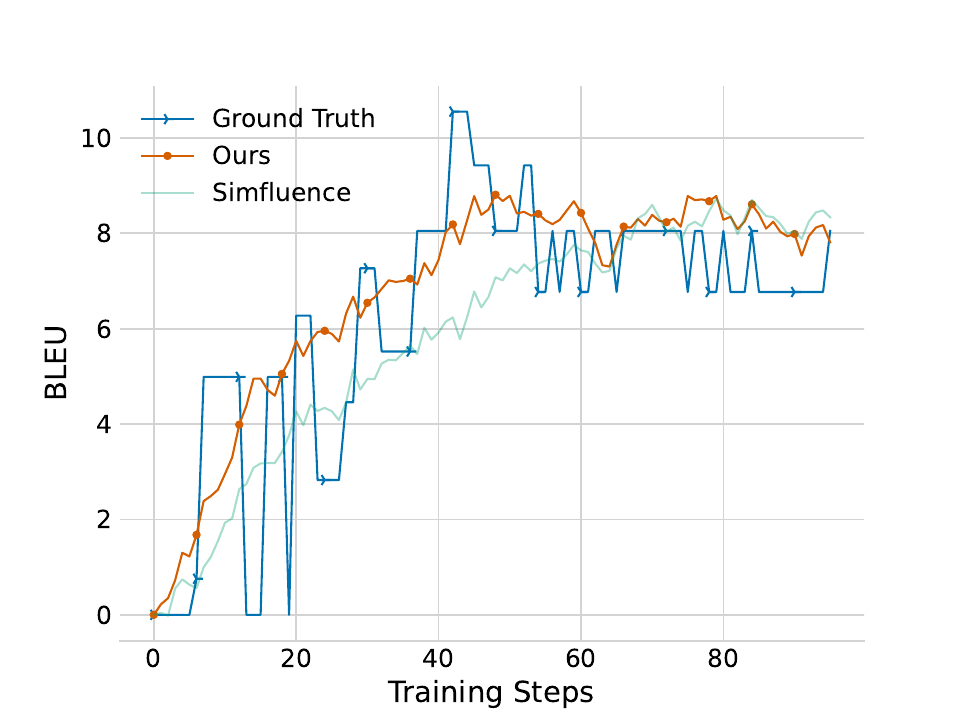}
        \includegraphics[width=0.3\linewidth]{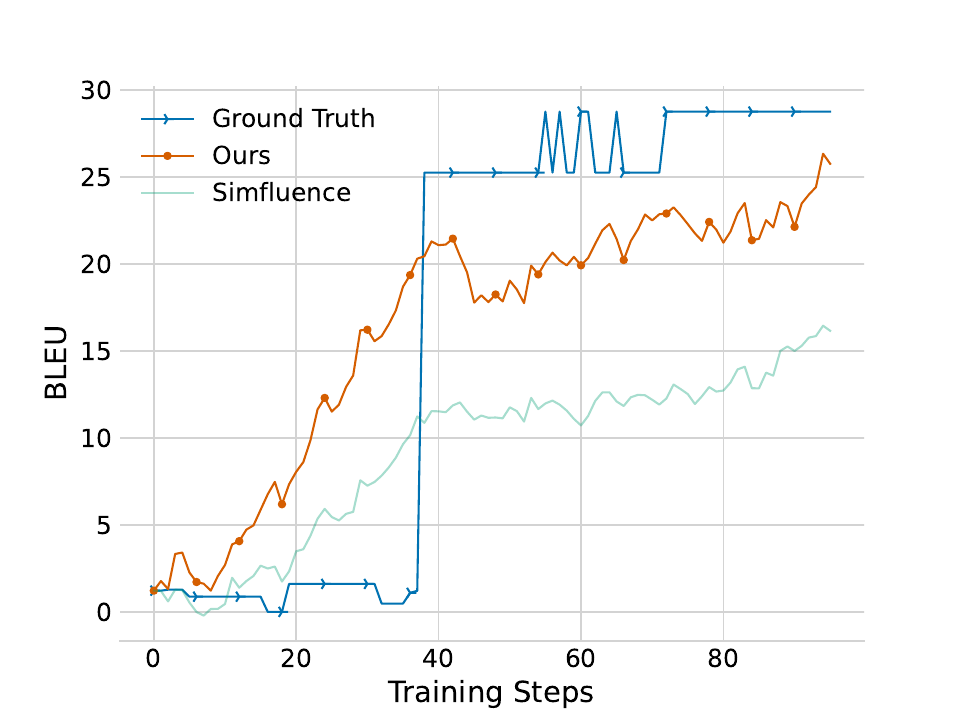}
        \includegraphics[width=0.3\linewidth]{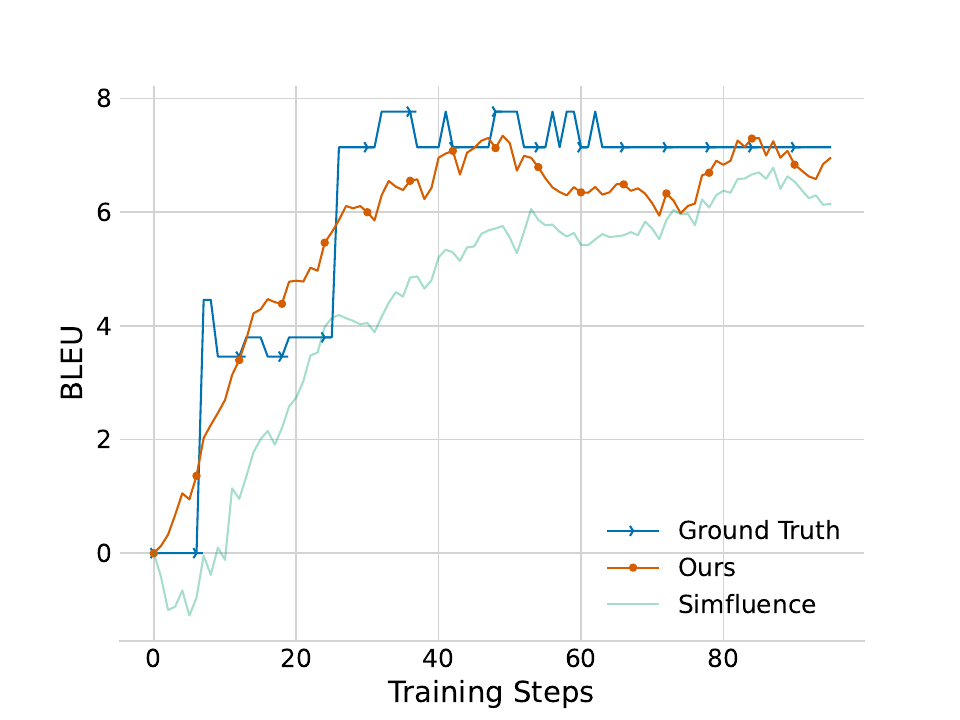}
    }
    \caption{BLEU metric simulation comparisons for \textit{instruction tuning} using {\name} versus Simfluence on Pythia-410M, across the WebNLG and WMT16 DE/EN datasets.}
    \label{fig:exm_it_bleu_410}
\end{figure*}

\begin{figure*}
    \centering
    \subfigure[WebNLG]{
        \includegraphics[width=0.3\linewidth]{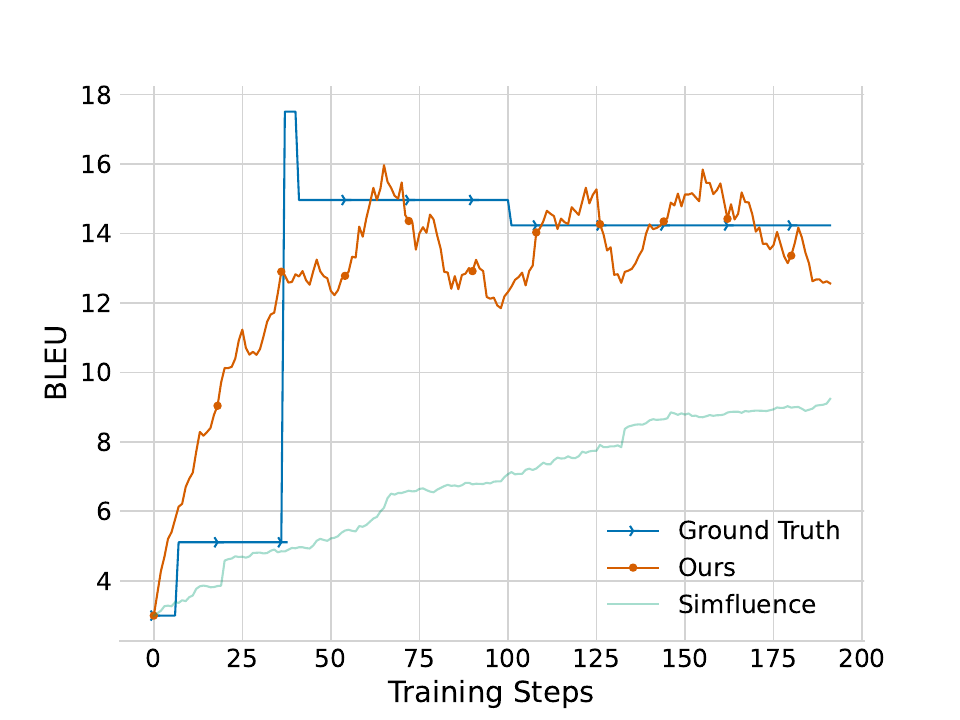}
        \includegraphics[width=0.3\linewidth]{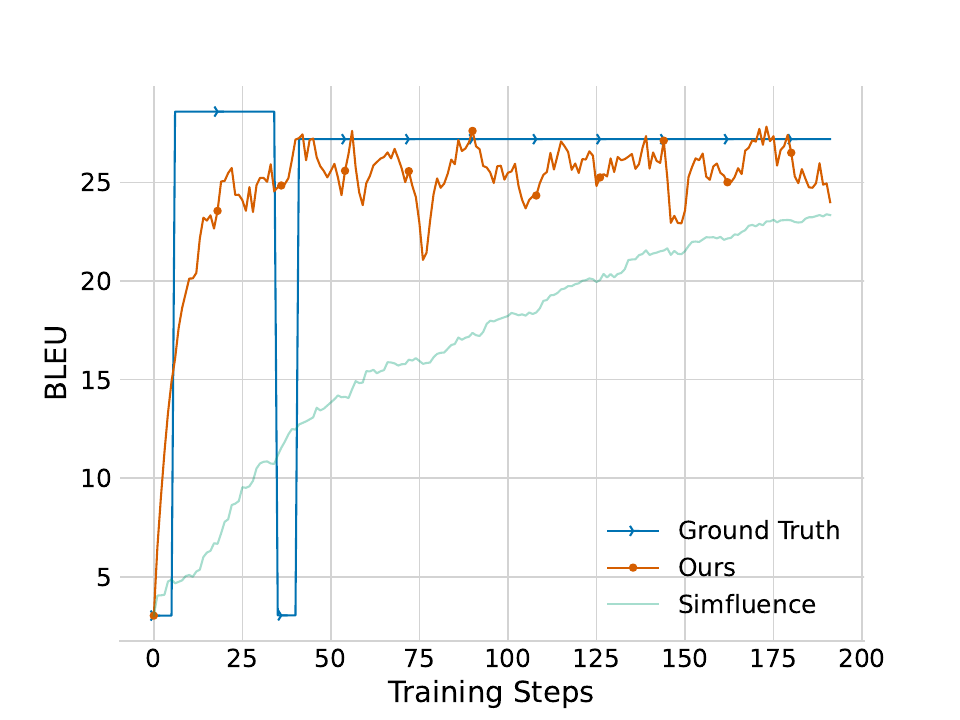}
        \includegraphics[width=0.3\linewidth]{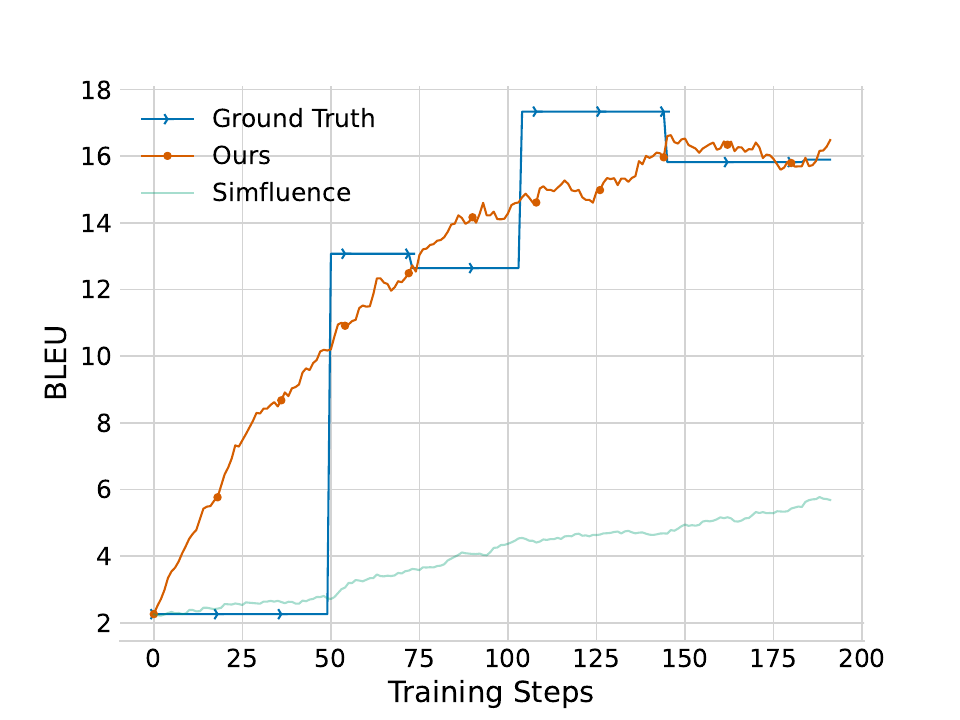}
    }
    
    \subfigure[WMT16 DE/EN]{
        \includegraphics[width=0.3\linewidth]{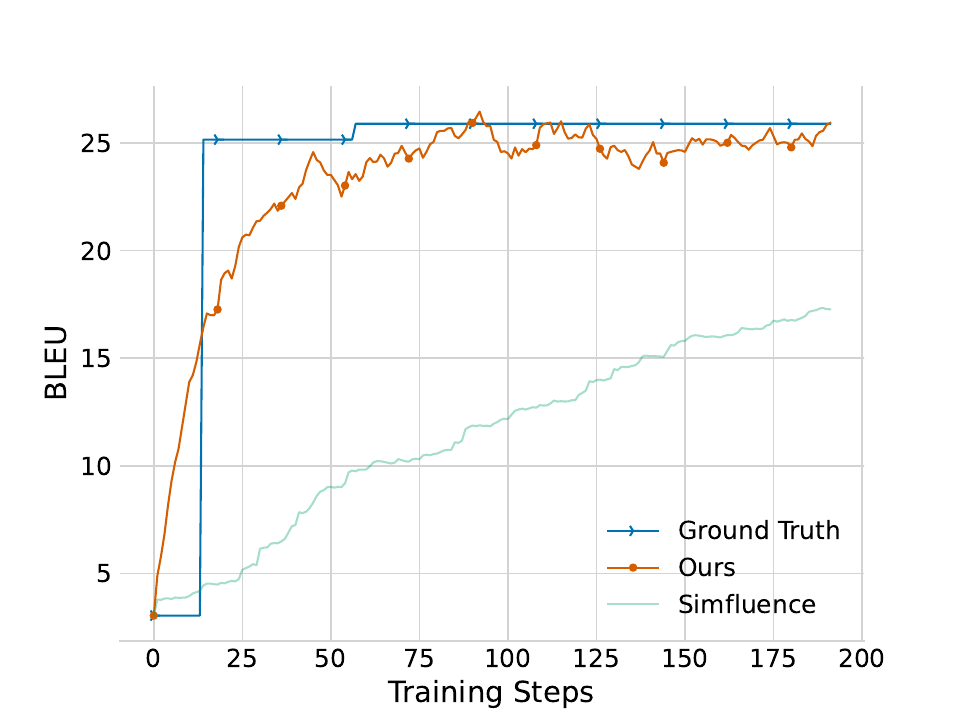}
        \includegraphics[width=0.3\linewidth]{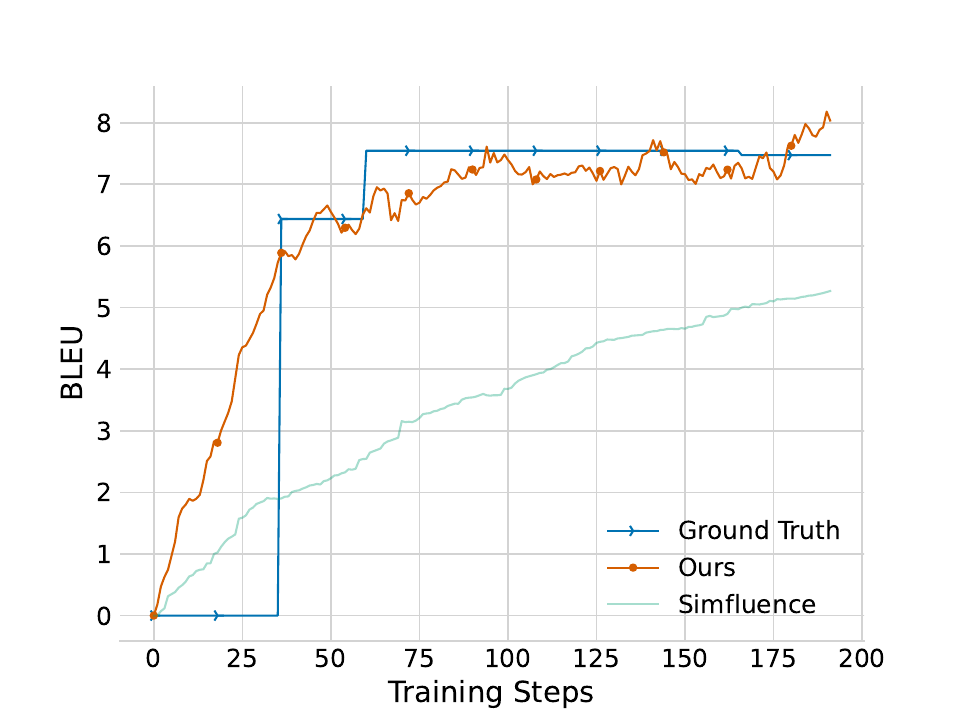}
        \includegraphics[width=0.3\linewidth]{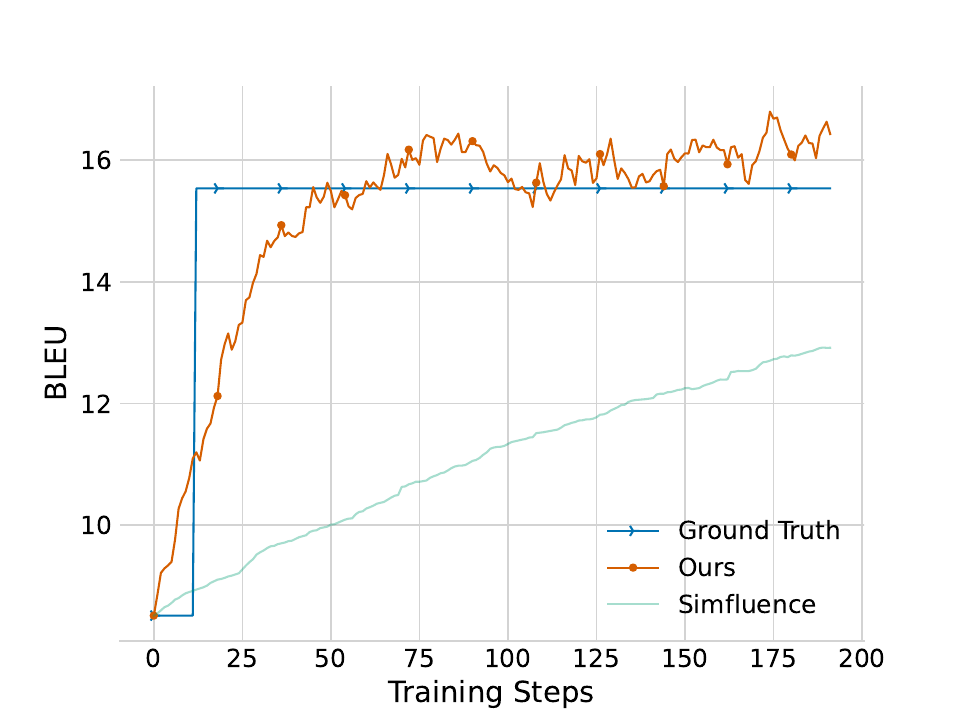}
    }
    \caption{Test examples of the BLEU simulation of {\name} and Simfluence for \textit{instruction tuning} with Pythia-1B on  WebNLG and WMT16 DE/EN datasets.}
    \label{fig:exm_it_bleu_1b}
\end{figure*}

\begin{figure*}
    \centering
    \subfigure[WebNLG]{
        \includegraphics[width=0.3\linewidth]{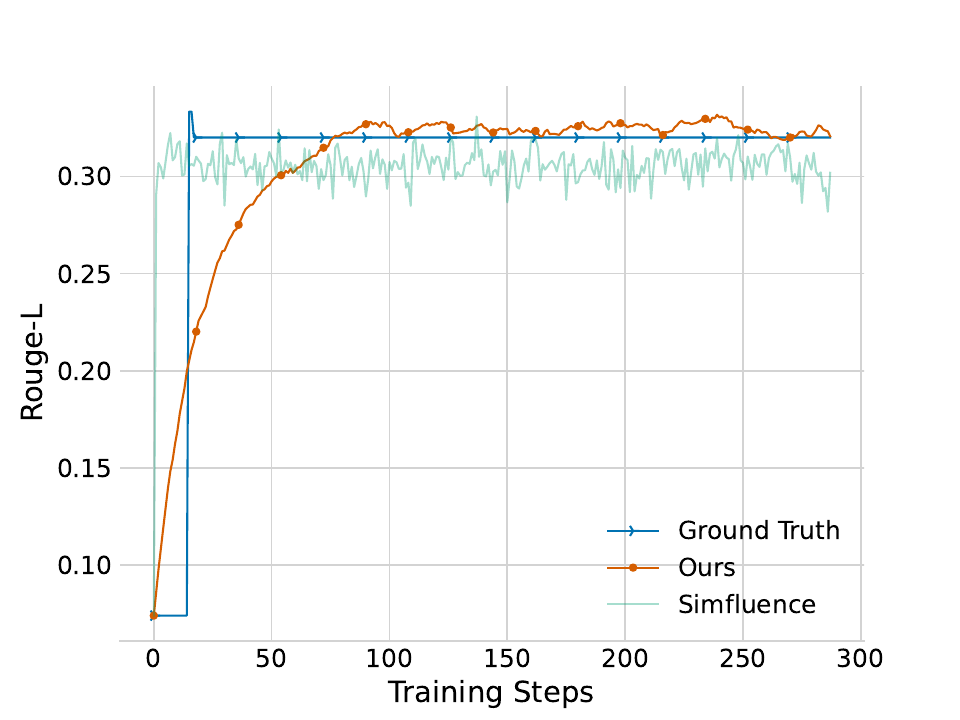}
        \includegraphics[width=0.3\linewidth]{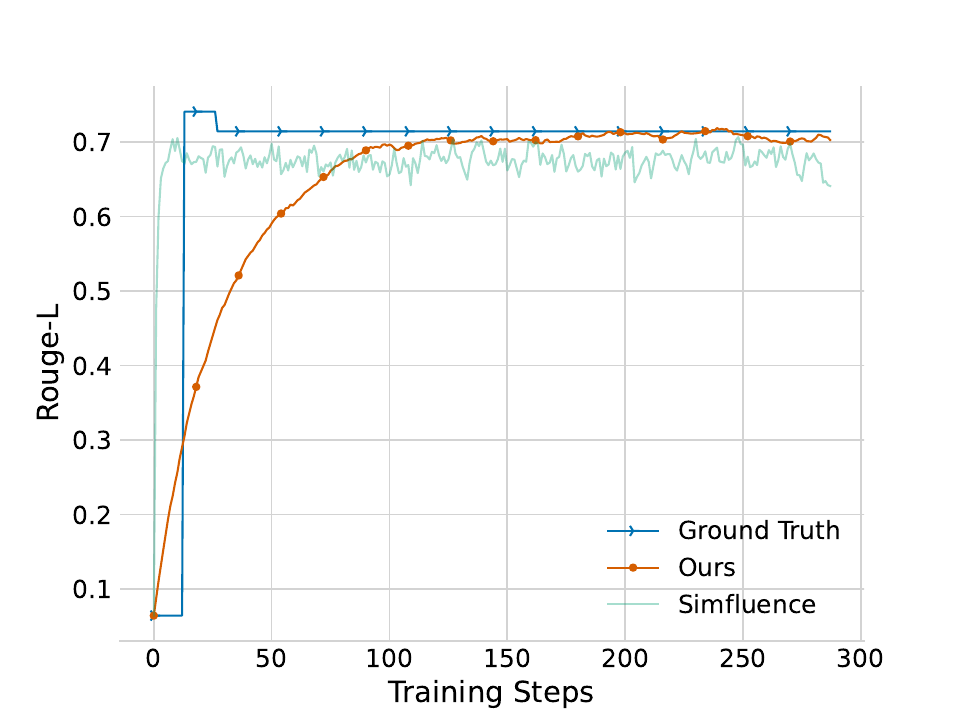}
        \includegraphics[width=0.3\linewidth]{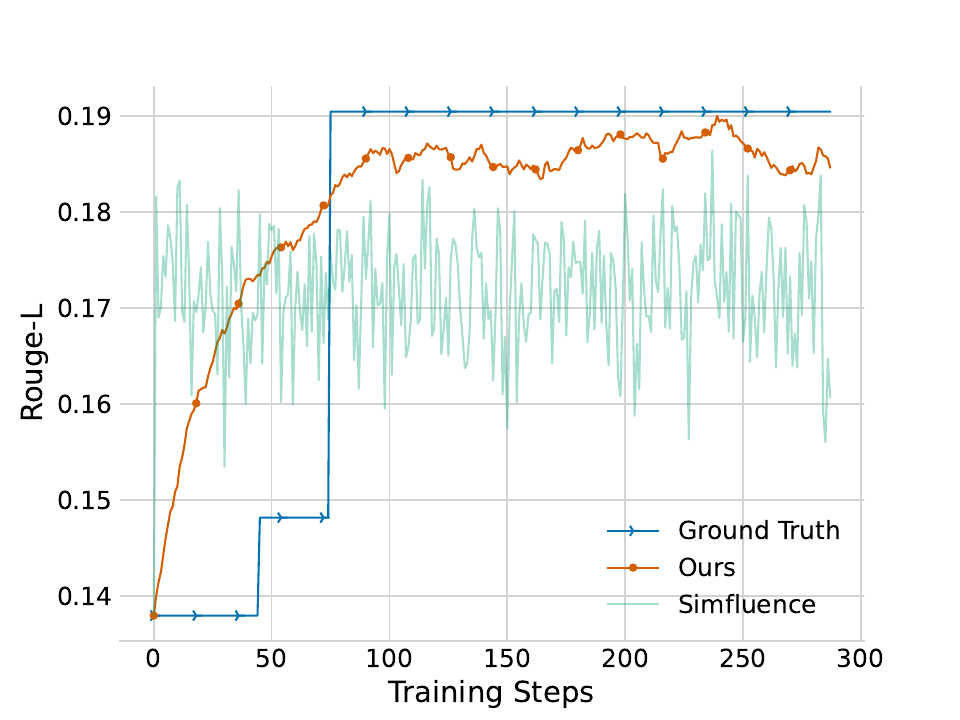}
    }
    
    \subfigure[WMT16 DE/EN]{
        \includegraphics[width=0.3\linewidth]{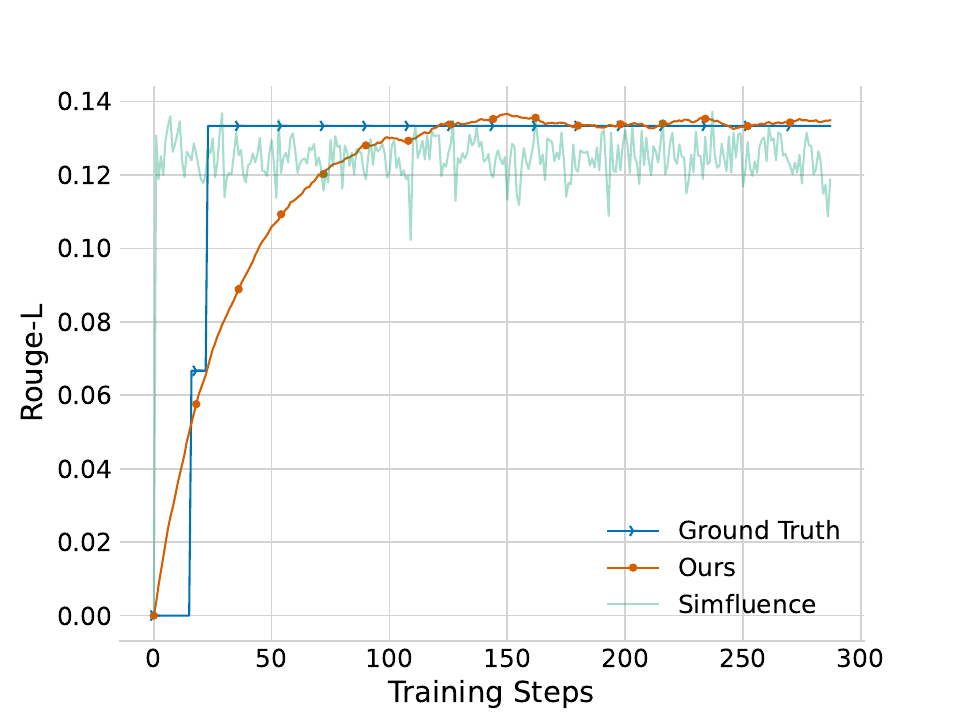}
        \includegraphics[width=0.3\linewidth]{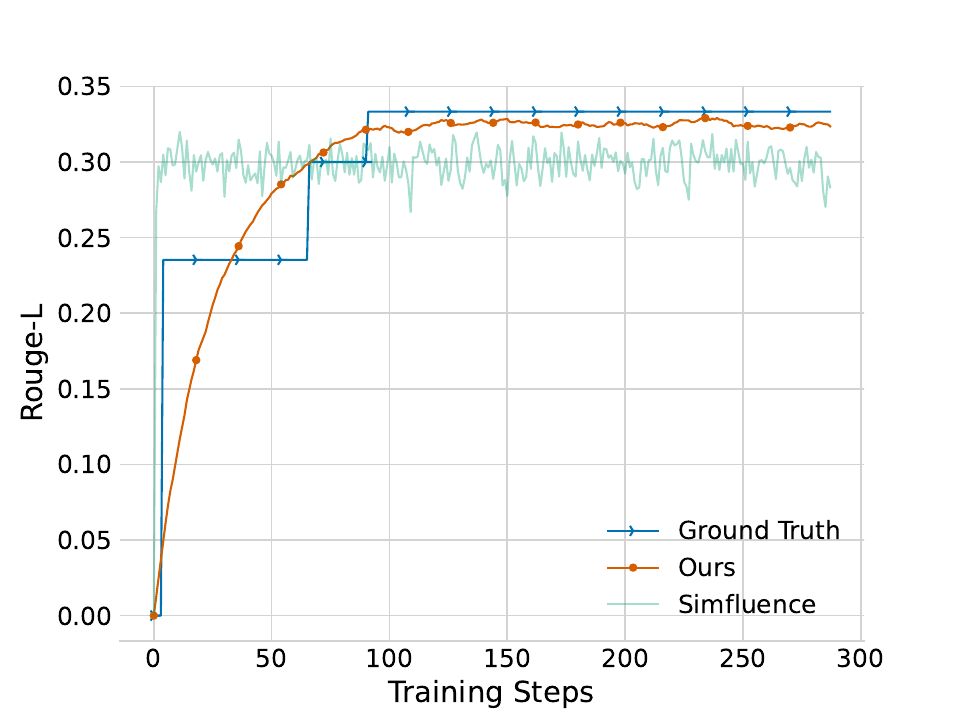}
        \includegraphics[width=0.3\linewidth]{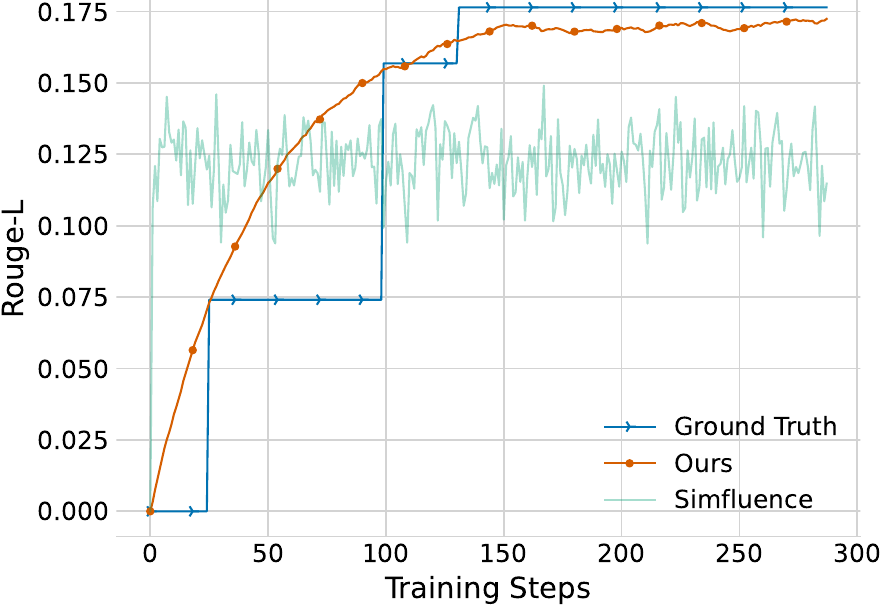}
    }
    \caption{Test examples of the ROUGE-L simulation of {\name} and Simfluence for \textit{instruction tuning} with Pythia-410M on  WebNLG and WMT16 DE/EN datasets.}
    \label{fig:exm_it_rougel_410}
\end{figure*}

\begin{figure*}
    \centering
    \subfigure[WebNLG]{
        \includegraphics[width=0.3\linewidth]{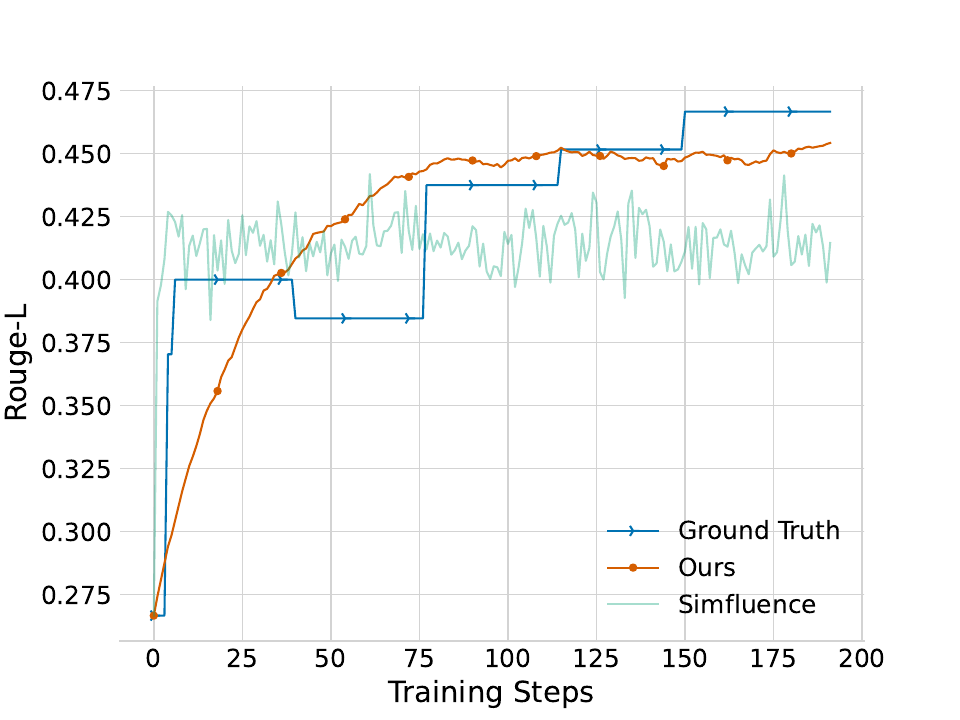}
        \includegraphics[width=0.3\linewidth]{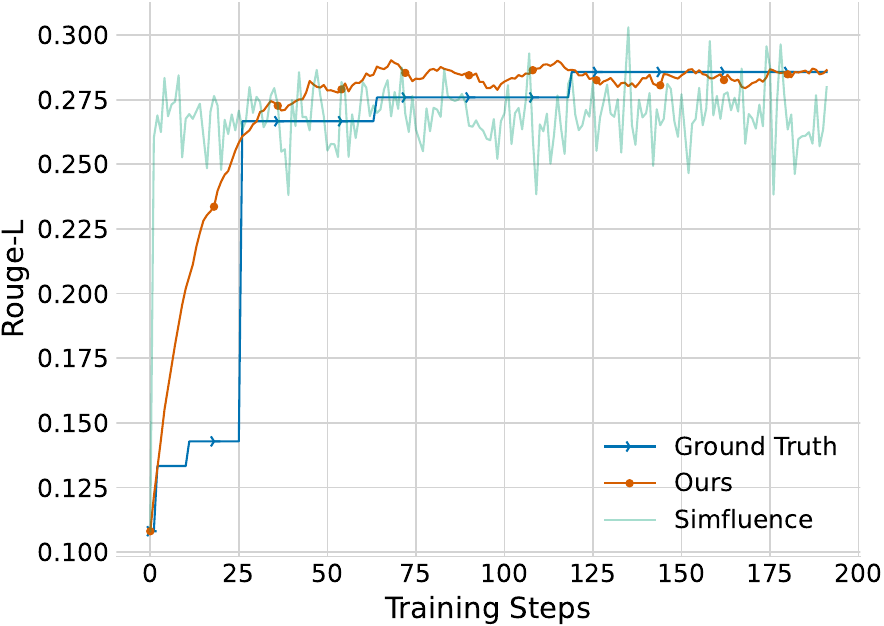}
        \includegraphics[width=0.3\linewidth]{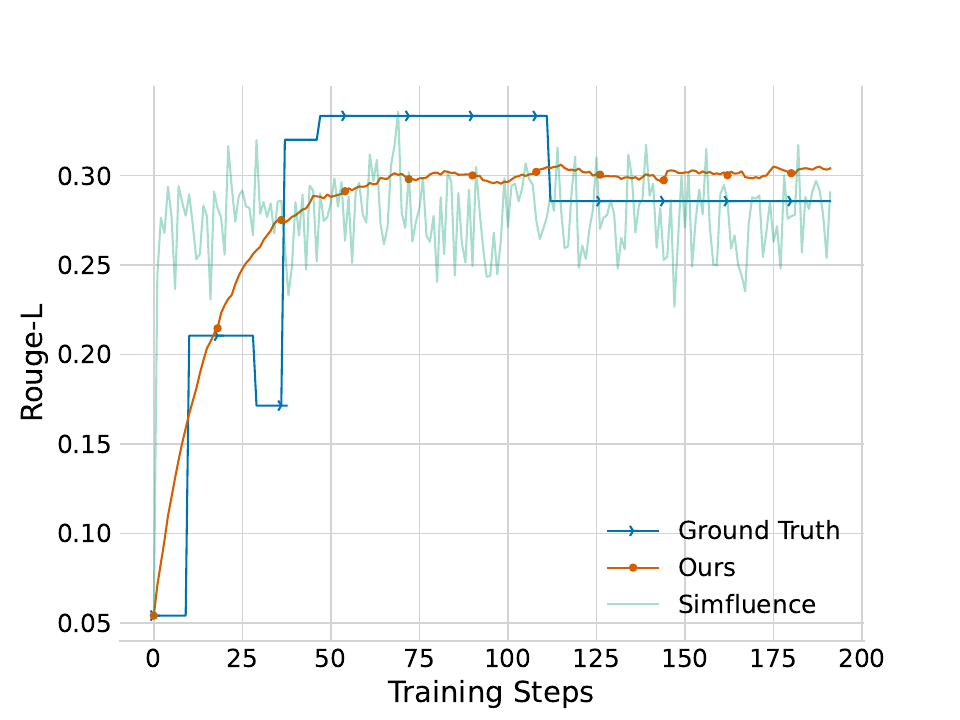}
    }
    
    \subfigure[WMT16 DE/EN]{
        \includegraphics[width=0.3\linewidth]{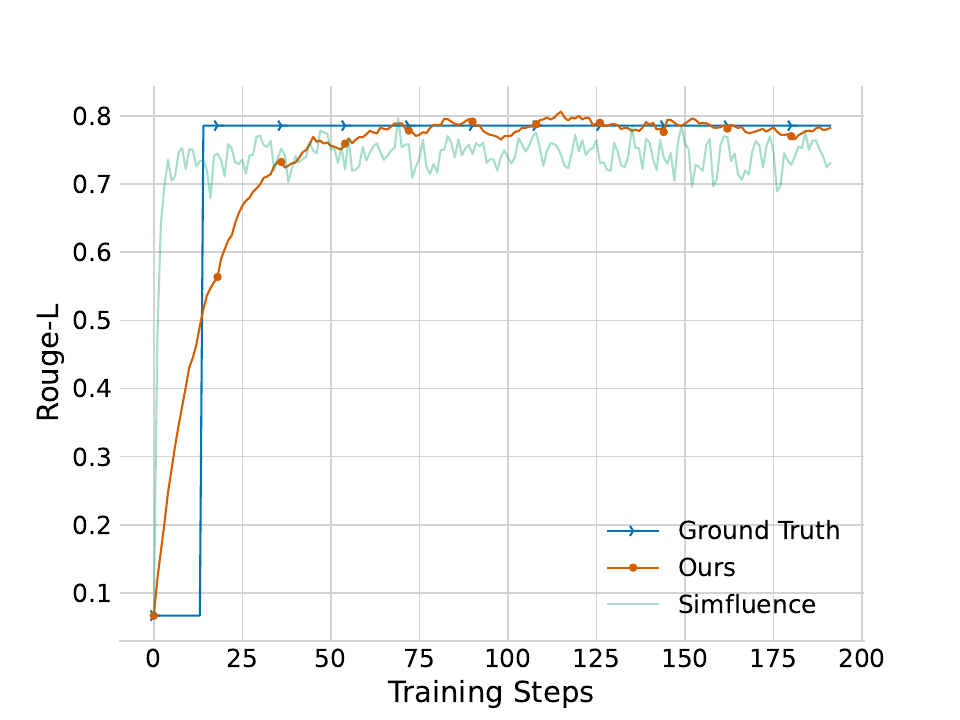}
        \includegraphics[width=0.3\linewidth]{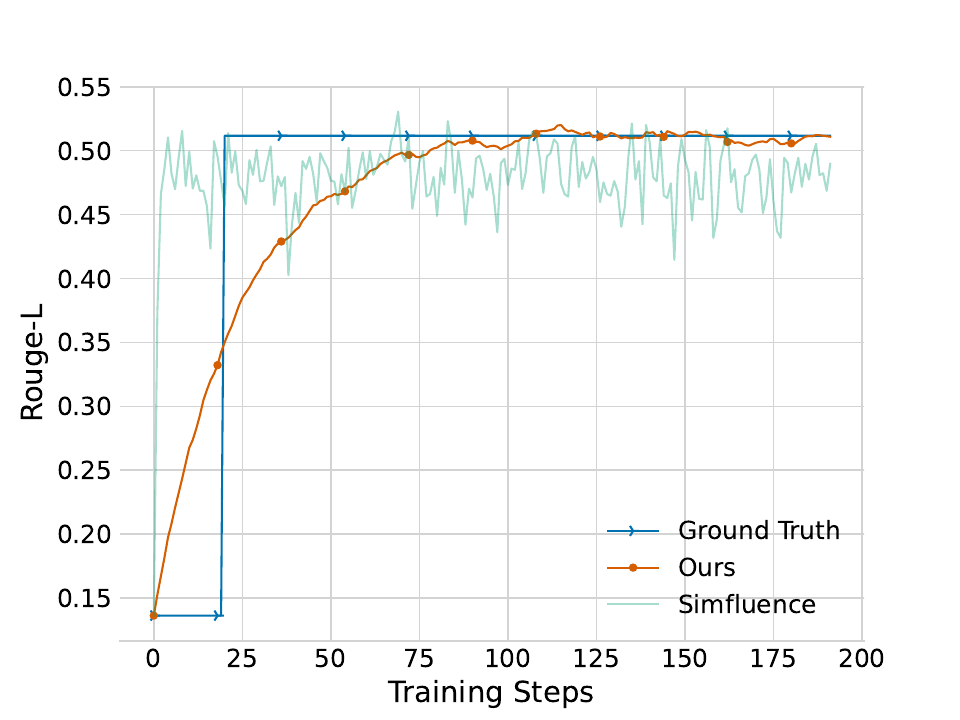}
        \includegraphics[width=0.3\linewidth]{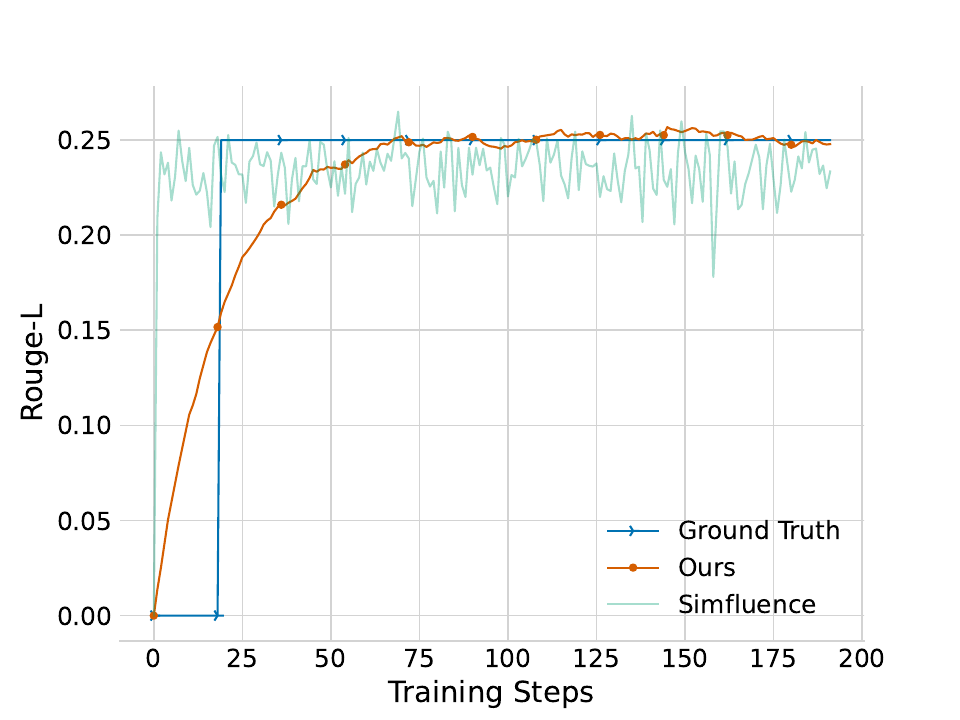}
    }
    \caption{Test examples of the ROUGE-L simulation of {\name} and Simfluence for \textit{instruction tuning} with Pythia-1B on  WebNLG and WMT16 DE/EN datasets.}
    \label{fig:exm_it_rougel_1b}
\end{figure*}

We provide additional qualitative examples for instruction-tuning simulations, highlighting test loss and performance metrics:
\begin{itemize}
    \item Simulation of test loss for Pythia-410M is shown in Fig.~\ref{fig:exm_it_loss_410}.
    \item Simulation of test loss for Pythia-1B is depicted in Fig.~\ref{fig:fig:exm_it_loss_1b}.
    \item BLEU metric simulation for Pythia-410M can be found in Fig.~\ref{fig:exm_it_bleu_410}.
    \item BLEU metric simulation for Pythia-1B is illustrated in Fig.~\ref{fig:exm_it_bleu_1b}.
    \item ROUGE-L metric simulation with Pythia-410M is presented in Fig.~\ref{fig:exm_it_rougel_410}.
    \item ROUGE-L metric simulation with Pythia-1B is detailed in Fig.~\ref{fig:exm_it_rougel_1b}.
\end{itemize}

\subsection{Simulation For Fine-Tuning}
\label{ap:examples_ft}
\begin{figure*}
    \centering
    \subfigure[BoolQ]{
        \includegraphics[width=0.3\linewidth]{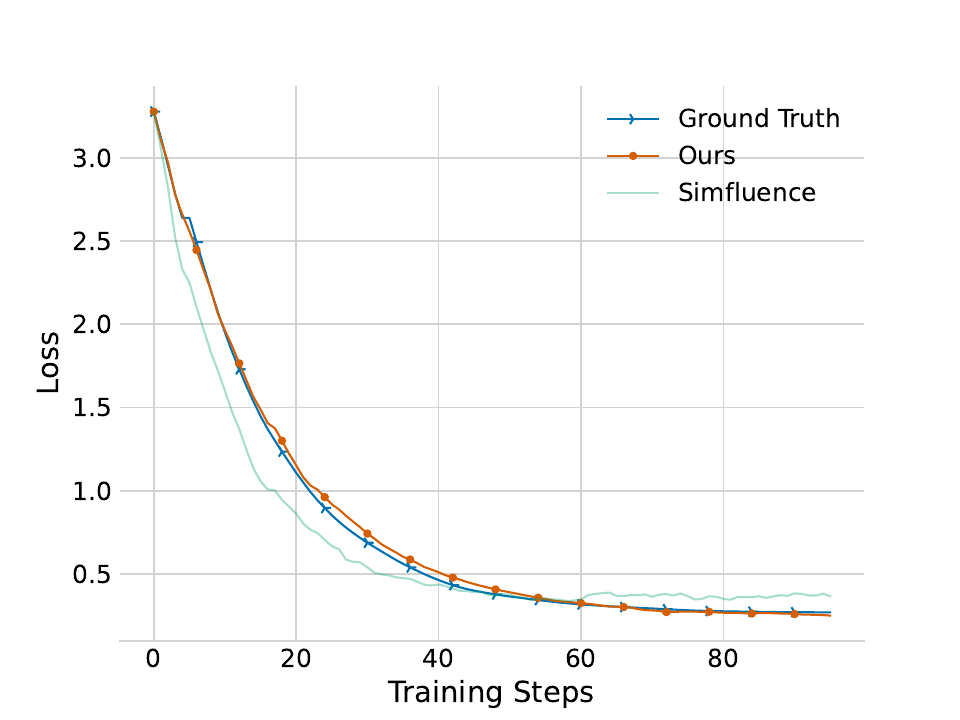}
        \includegraphics[width=0.3\linewidth]{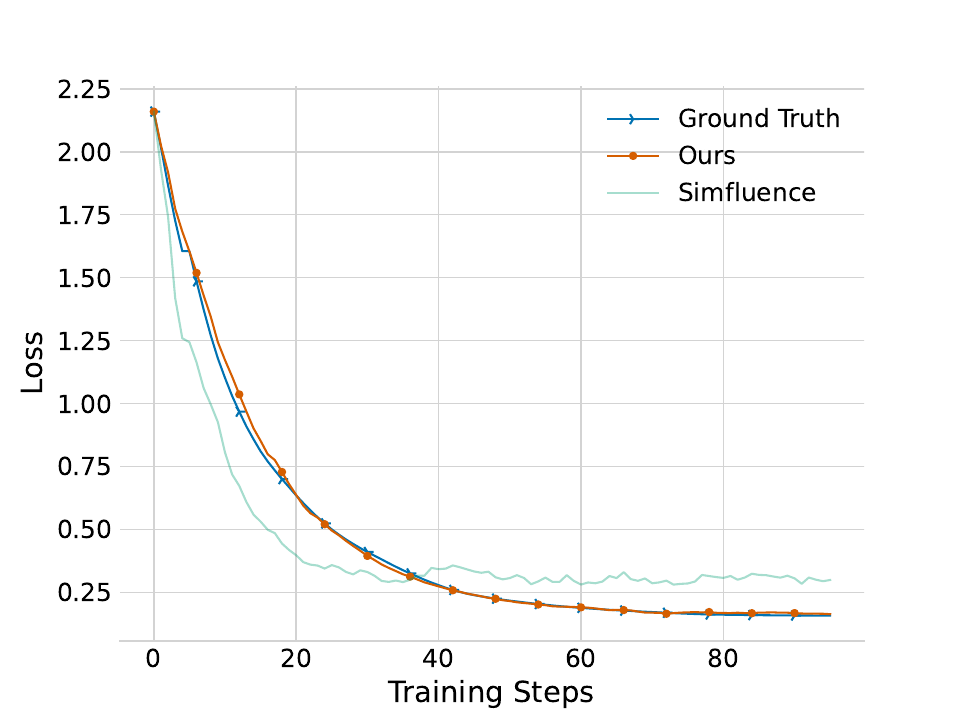}
        \includegraphics[width=0.3\linewidth]{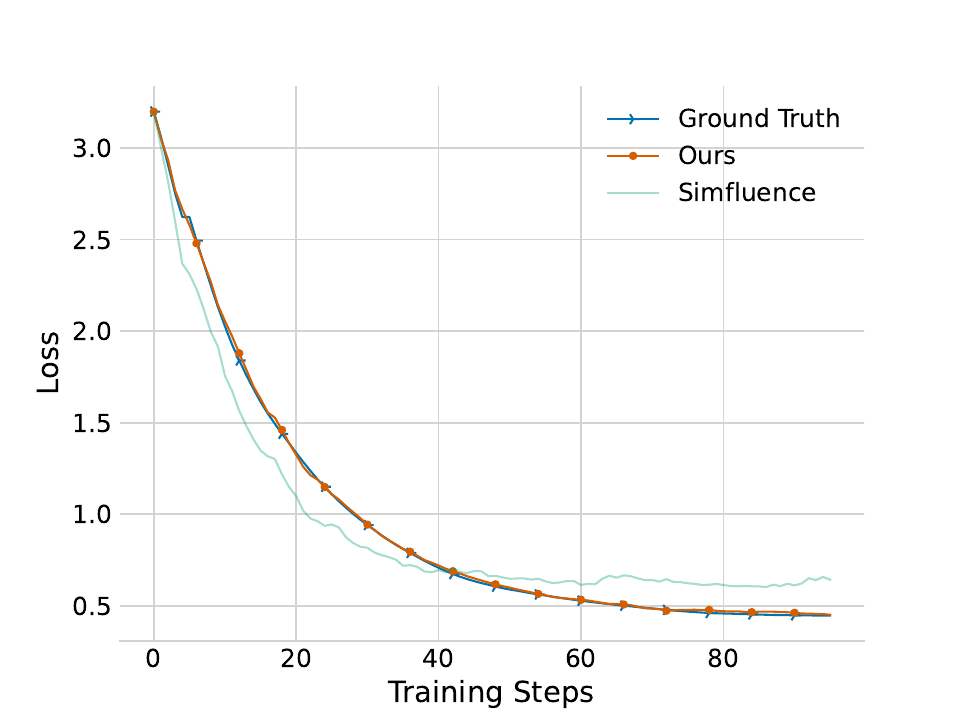}
    }
    
    \subfigure[RTE]{
        \includegraphics[width=0.3\linewidth]{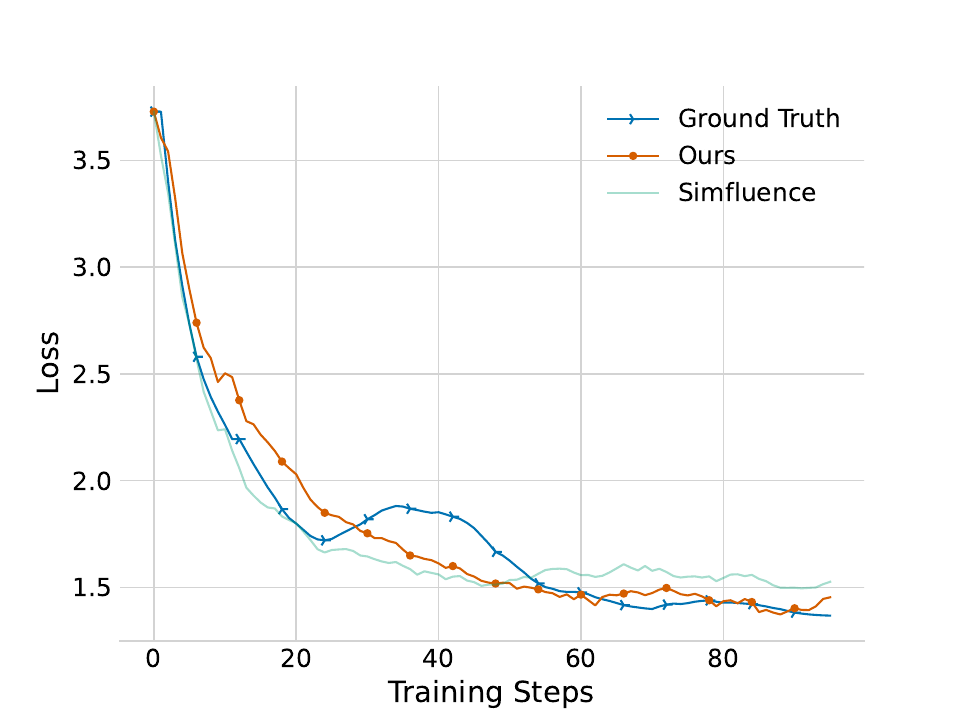}
        \includegraphics[width=0.3\linewidth]{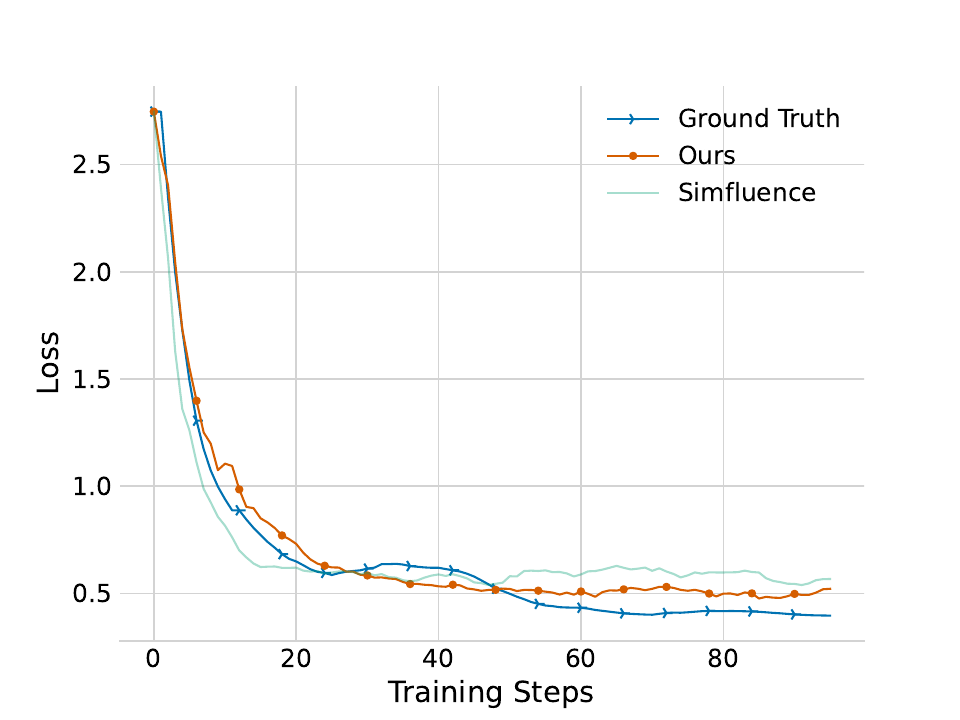}
        \includegraphics[width=0.3\linewidth]{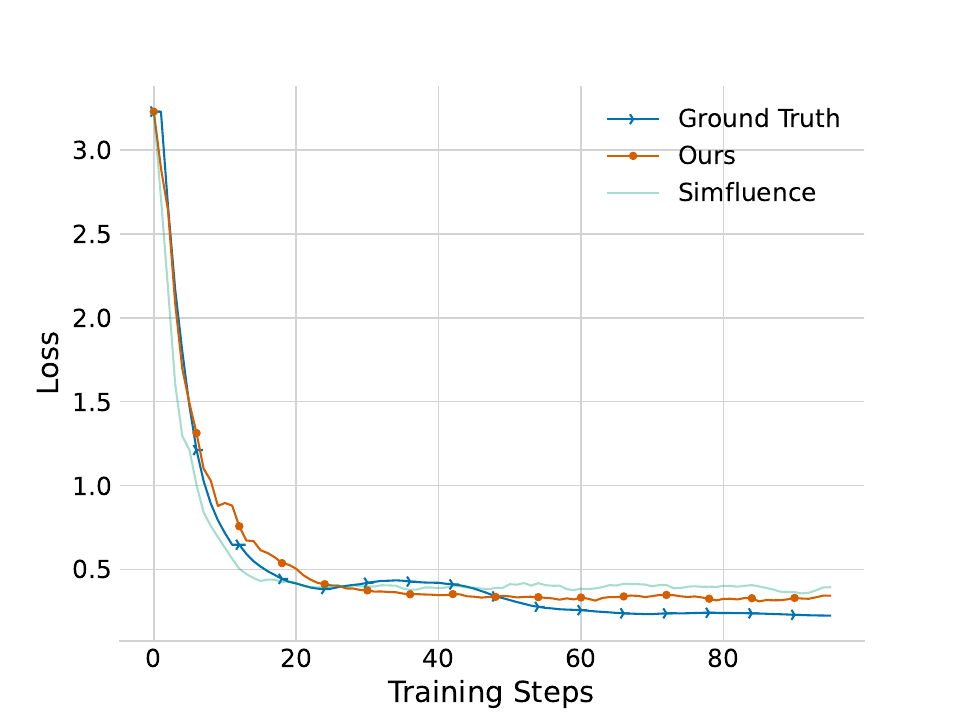}
    }
    
    \subfigure[SST-2]{
        \includegraphics[width=0.3\linewidth]{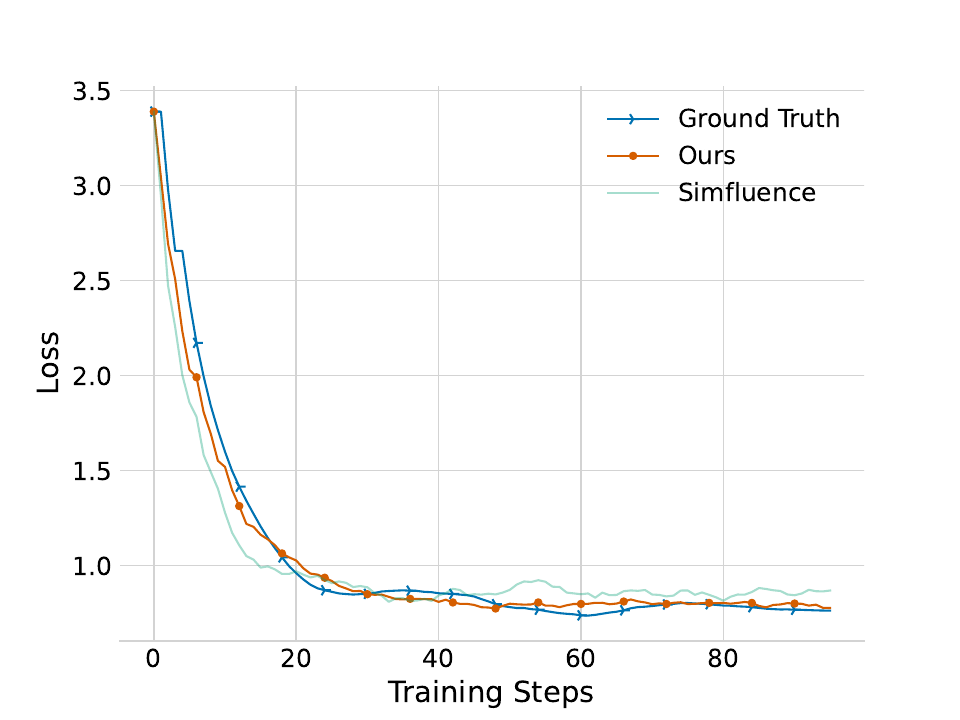}
        \includegraphics[width=0.3\linewidth]{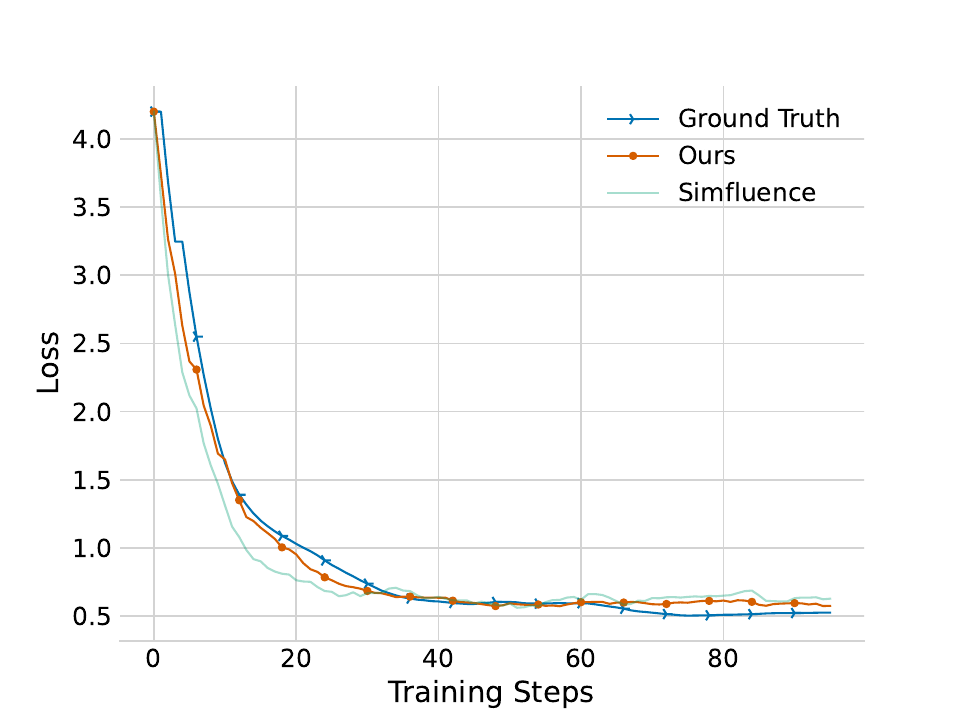}
        \includegraphics[width=0.3\linewidth]{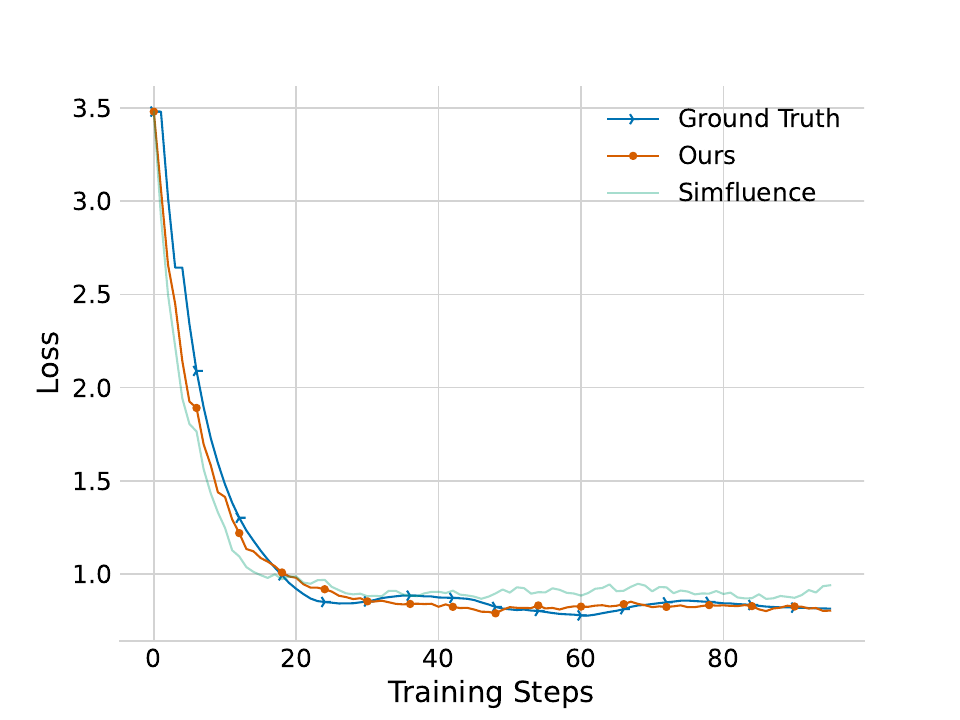}
    }

    \subfigure[WebNLG]{
        \includegraphics[width=0.3\linewidth]{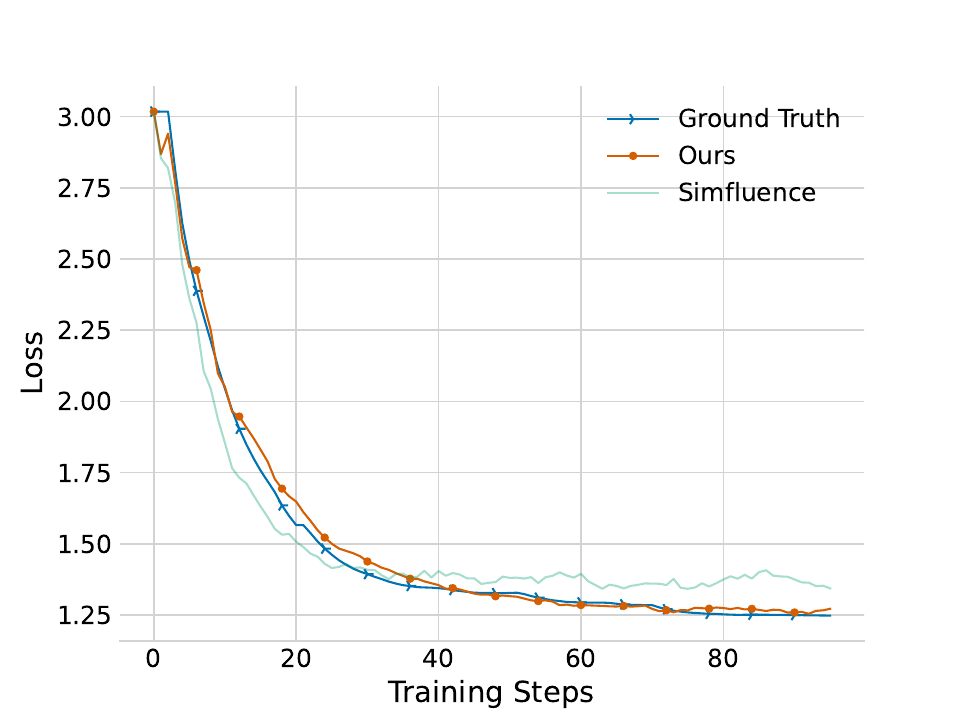}
        \includegraphics[width=0.3\linewidth]{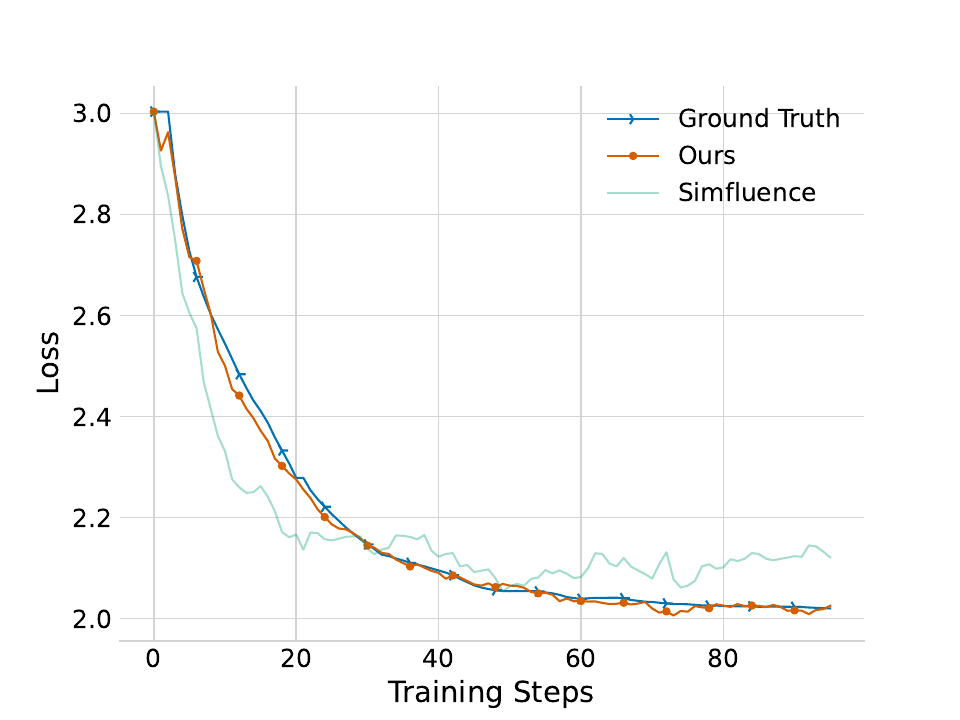}
        \includegraphics[width=0.3\linewidth]{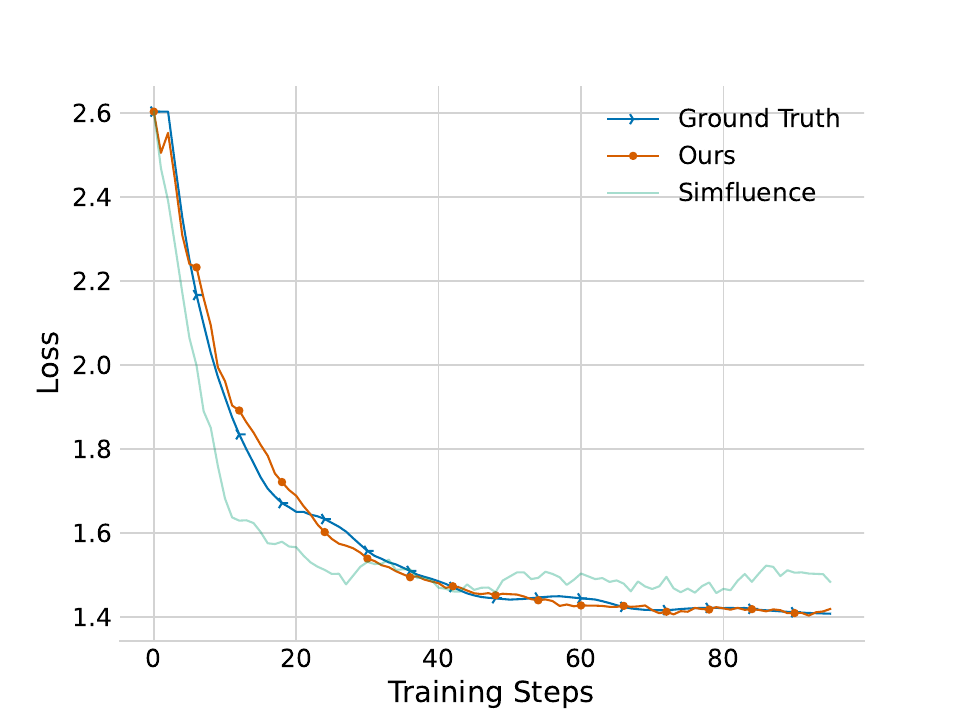}
    }

    \subfigure[WMT16 DE/EN]{
        \includegraphics[width=0.3\linewidth]{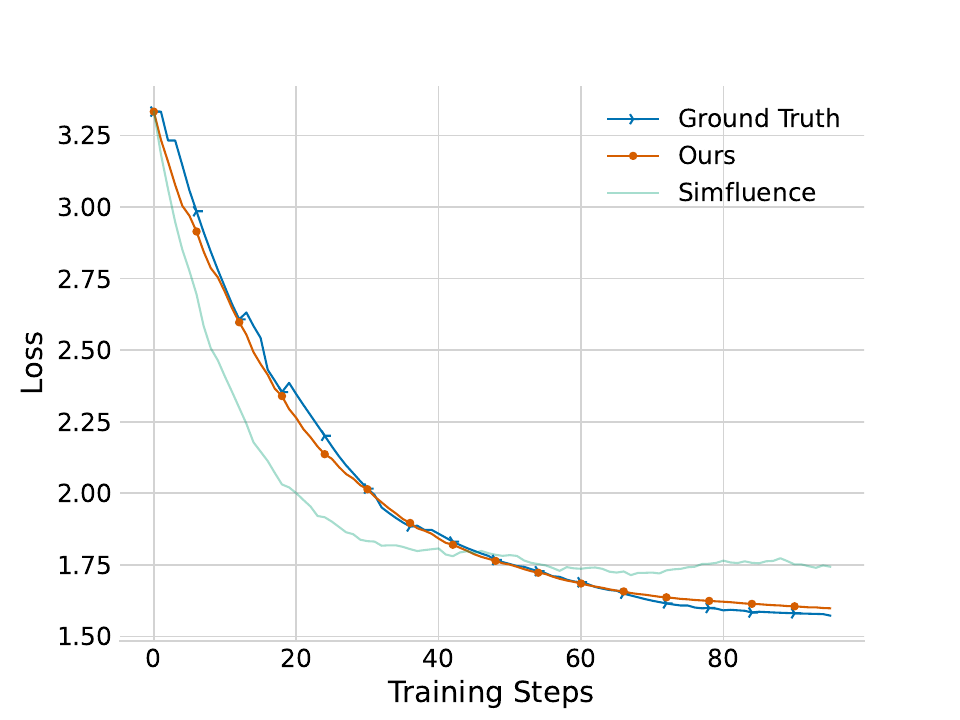}
        \includegraphics[width=0.3\linewidth]{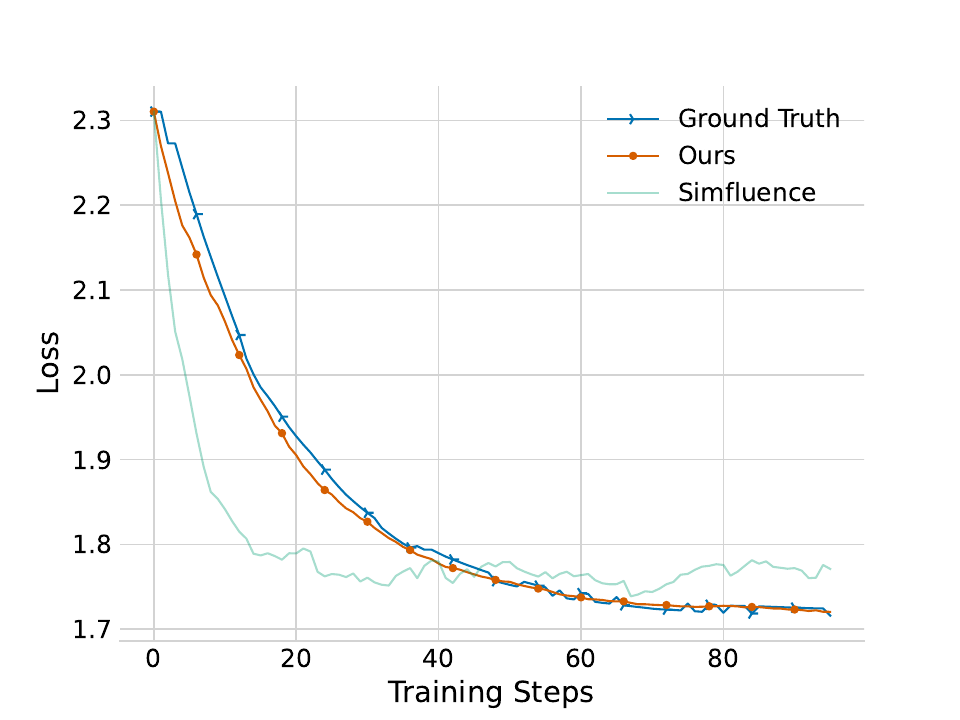}
        \includegraphics[width=0.3\linewidth]{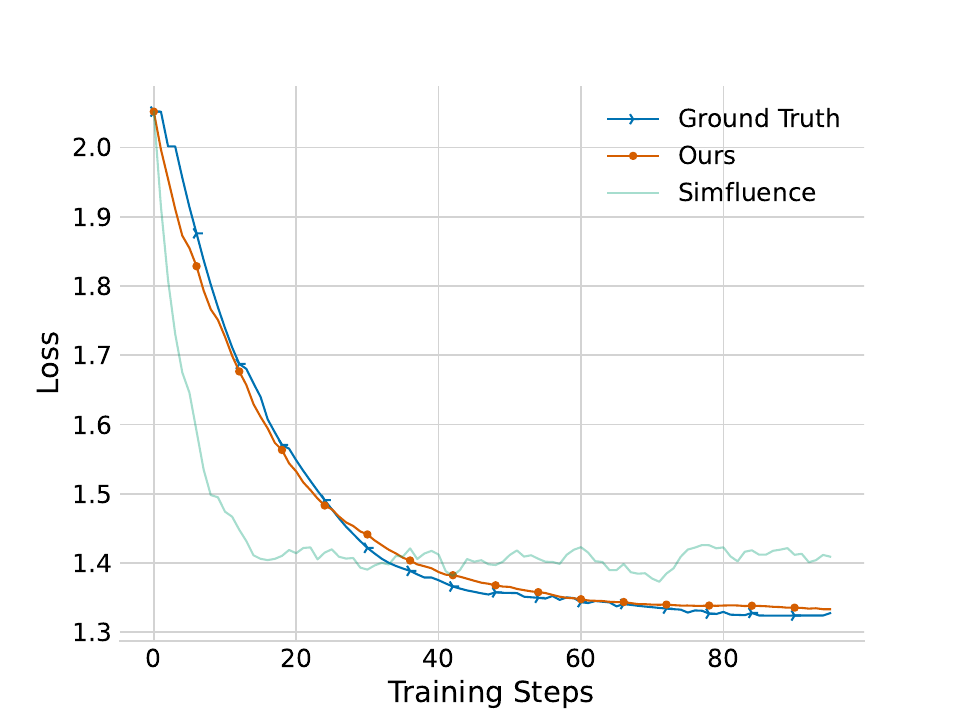}
    }
    \caption{Loss simulation comparisons of {\name} versus Simfluence for \textit{fine-tuning} on Pythia-410M across BoolQ, RTE, SST-2, WebNLG, and WMT16 DE/EN datasets.}
    \label{fig:exam_ft_loss}
\end{figure*}

\begin{figure*}
    \centering
    \subfigure[WebNLG]{
        \includegraphics[width=0.3\linewidth]{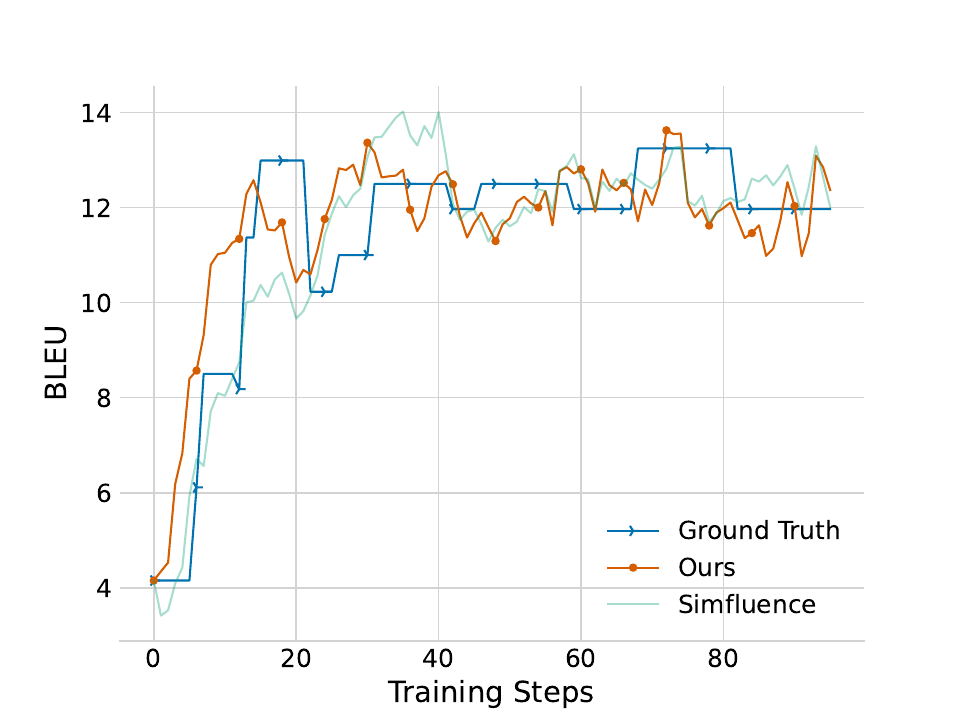}
        \includegraphics[width=0.3\linewidth]{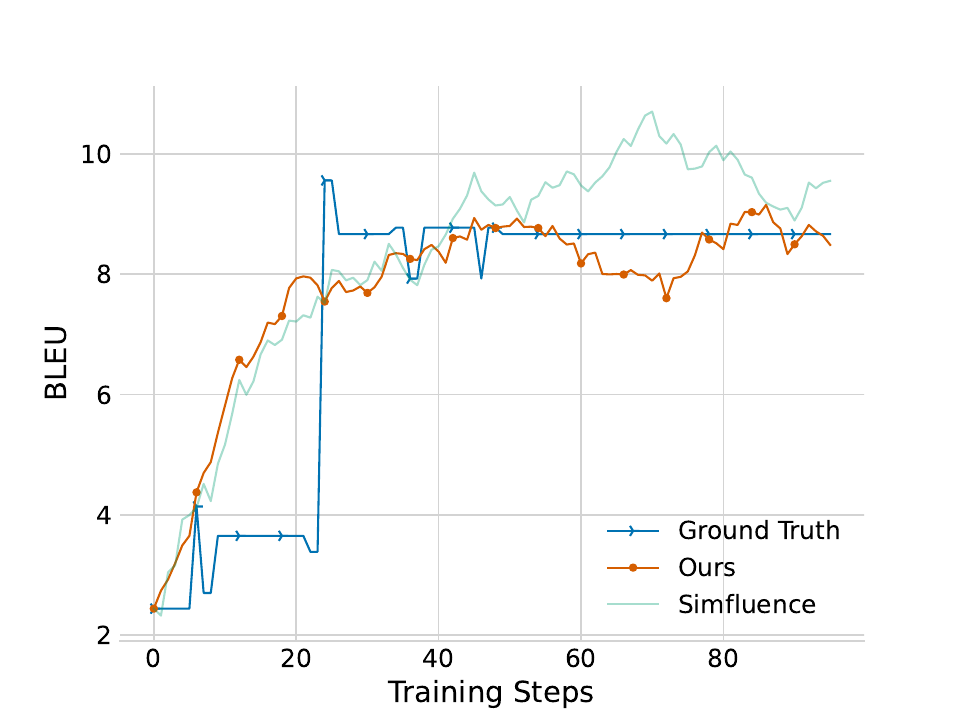}
        \includegraphics[width=0.3\linewidth]{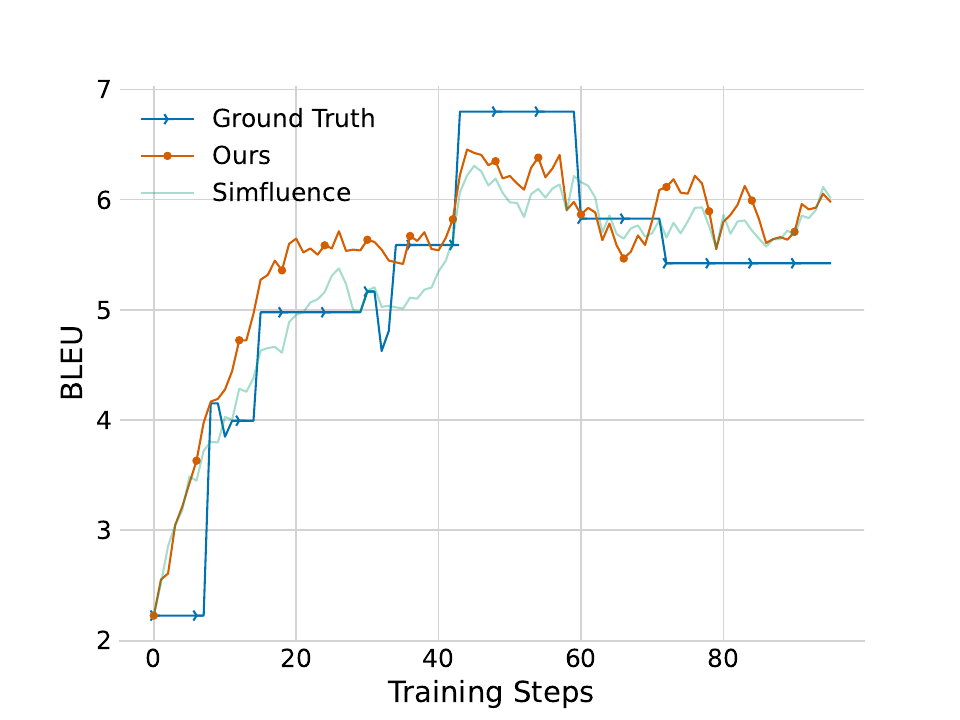}
    }
    
    \subfigure[WMT16 DE/EN]{
        \includegraphics[width=0.3\linewidth]{figure/curves/fine-tuning/bleu_410m_wmt16_de_en/__test_sample_41_run_0.pdf}
        \includegraphics[width=0.3\linewidth]{figure/curves/fine-tuning/bleu_410m_wmt16_de_en/__test_sample_45_run_0.pdf}
        \includegraphics[width=0.3\linewidth]{figure/curves/fine-tuning/bleu_410m_wmt16_de_en/__test_sample_58_run_0.pdf}
    }
    \caption{BLEU metric simulation comparison for \textit{fine-tuning} Pythia-410M using {\name} and Simfluence on the WebNLG and WMT16 DE/EN tasks.}
    \label{fig:exam_ft_bleu}
\end{figure*}

\begin{figure*}
    \centering
    \subfigure[WebNLG]{
        \includegraphics[width=0.3\linewidth]{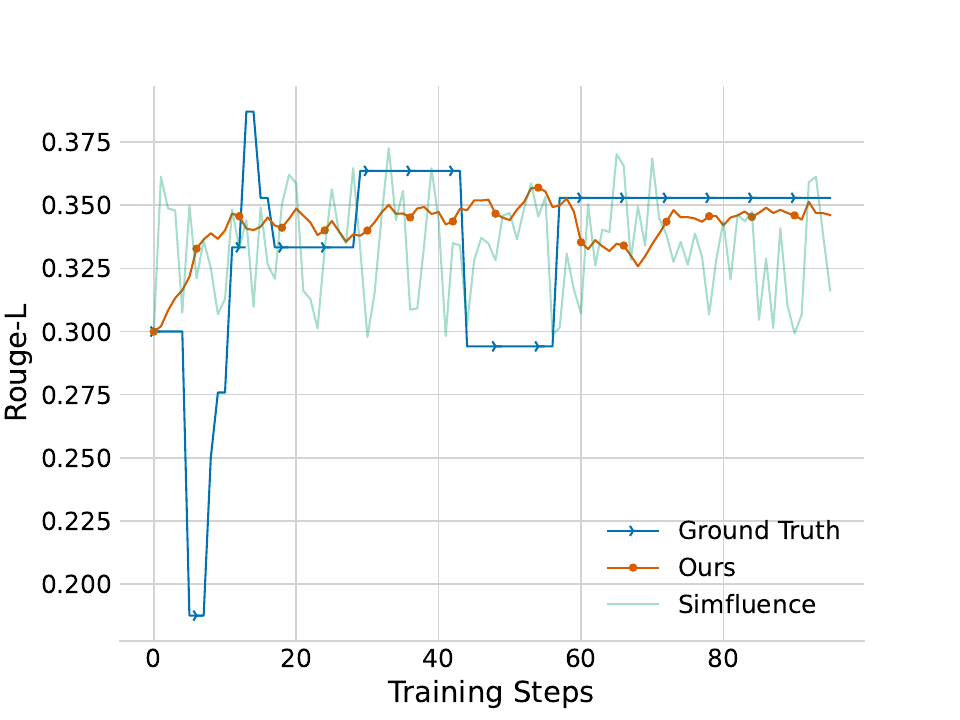}
        \includegraphics[width=0.3\linewidth]{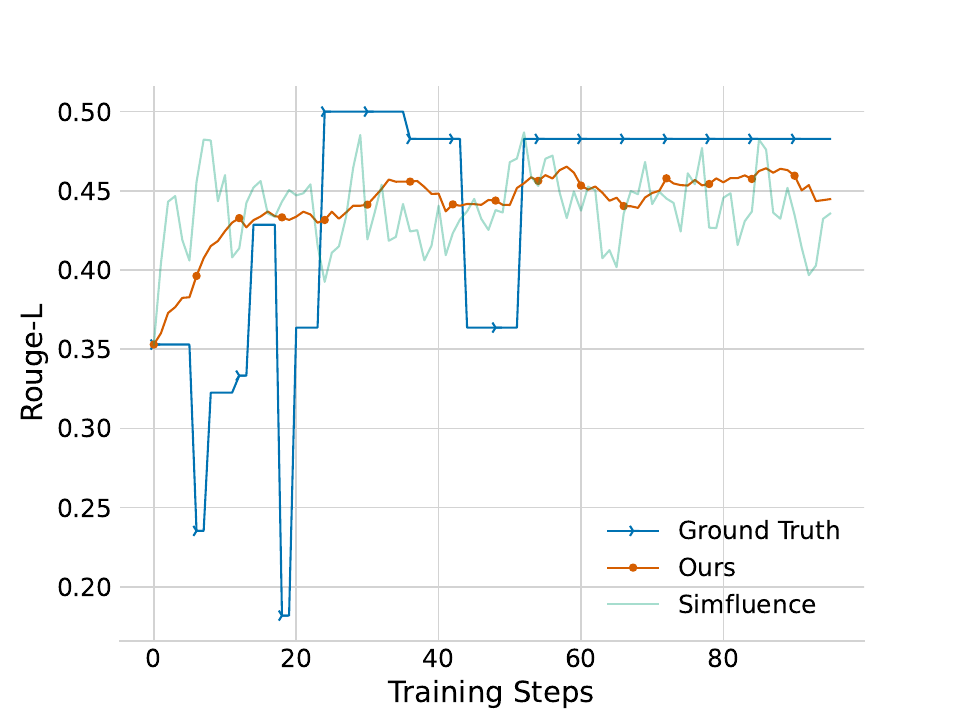}
        \includegraphics[width=0.3\linewidth]{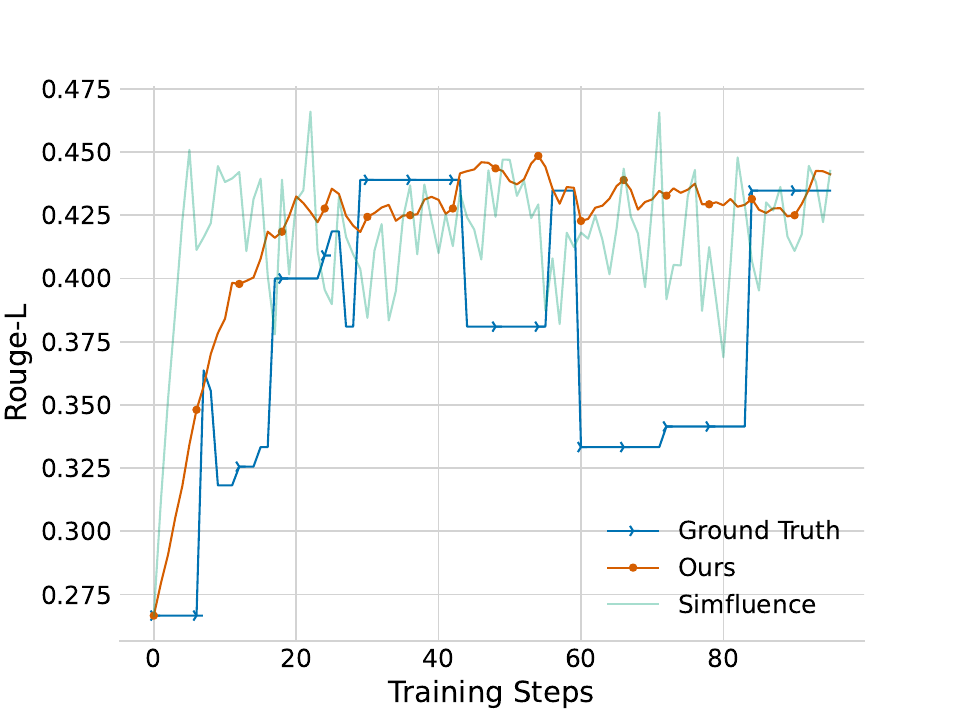}
    }
    \subfigure[WMT16 DE/EN]{
        \includegraphics[width=0.3\linewidth]{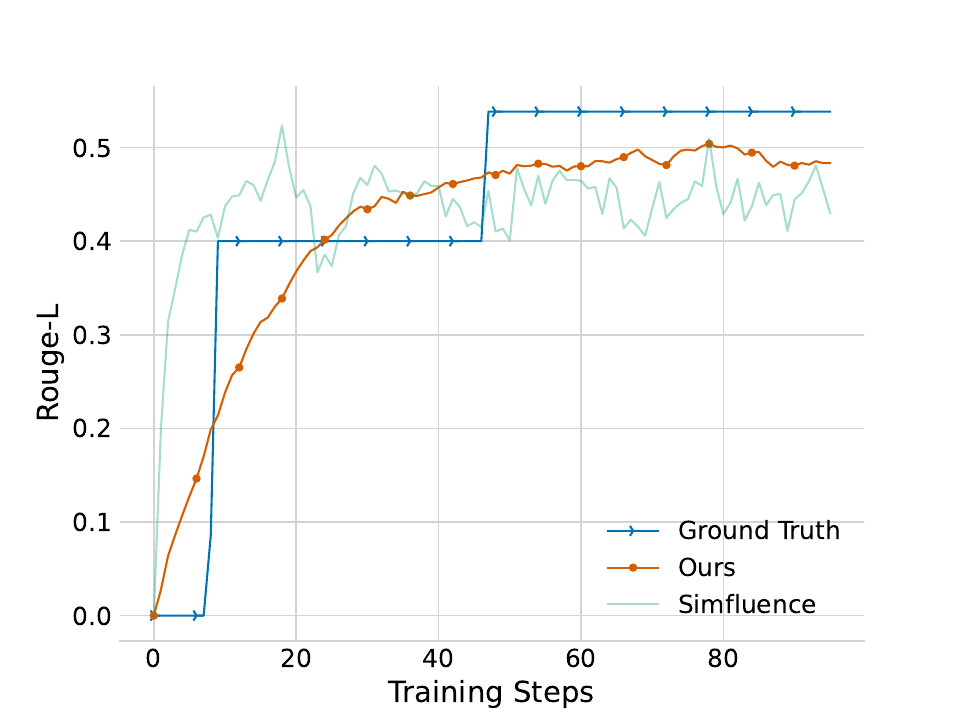}
        \includegraphics[width=0.3\linewidth]{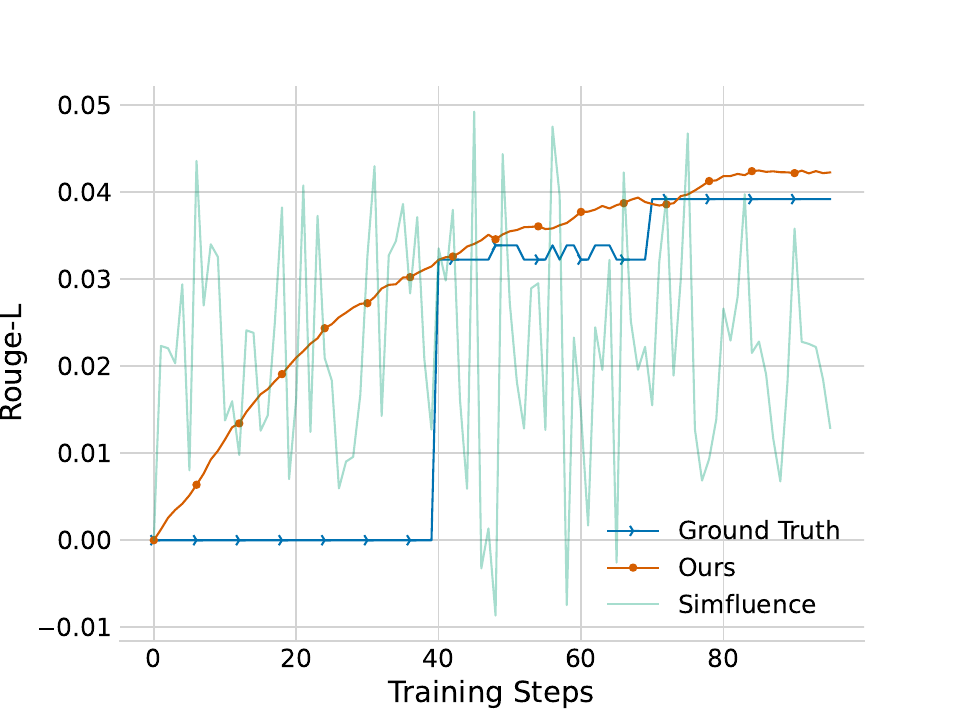}
        \includegraphics[width=0.3\linewidth]{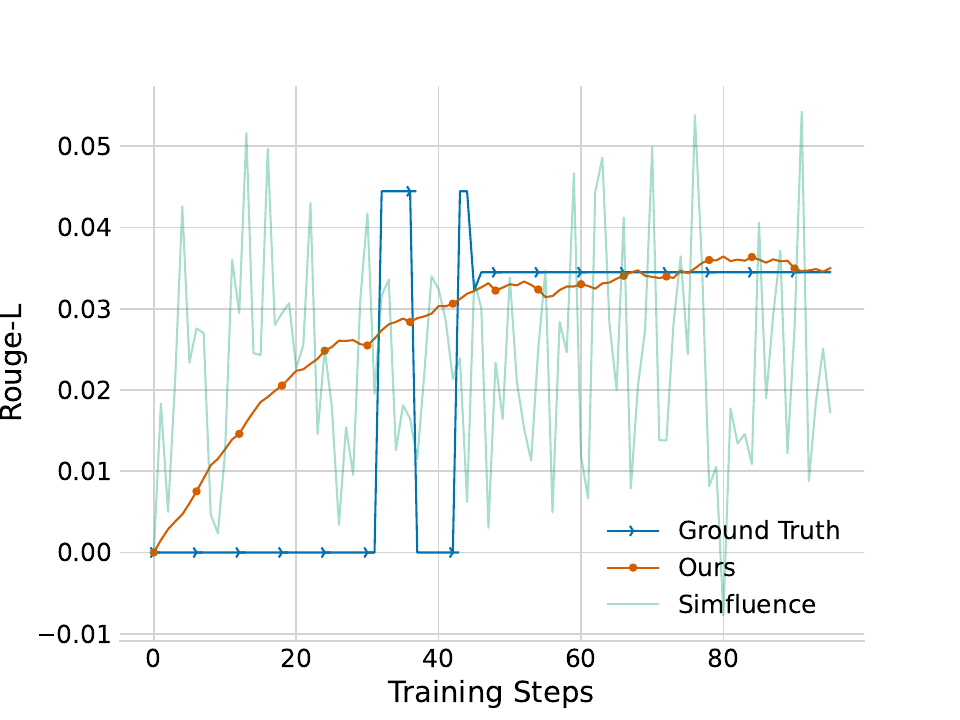}
    }
    \caption{ROUGE-L metric simulation comparison for \textit{fine-tuning} Pythia-410M with {\name} and Simfluence on the WebNLG and WMT16 DE/EN tasks.}
    \label{fig:exam_ft_rougel}
\end{figure*}

We provide additional qualitative examples showcasing simulations of test loss and performance metrics for fine-tuning, as follows:
\begin{itemize}
    \item For test loss simulation, see Fig.~\ref{fig:exam_ft_loss}.
    \item For BLEU metric simulation, refer to Fig.~\ref{fig:exam_ft_bleu}.
    \item For ROUGE-L metric simulation, see Fig.~\ref{fig:exam_ft_rougel}.
\end{itemize}

\subsection{Simuation with Unseen Data}
\label{ap:examples_unseen}

\begin{figure*}
    \centering
    \subfigure[Test Example1]{
        \includegraphics[width=0.3\linewidth]{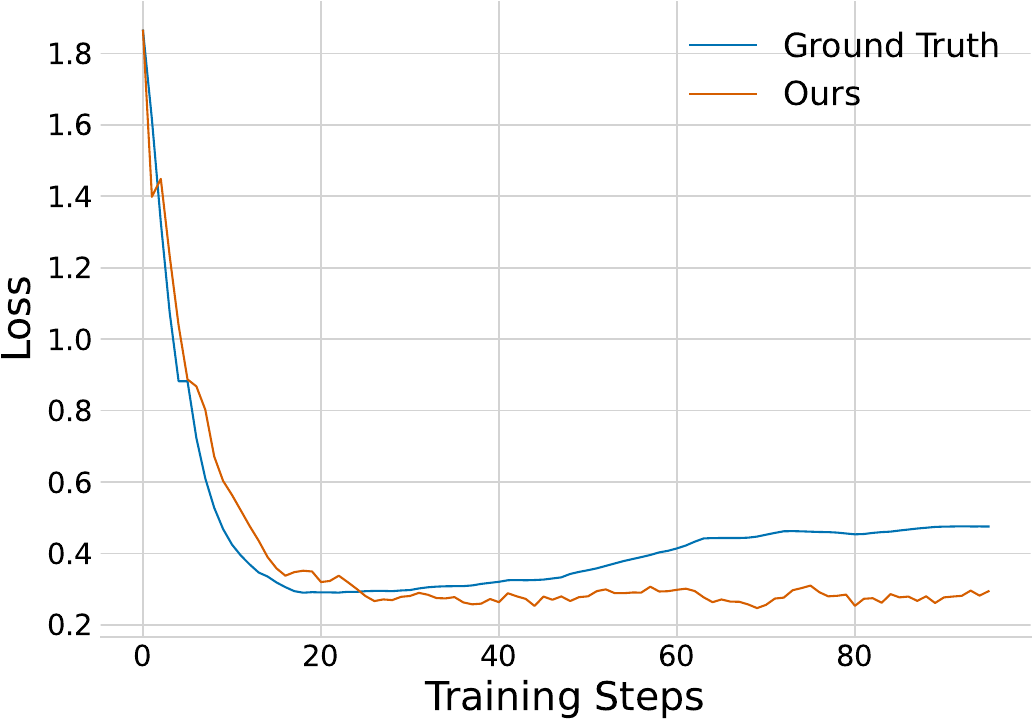}
    }
    \subfigure[Test Example2]{
        \includegraphics[width=0.3\linewidth]{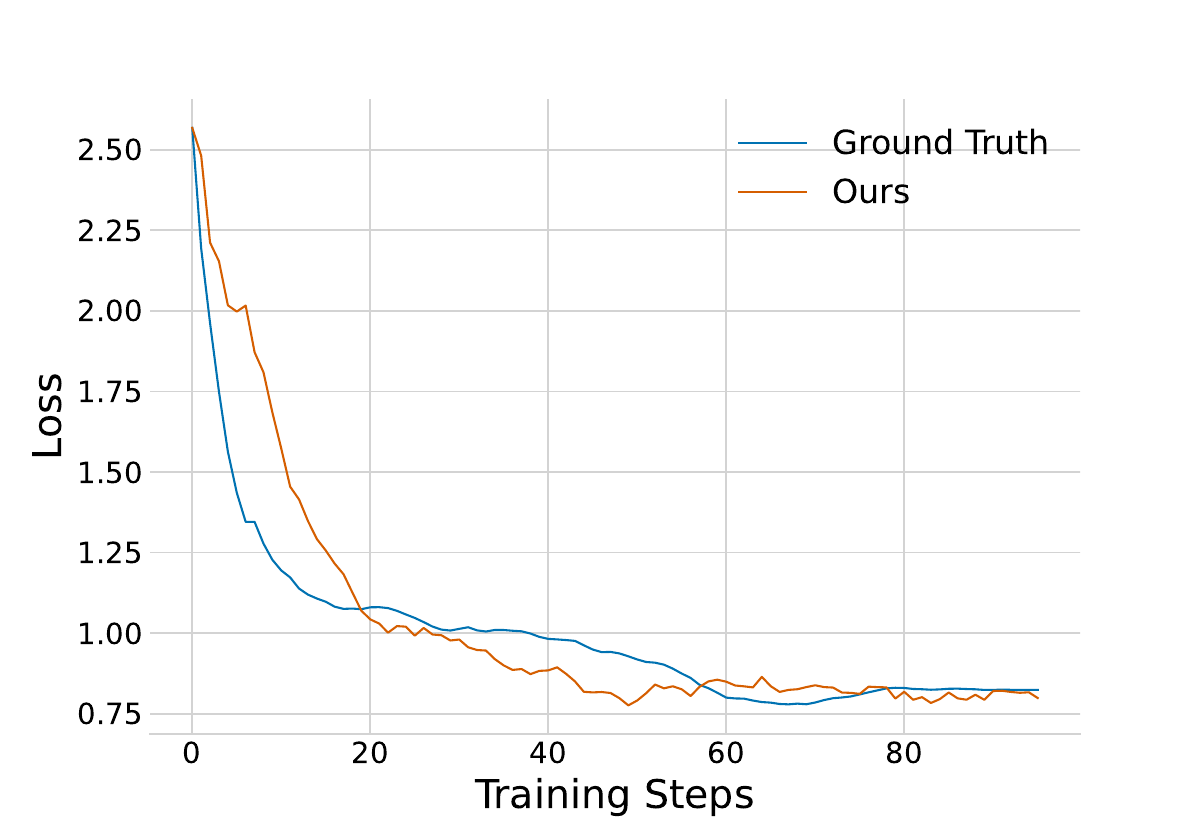}
    }
    \subfigure[Test Example3]{
        \includegraphics[width=0.3\linewidth]{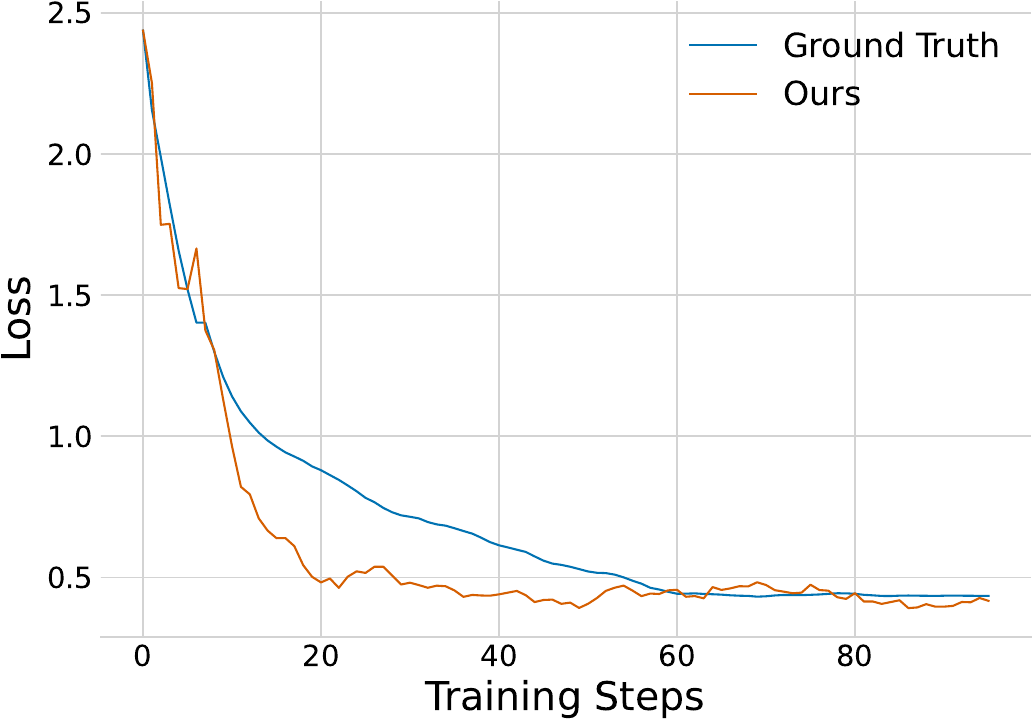}
    }   
    \caption{Examples of loss simulation of {\name} for the RTE task on \textbf{\textit{unseen test data}}.}
    \label{fig:rte_unseen_test}
\end{figure*}

\begin{figure*}
    \centering
    \subfigure[Test Example1]{
        \includegraphics[width=0.3\linewidth]{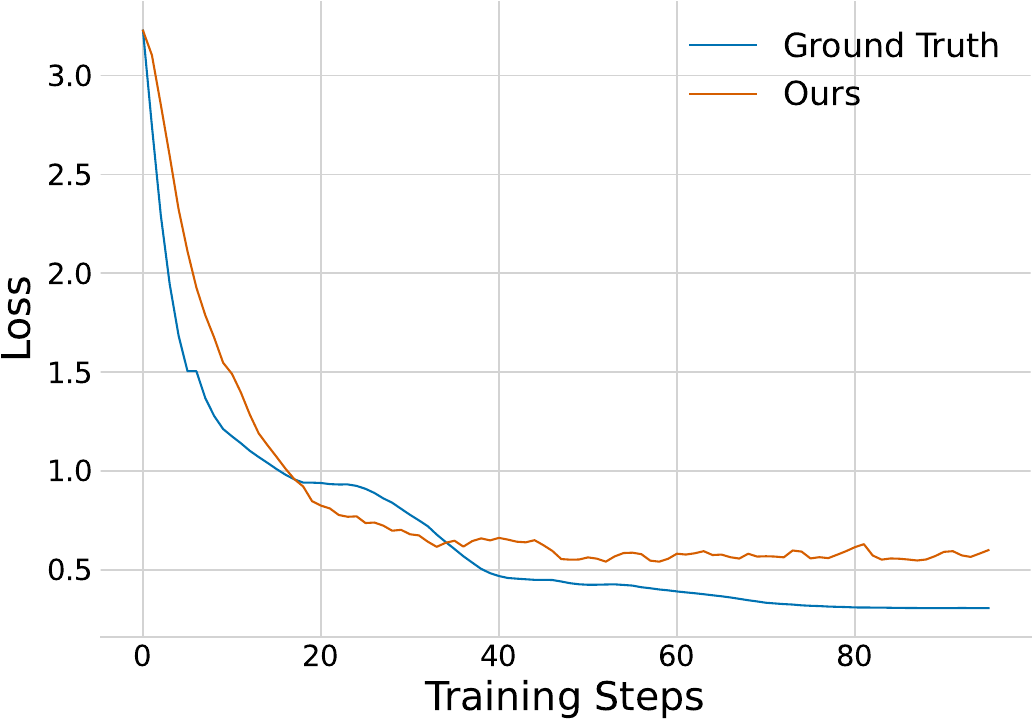}
    }
    \subfigure[Test Example2]{
        \includegraphics[width=0.3\linewidth]{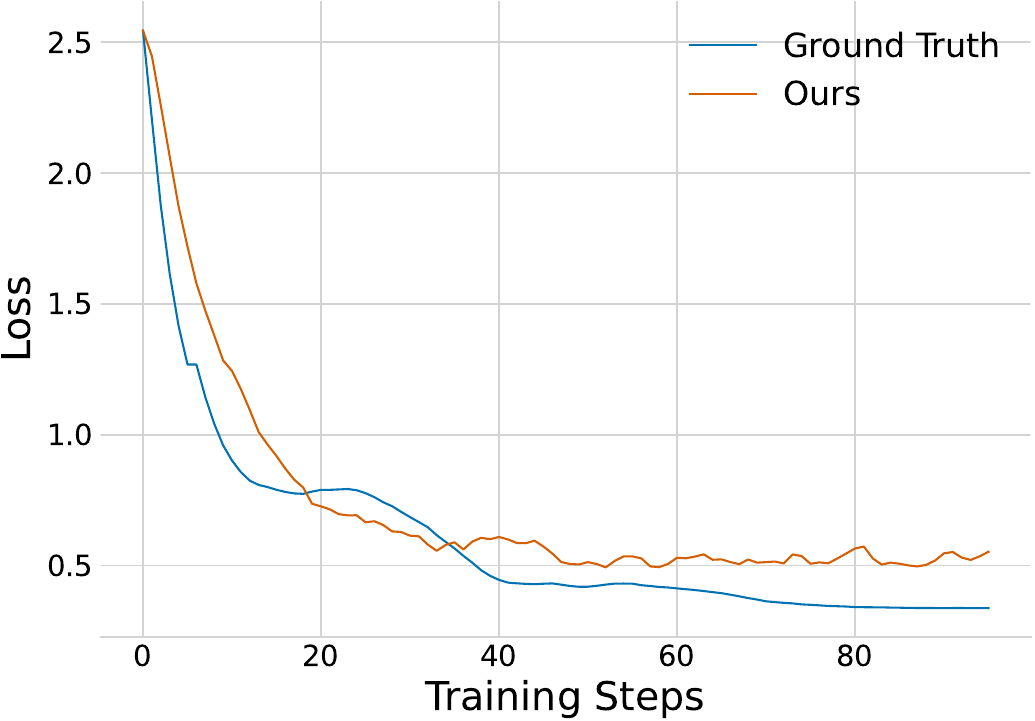}
    }
    \subfigure[Test Example3]{
        \includegraphics[width=0.3\linewidth]{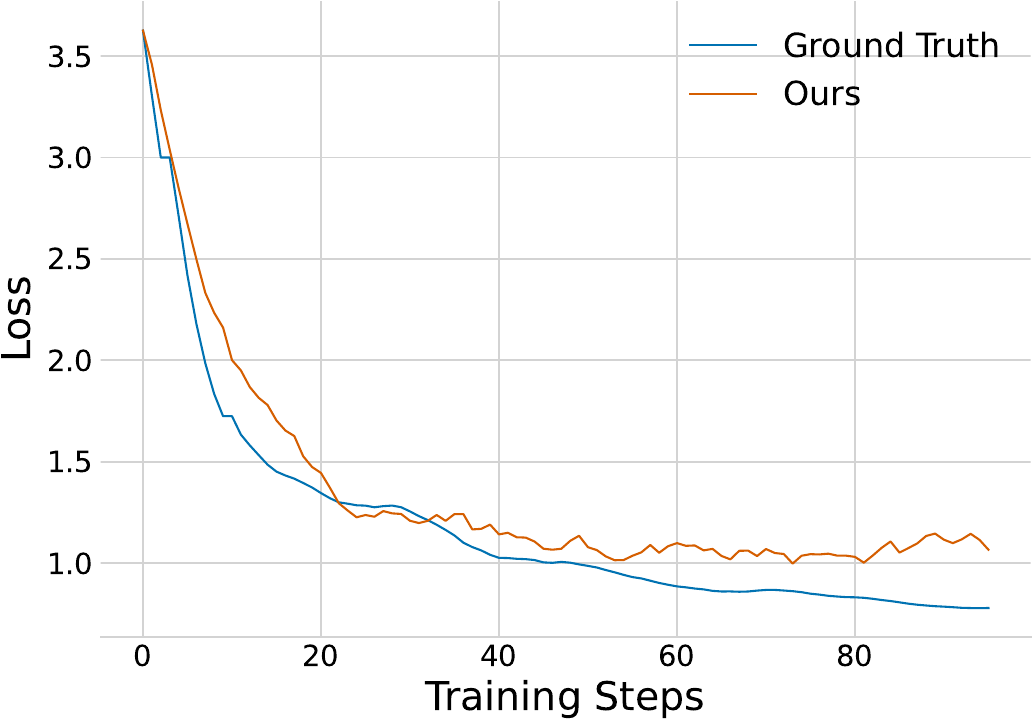}
    }   
    \caption{Examples of loss simulation of {\name} for the RTE task on \textbf{\textit{unseen training data}}.}
    \label{fig:rte_unseen_training}
\end{figure*}

\begin{figure*}
    \centering
    \subfigure[Test Example1]{
        \includegraphics[width=0.3\linewidth]{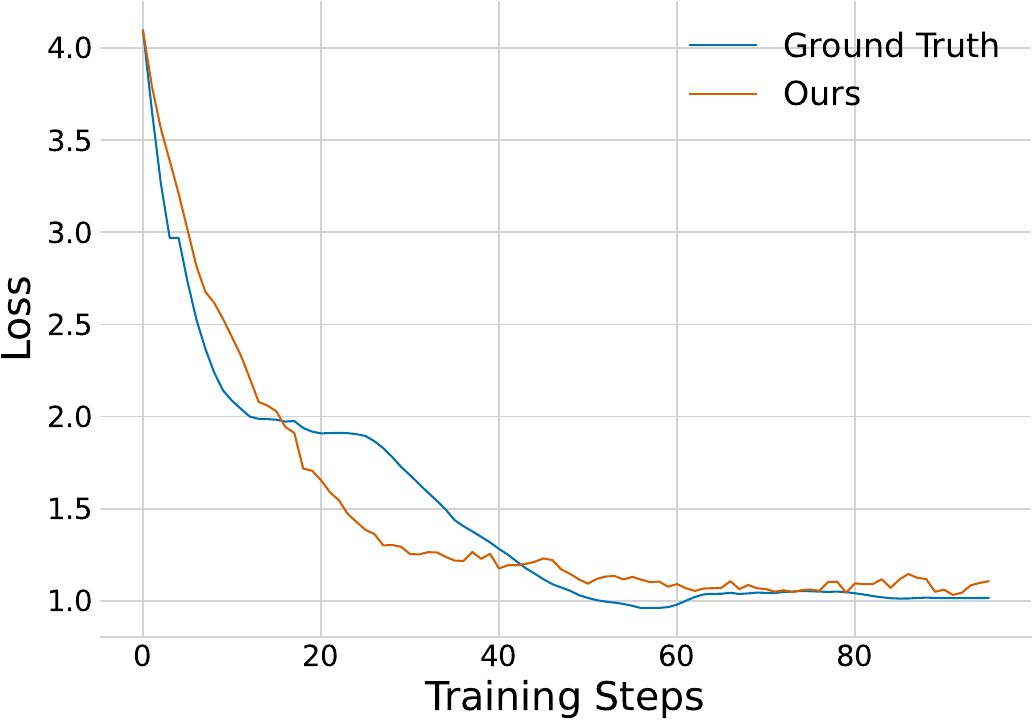}
    }
    \subfigure[Test Example2]{
        \includegraphics[width=0.3\linewidth]{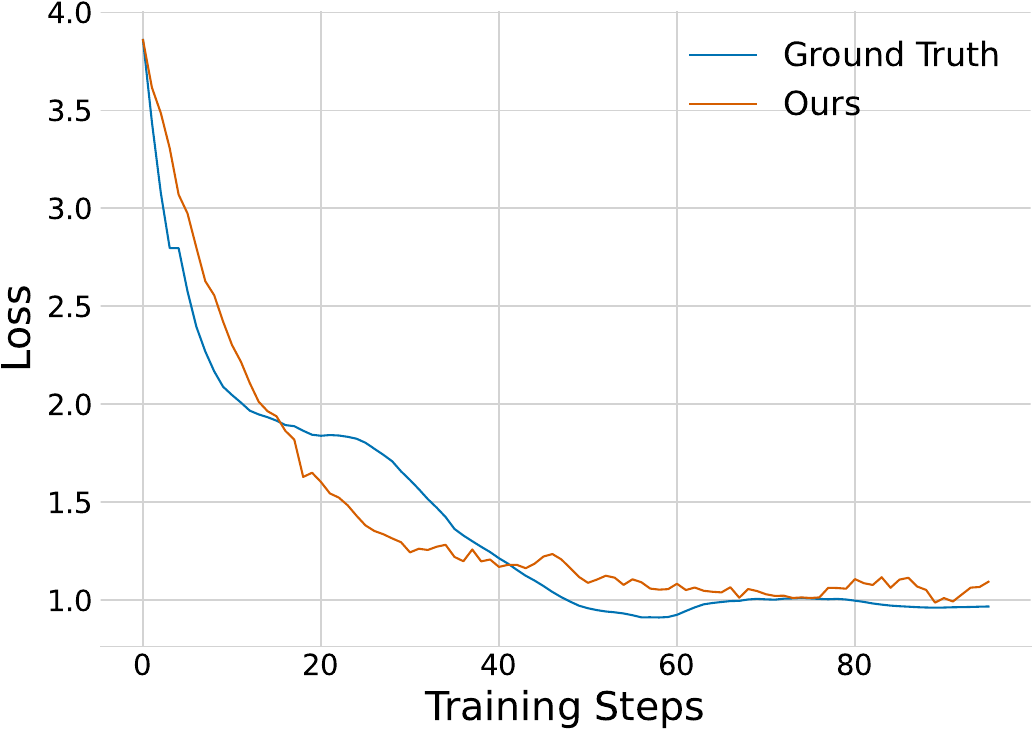}
    }
    \subfigure[Test Example3]{
        \includegraphics[width=0.3\linewidth]{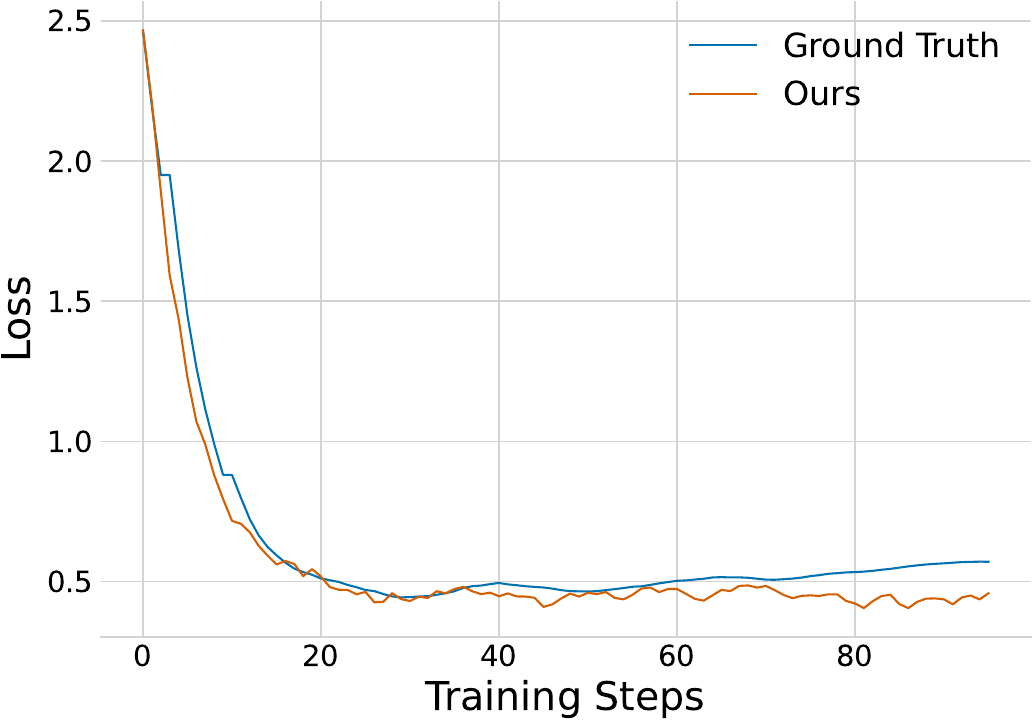}
    }   
    \caption{Examples of loss simulation of {\name} for the RTE task on \textbf{\textit{unseen training and test data}}.}
    \label{fig:rte_unseen_training_test}
\end{figure*}

\begin{figure*}
    \centering
    \subfigure[Test Example1]{
        \includegraphics[width=0.3\linewidth]{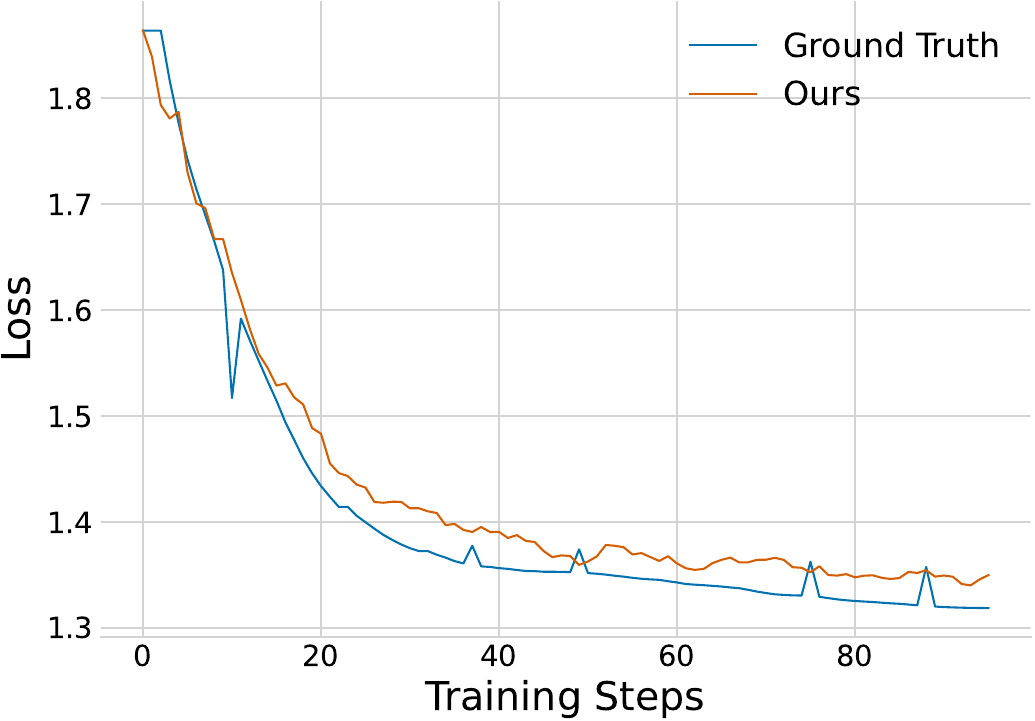}
    }
    \subfigure[Test Example2]{
        \includegraphics[width=0.3\linewidth]{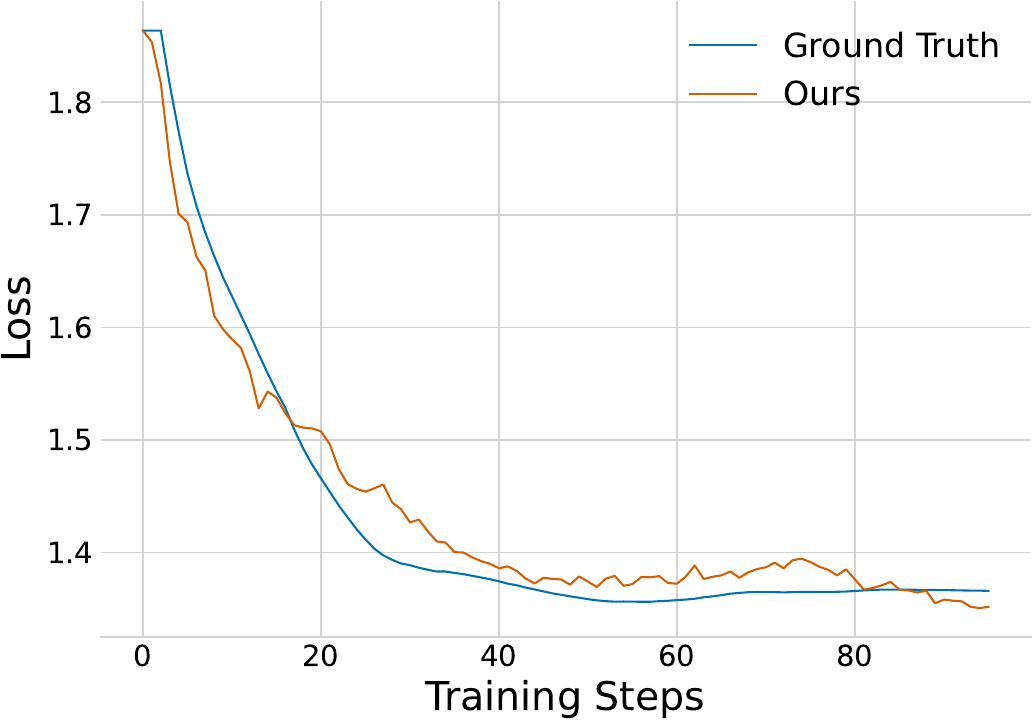}
    }
    \subfigure[Test Example3]{
        \includegraphics[width=0.3\linewidth]{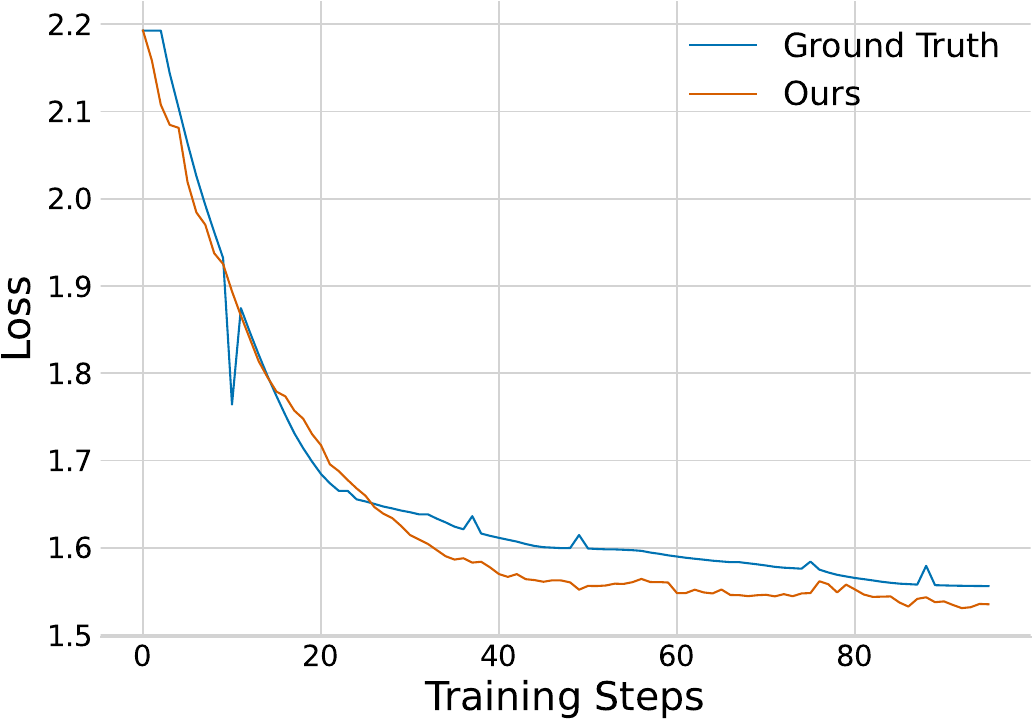}
    }   
    \caption{Examples of loss simulation of {\name} for the WebNLG task on \textbf{\textit{unseen test data}}.}
    \label{fig:webnlg_loss_unseen_test}
\end{figure*}

\begin{figure*}
    \centering
    \subfigure[Test Example1]{
        \includegraphics[width=0.3\linewidth]{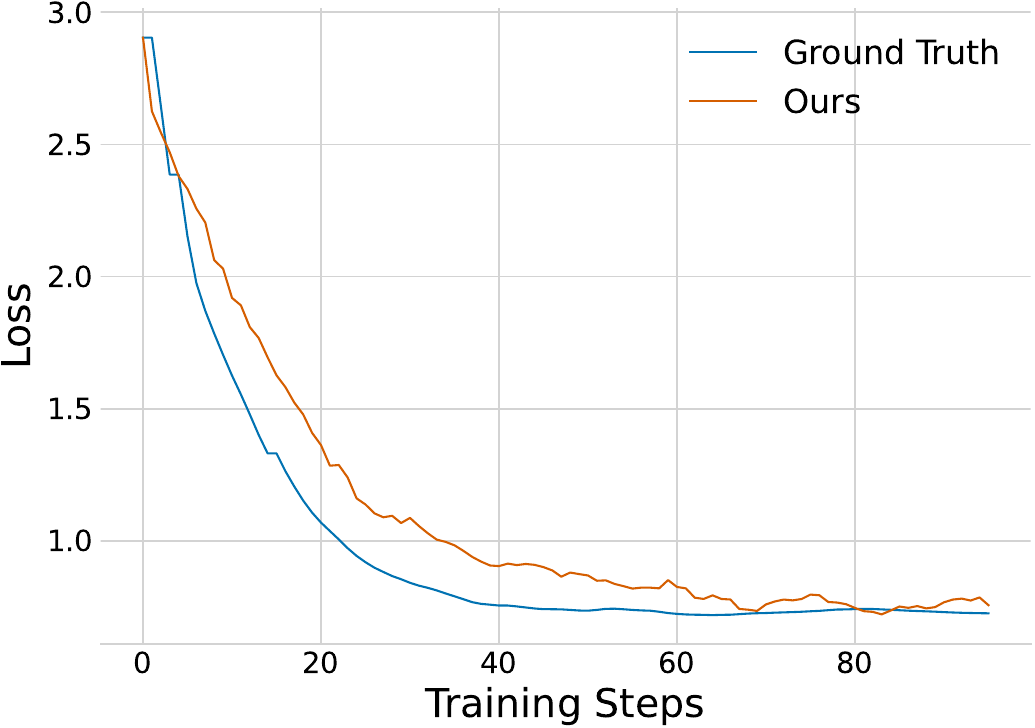}
    }
    \subfigure[Test Example2]{
        \includegraphics[width=0.3\linewidth]{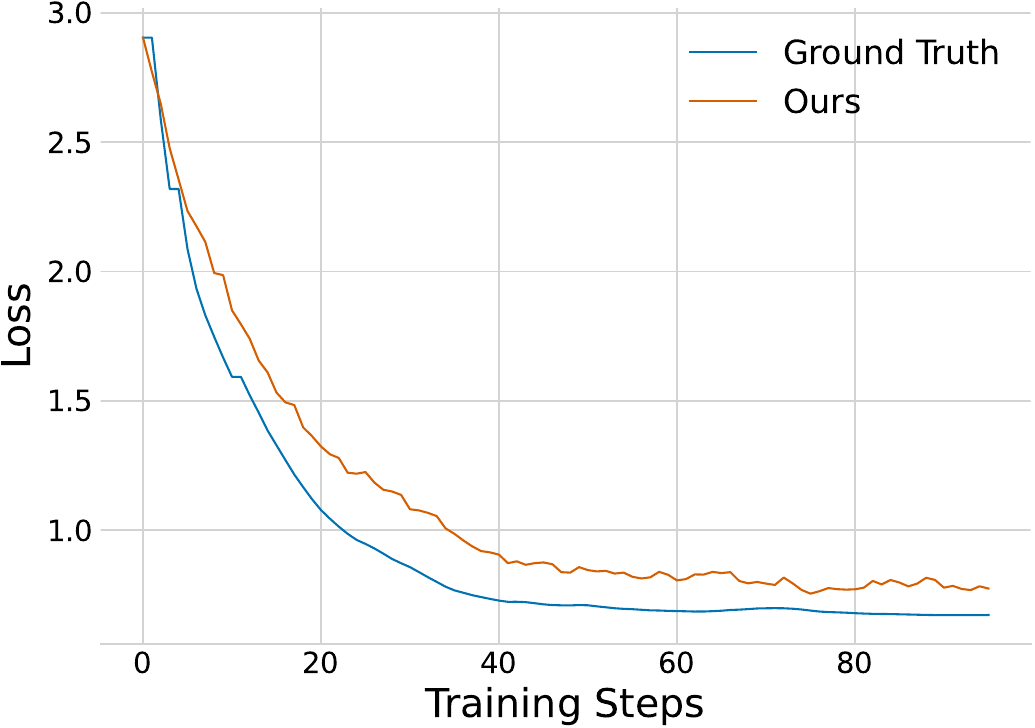}
    }
    \subfigure[Test Example3]{
        \includegraphics[width=0.3\linewidth]{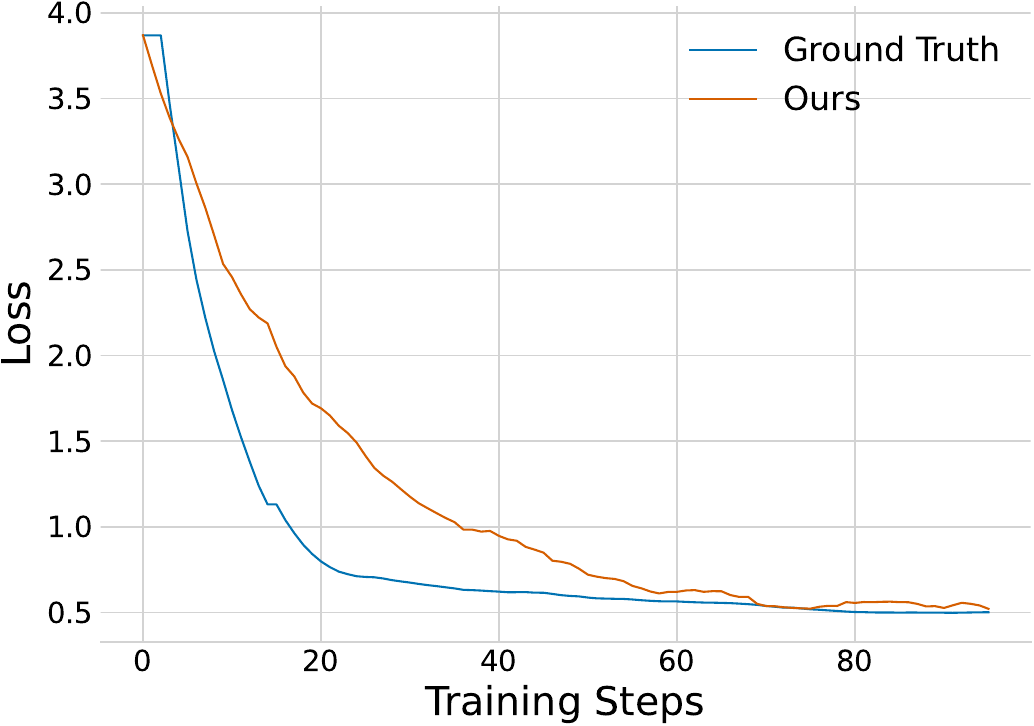}
    }   
    \caption{Examples of loss simulation of {\name} for the WebNLG task on \textbf{\textit{unseen training data}}.}
    \label{fig:webnlg_loss_unseen_training}
\end{figure*}

\begin{figure*}
    \centering
    \subfigure[Test Example1]{
        \includegraphics[width=0.3\linewidth]{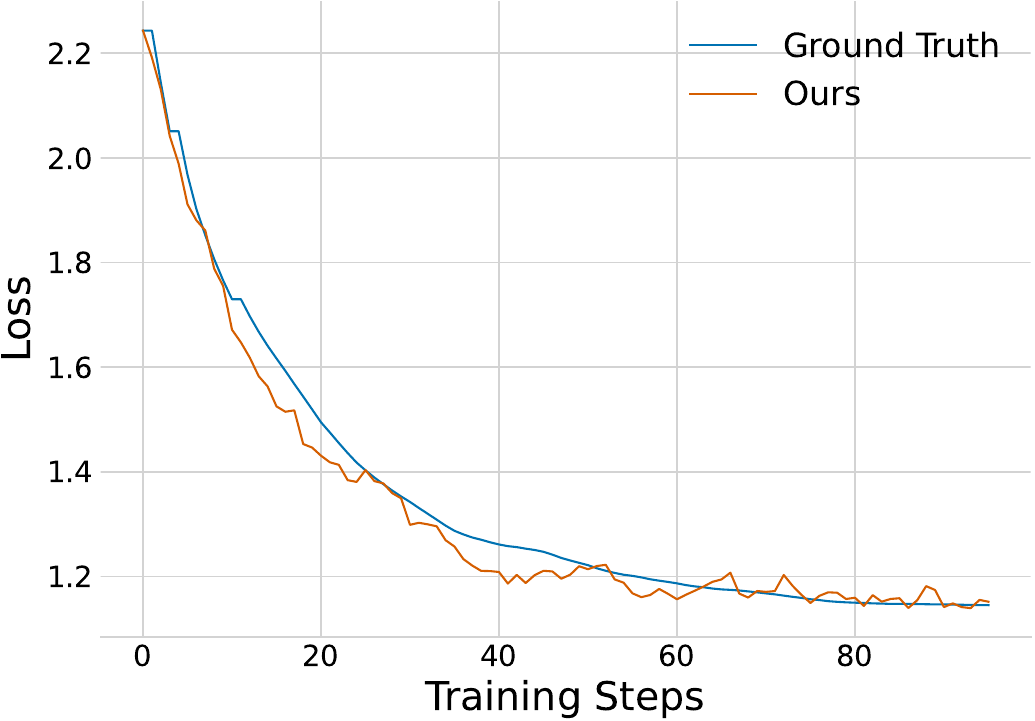}
    }
    \subfigure[Test Example2]{
        \includegraphics[width=0.3\linewidth]{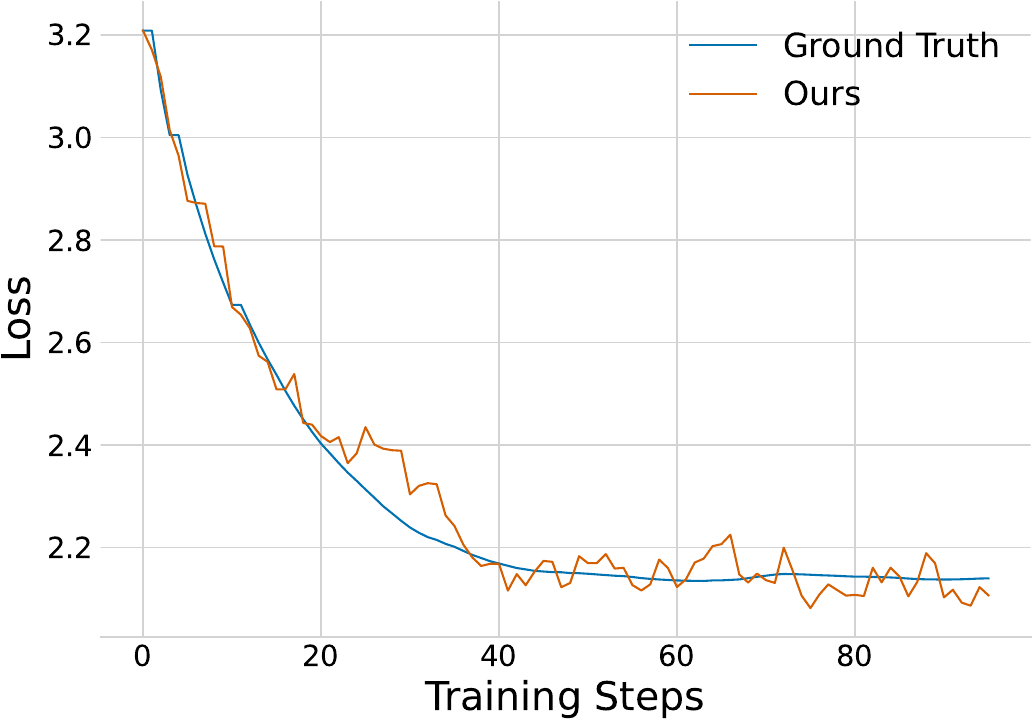}
    }
    \subfigure[Test Example3]{
        \includegraphics[width=0.3\linewidth]{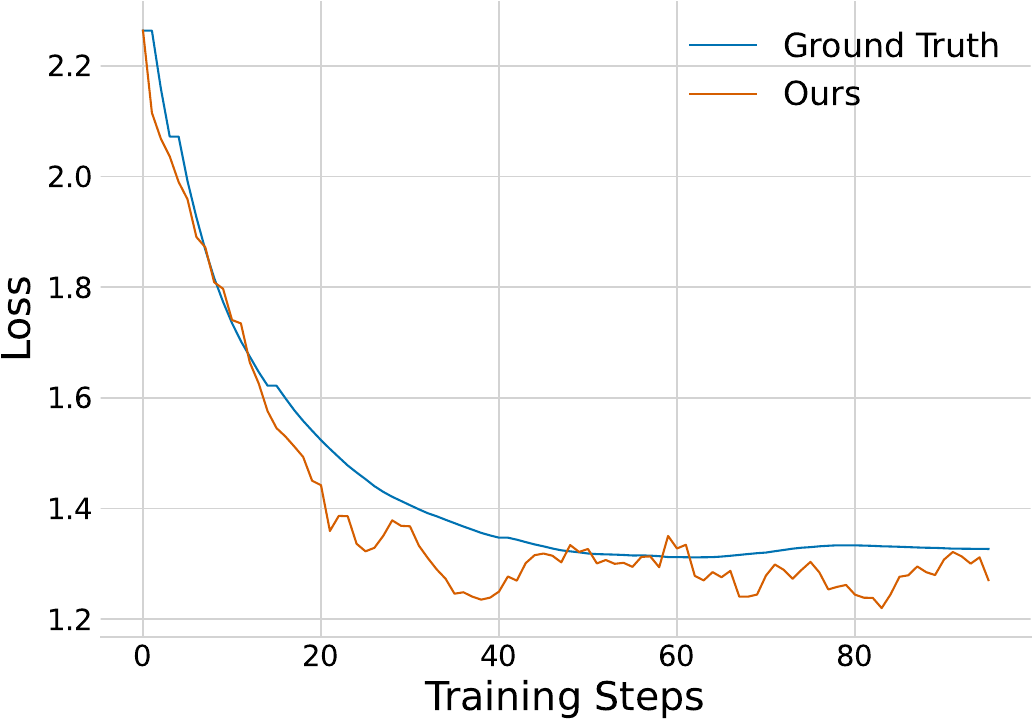}
    }   
    \caption{Examples of loss simulation of {\name} for the WebNLG task on \textbf{\textit{unseen training and test data}}.}
    \label{fig:webnlg_loss_unseen_training_test}
\end{figure*}

\begin{figure*}
    \centering
    \subfigure[Test Example1]{
        \includegraphics[width=0.3\linewidth]{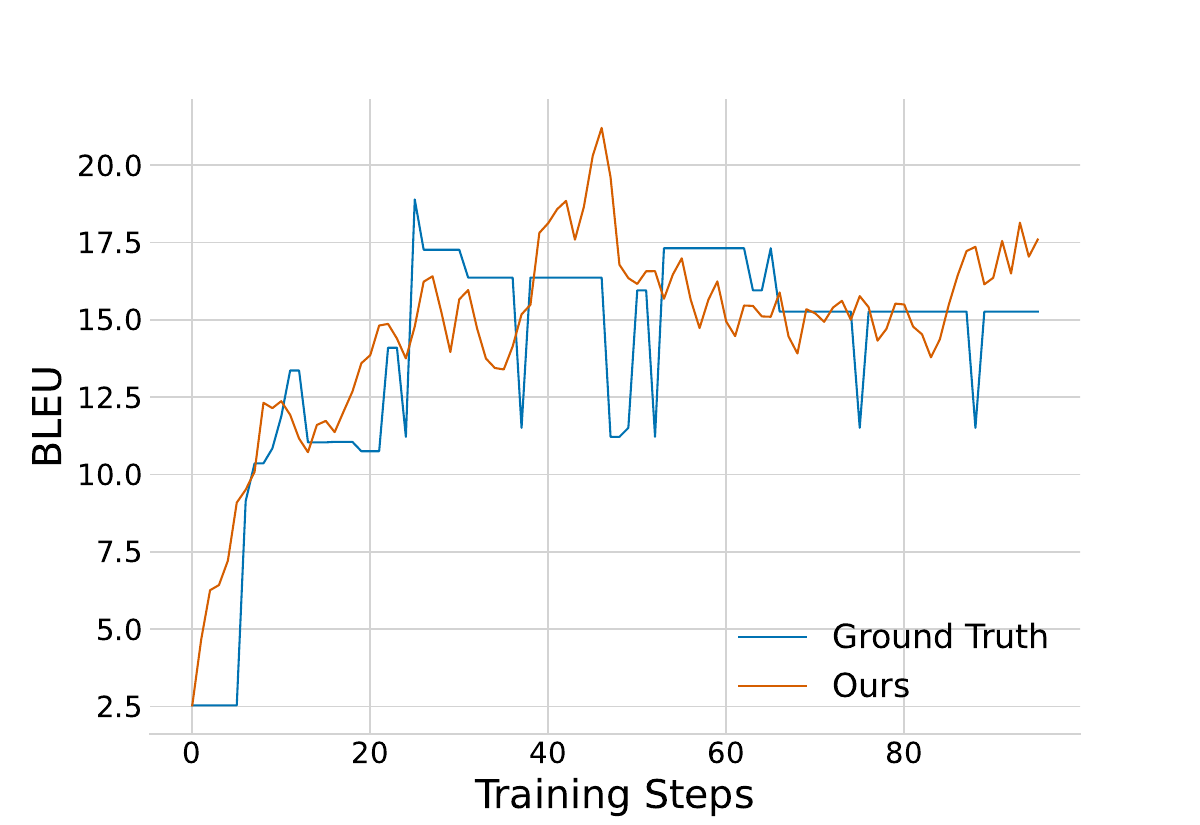}
    }
    \subfigure[Test Example2]{
        \includegraphics[width=0.3\linewidth]{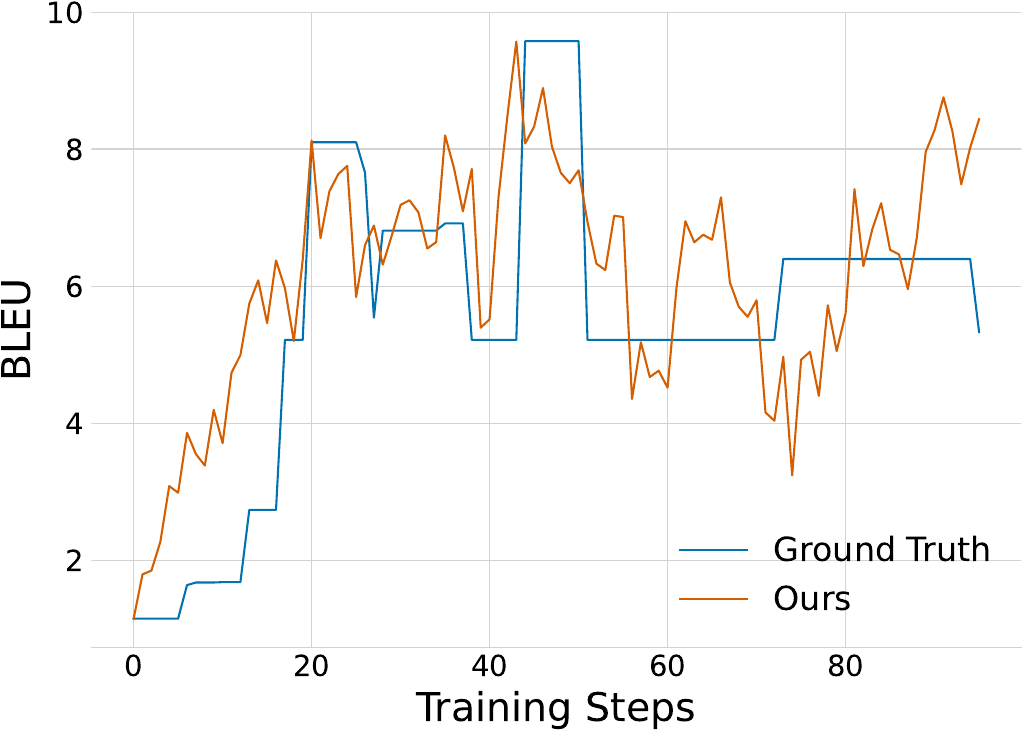}
    }
    \subfigure[Test Example3]{
        \includegraphics[width=0.3\linewidth]{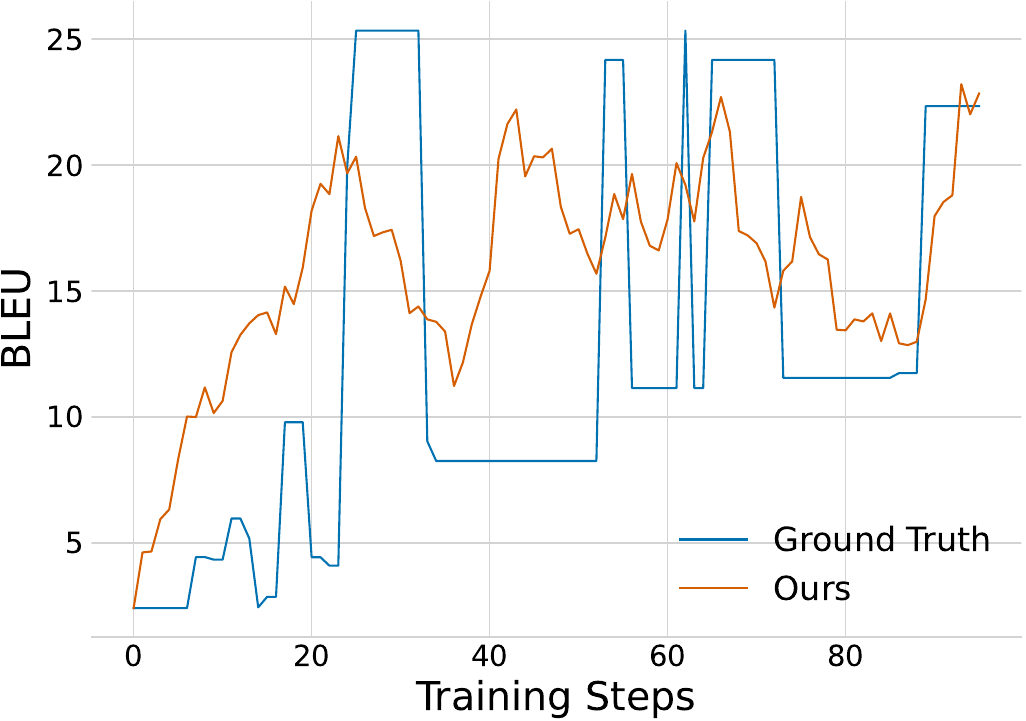}
    }   
    \caption{Examples of BLEU simulation of {\name} for the WebNLG task on \textbf{\textit{unseen test data}}.}
    \label{fig:webnlg_bleu_unseen_test}
\end{figure*}

\begin{figure*}
    \centering
    \subfigure[Test Example1]{
        \includegraphics[width=0.3\linewidth]{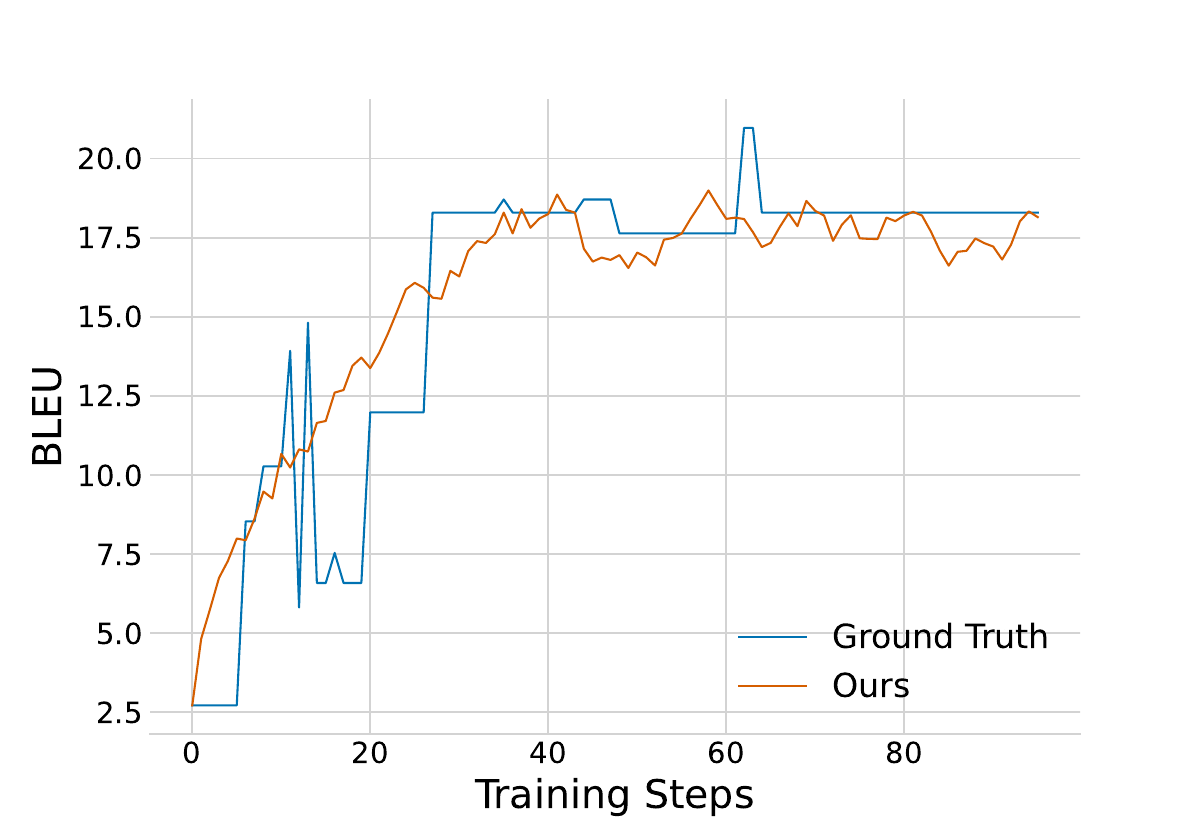}
    }
    \subfigure[Test Example2]{
        \includegraphics[width=0.3\linewidth]{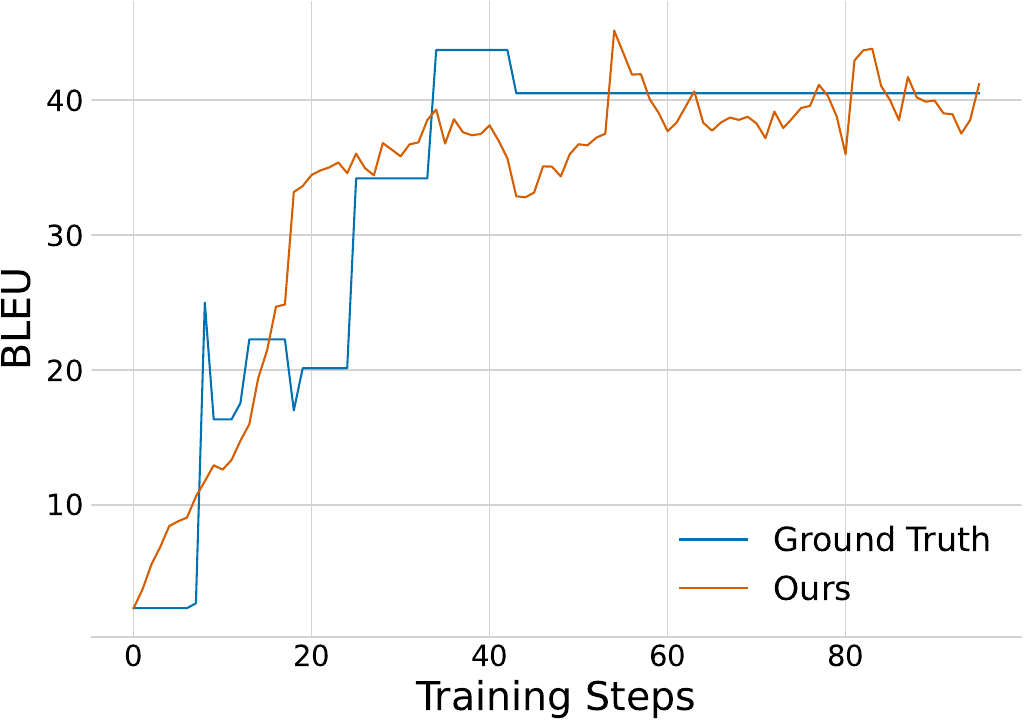}
    }
    \subfigure[Test Example3]{
        \includegraphics[width=0.3\linewidth]{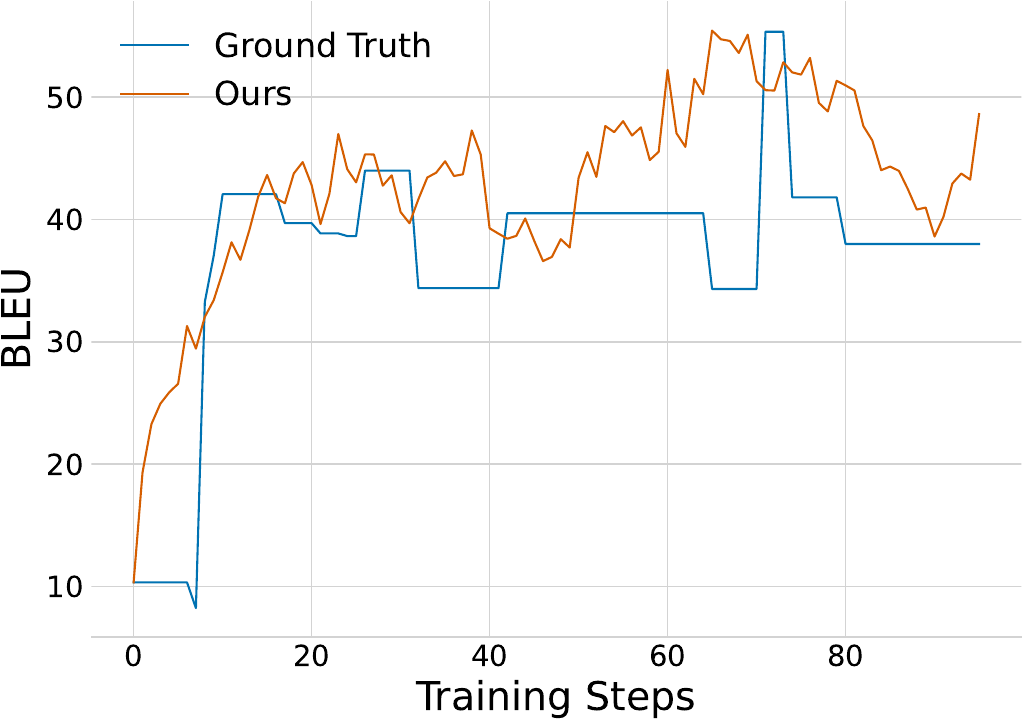}
    }   
    \caption{Examples of BLEU simulation of {\name} for the WebNLG task on \textbf{\textit{unseen training data}}.}
    \label{fig:webnlg_bleu_unseen_training}
\end{figure*}

\begin{figure*}
    \centering
    \subfigure[Test Example1]{
        \includegraphics[width=0.3\linewidth]{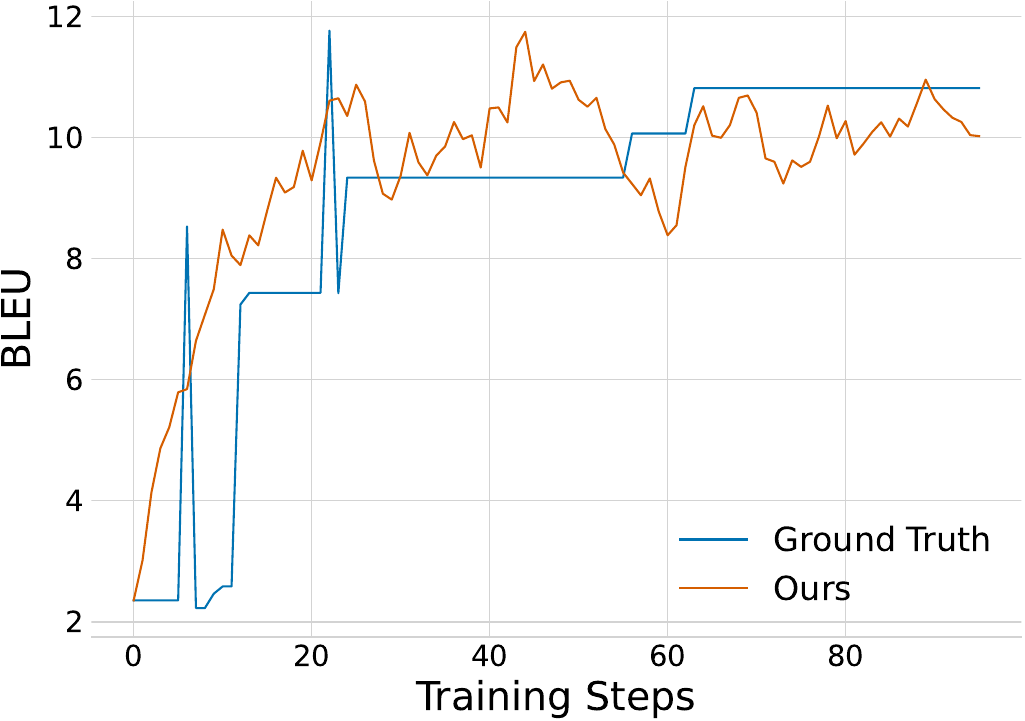}
    }
    \subfigure[Test Example2]{
        \includegraphics[width=0.3\linewidth]{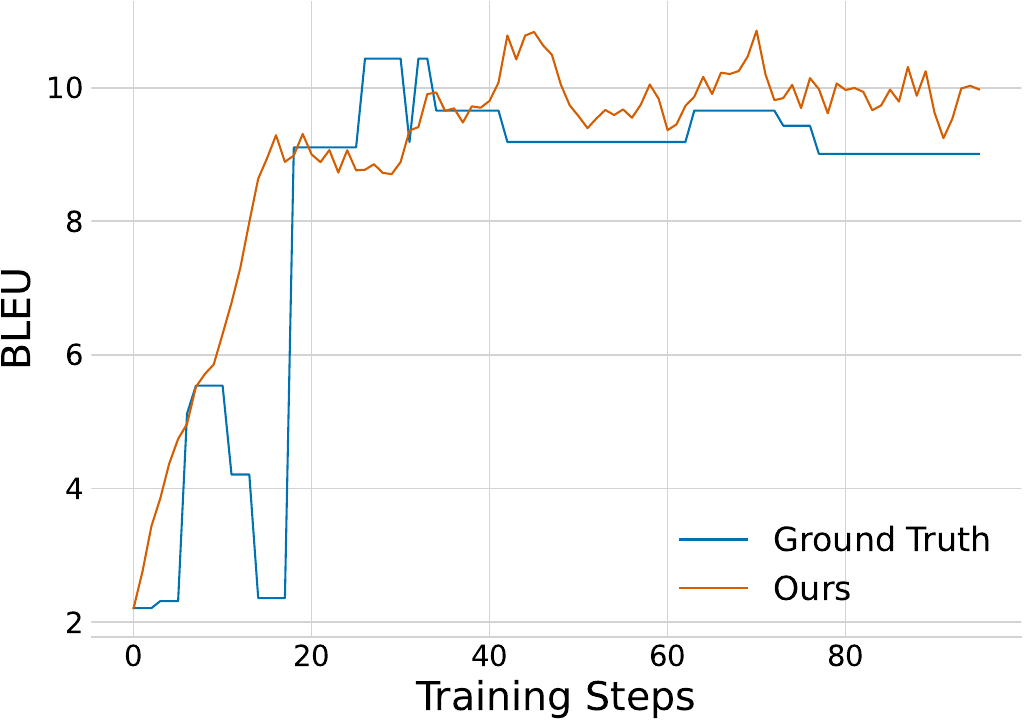}
    }
    \subfigure[Test Example3]{
        \includegraphics[width=0.3\linewidth]{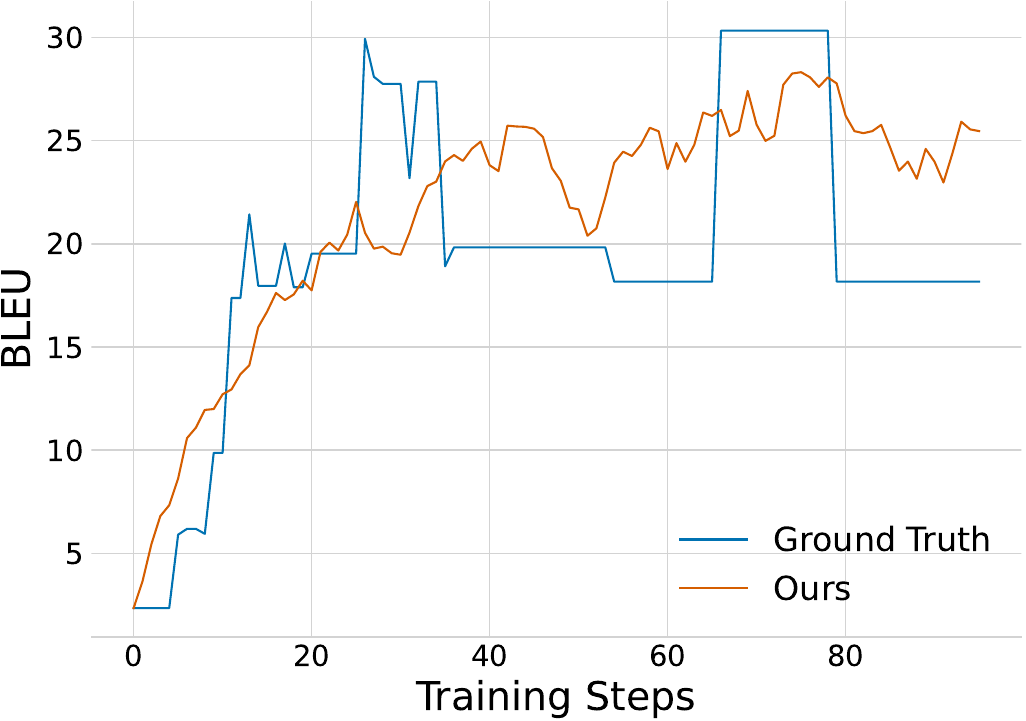}
    }   
    \caption{Examples of BLEU simulation of {\name} for the WebNLG task on \textbf{\textit{unseen training and test data}}.}
    \label{fig:webnlg_bleu_unseen_training_test}
\end{figure*}

\begin{figure*}
    \centering
    \subfigure[Test Example1]{
        \includegraphics[width=0.3\linewidth]{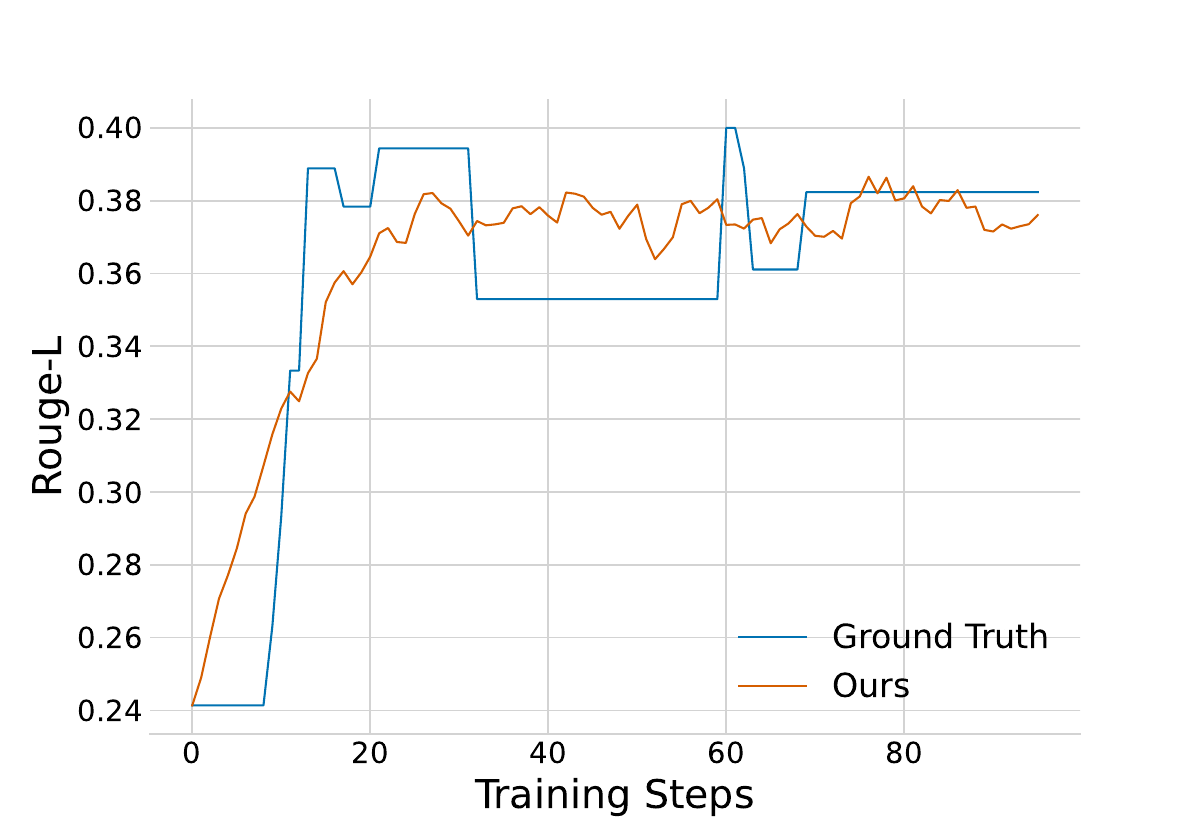}
    }
    \subfigure[Test Example2]{
        \includegraphics[width=0.3\linewidth]{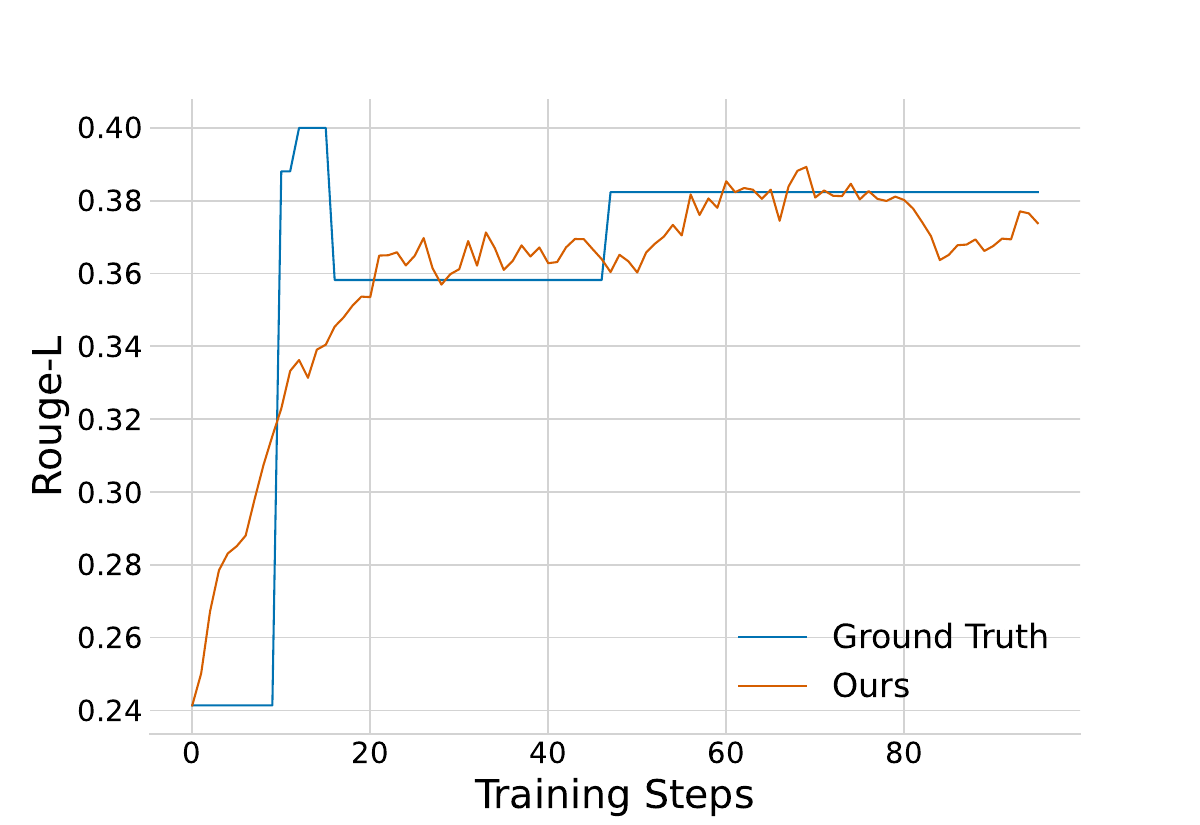}
    }
    \subfigure[Test Example3]{
        \includegraphics[width=0.3\linewidth]{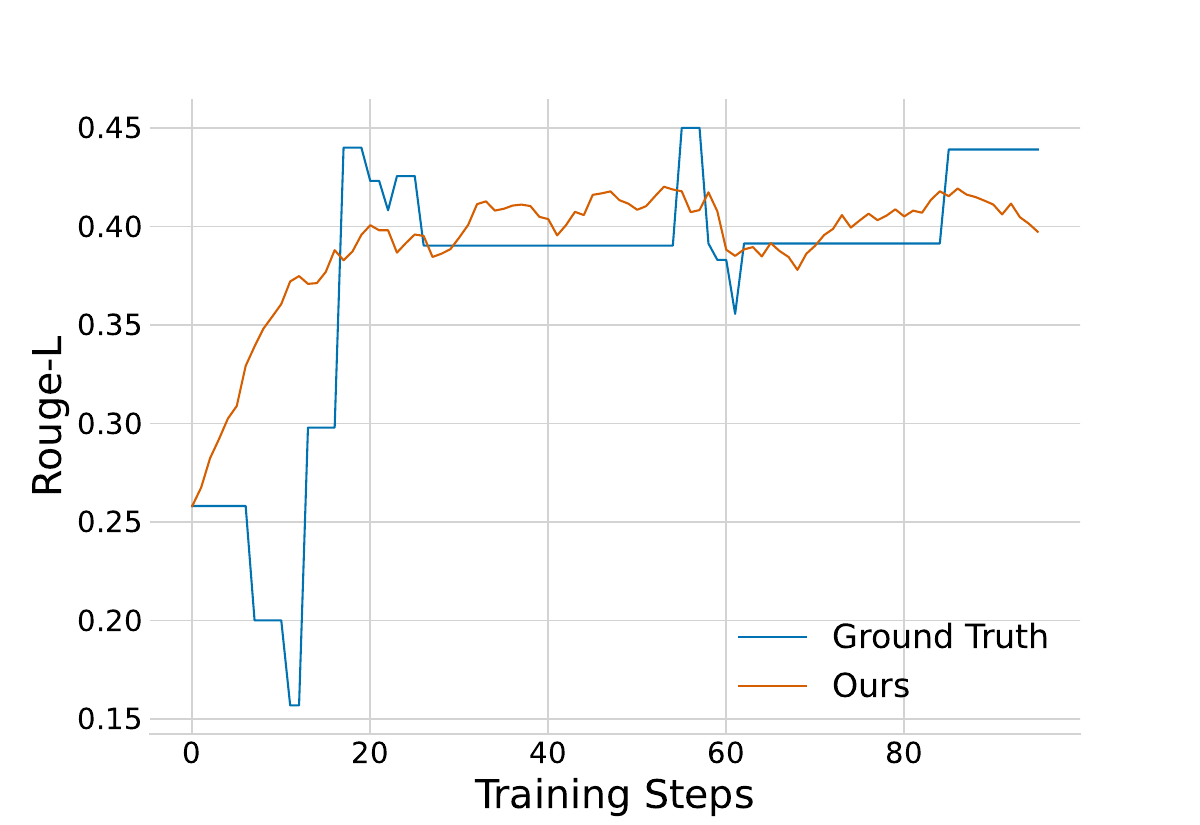}
    }   
    \caption{Examples of the ROUGE-L simulation of {\name}  for the WebNLG task on \textbf{\textit{unseen test data}}.}
    \label{fig:webnlg_rougeL_unseen_test}
\end{figure*}

\begin{figure*}
    \centering
    \subfigure[Test Example1]{
        \includegraphics[width=0.3\linewidth]{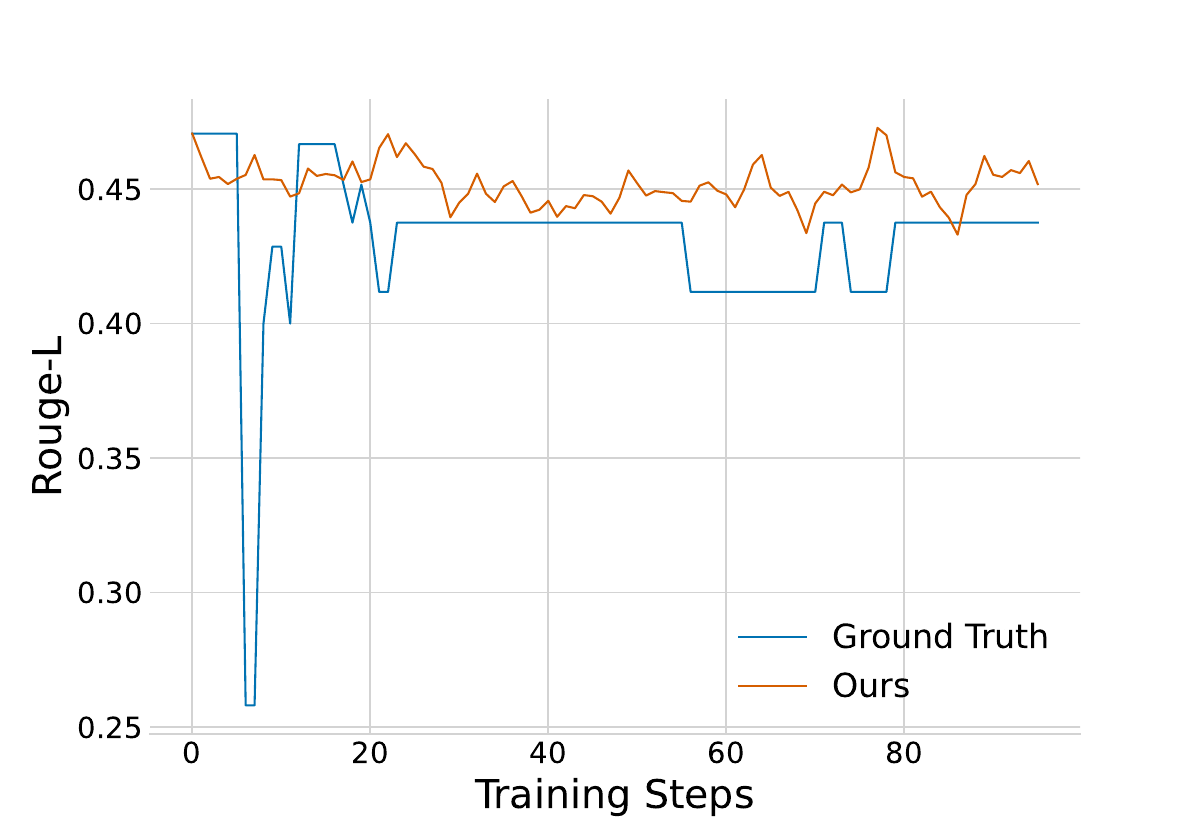}
    }
    \subfigure[Test Example2]{
        \includegraphics[width=0.3\linewidth]{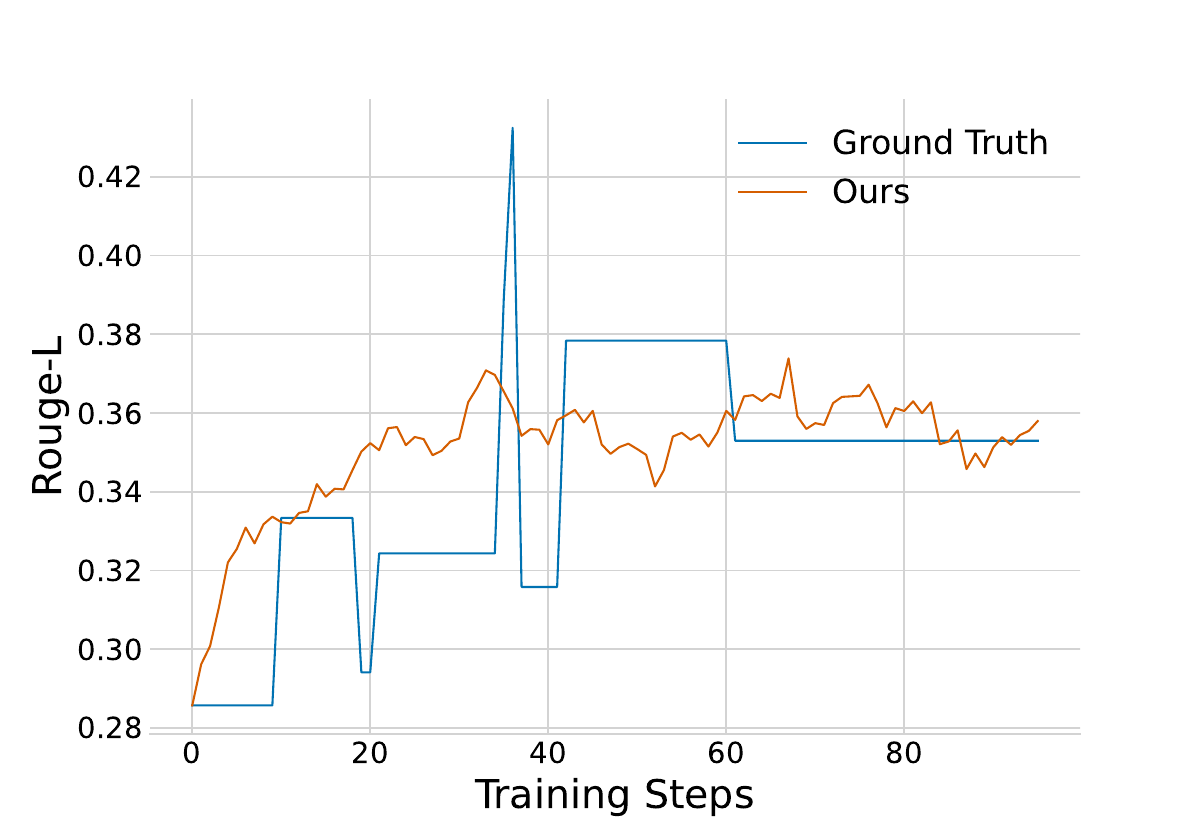}
    }
    \subfigure[Test Example3]{
        \includegraphics[width=0.3\linewidth]{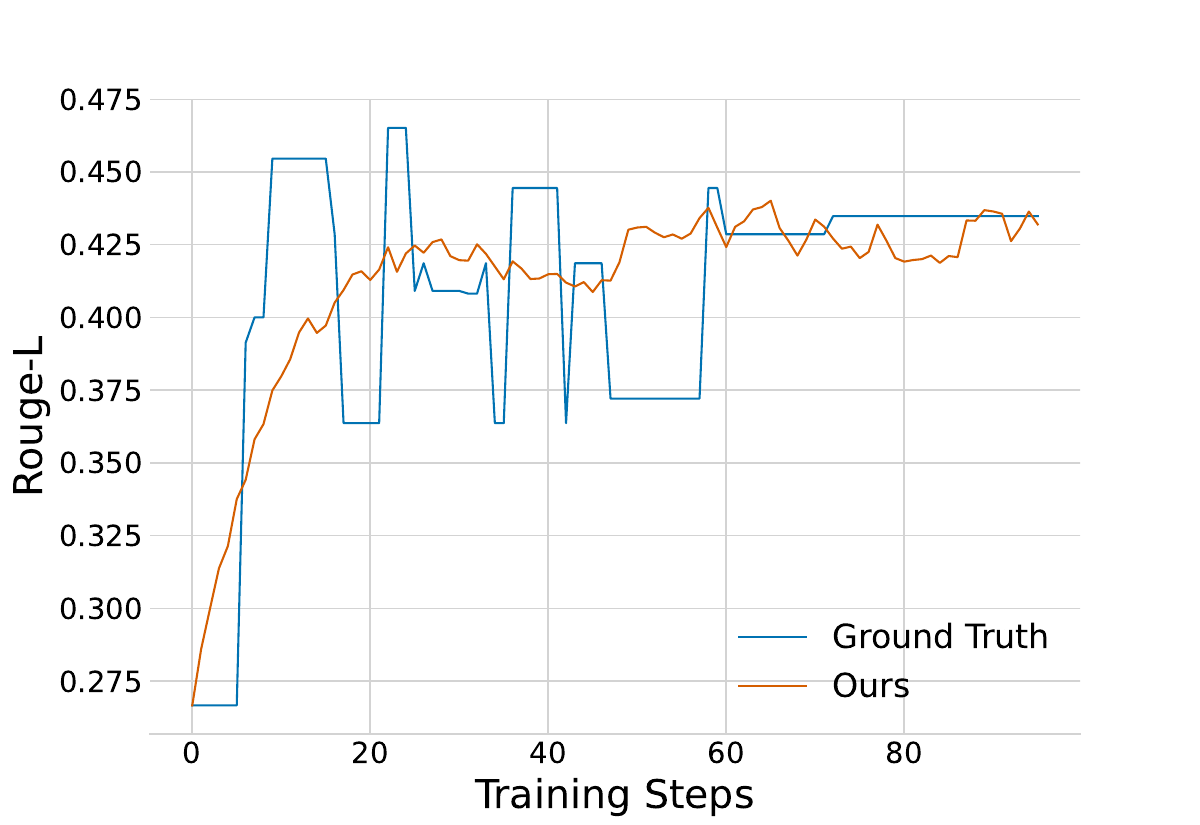}
    }   
    \caption{Examples of the ROUGE-L simulation of {\name}  for the WebNLG task on \textbf{\textit{unseen training data}}.}
    \label{fig:webnlg_rougeL_unseen_training}
\end{figure*}

\begin{figure*}
    \centering
    \subfigure[Test Example1]{
        \includegraphics[width=0.3\linewidth]{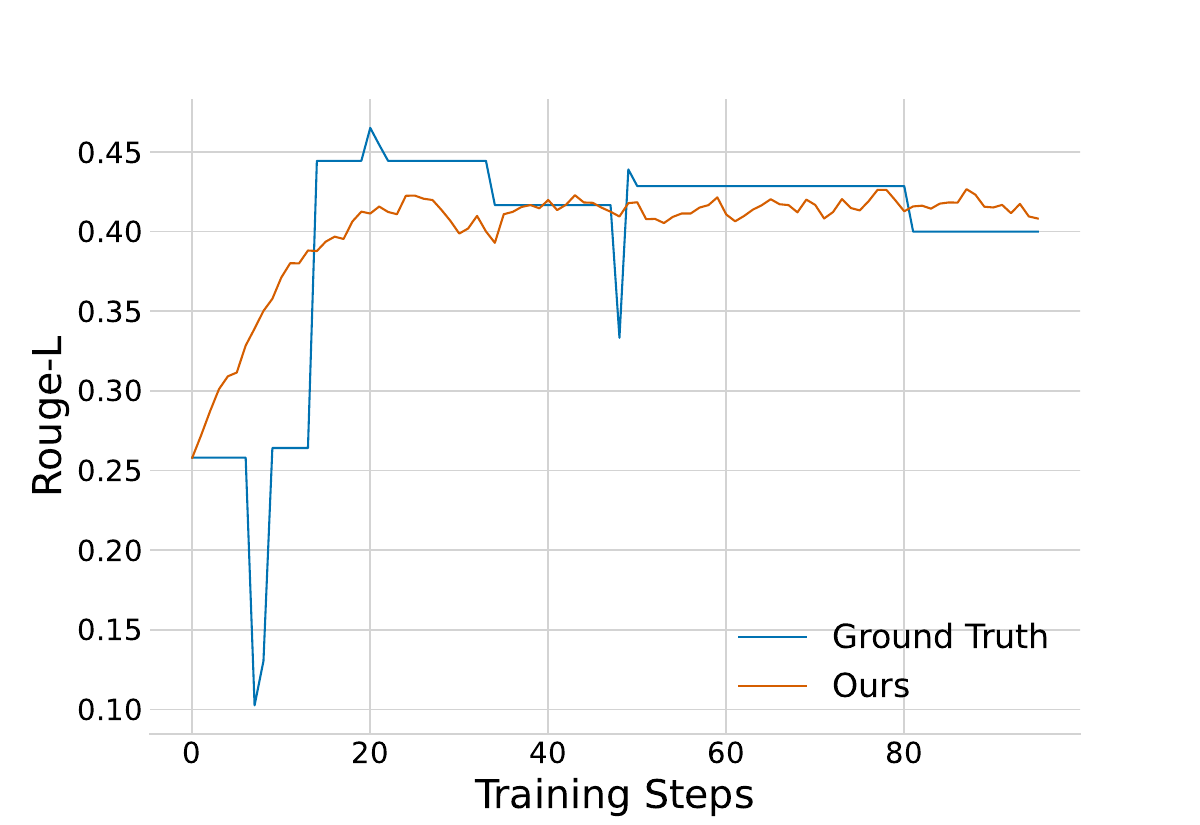}
    }
    \subfigure[Test Example2]{
        \includegraphics[width=0.3\linewidth]{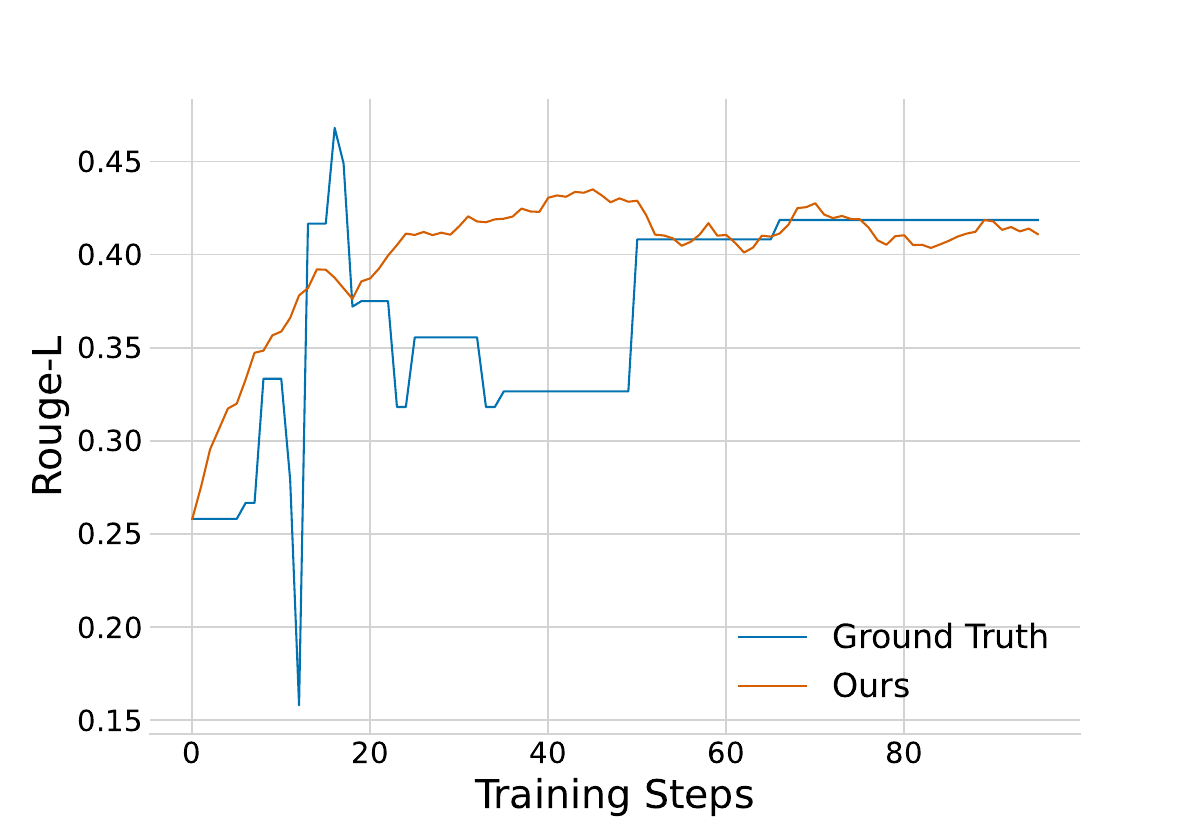}
    }
    \subfigure[Test Example3]{
        \includegraphics[width=0.3\linewidth]{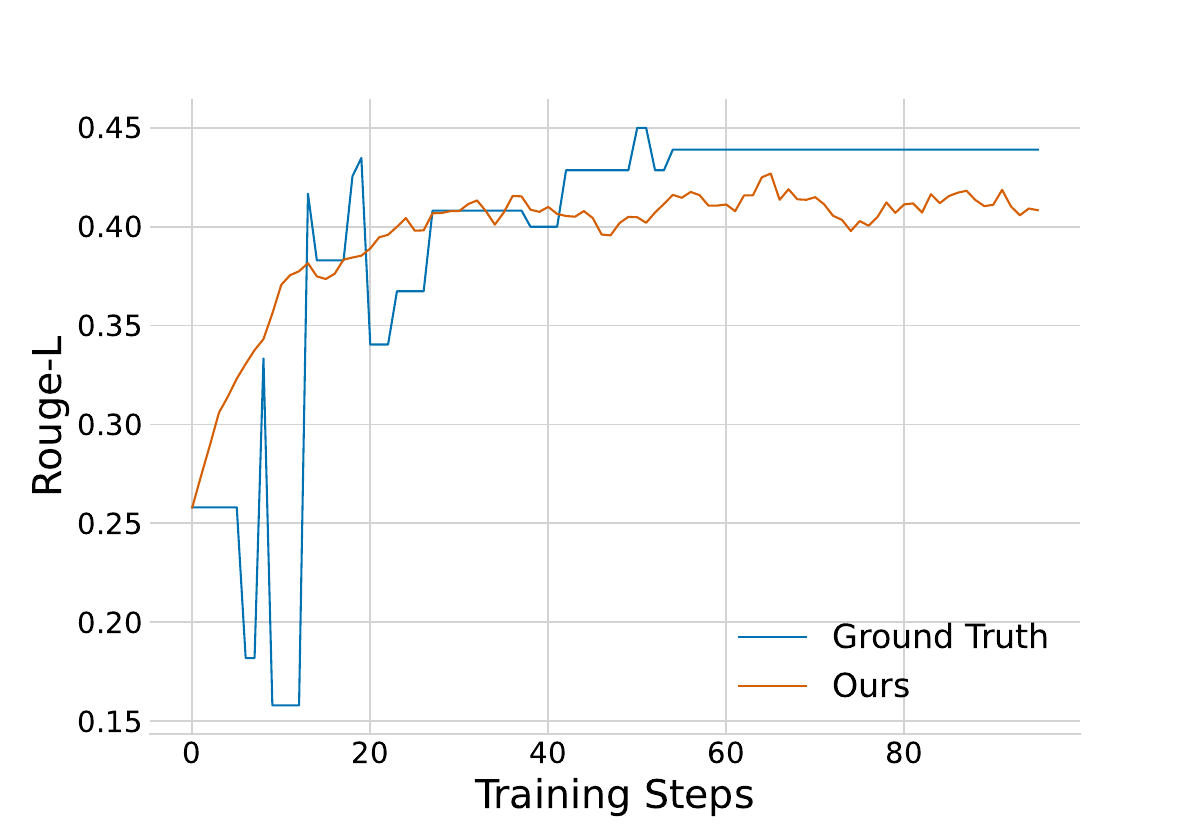}
    }   
    \caption{Examples of the ROUGE-L simulation of {\name}  for the WebNLG task on \textbf{\textit{unseen training and test data}}.}
    \label{fig:webnlg_rougeL_unseen_training_test}
\end{figure*}

We provide detailed simulations of test loss and performance metrics across different tasks and scenarios, as detailed below:
\begin{itemize}
    \item For the RTE task, test loss simulations under various conditions are presented in Fig.~\ref{fig:rte_unseen_test} (unseen test data), Fig.~\ref{fig:rte_unseen_training} (unseen training data), and Fig.~\ref{fig:rte_unseen_training_test} (unseen training and test data).
    \item For the WebNLG task, test loss simulations are shown in Fig.~\ref{fig:webnlg_loss_unseen_test} (unseen test data), Fig.~\ref{fig:webnlg_loss_unseen_training} (unseen training data), and Fig.~\ref{fig:webnlg_loss_unseen_training_test} (unseen training and test data).
    \item BLEU metric simulations for the WebNLG task are illustrated in Fig.~\ref{fig:webnlg_bleu_unseen_test} (unseen test data), Fig.~\ref{fig:webnlg_bleu_unseen_training} (unseen training data), and Fig.~\ref{fig:webnlg_bleu_unseen_training_test} (unseen training and test data).
    \item ROUGE-L metric simulations for the WebNLG task are depicted in Fig.~\ref{fig:webnlg_rougeL_unseen_test} (unseen test data), Fig.~\ref{fig:webnlg_rougeL_unseen_training} (unseen training data) and Fig.~\ref{fig:webnlg_rougeL_unseen_training_test} (unseen training and test data).
\end{itemize}

    

\end{document}